\documentclass[nohyperref]{article}

\usepackage{microtype}
\usepackage{graphicx}
\usepackage{subfigure}
\usepackage{booktabs} 

\usepackage{hyperref}
\usepackage{url}

\usepackage[accepted]{icml2022}

\usepackage{amsmath,amsfonts,bm}

\def\eqref#1{equation~\ref{#1}}

\def\1{\bm{1}}

\DeclareMathAlphabet{\mathsfit}{\encodingdefault}{\sfdefault}{m}{sl}
\SetMathAlphabet{\mathsfit}{bold}{\encodingdefault}{\sfdefault}{bx}{n}

\newcommand{\E}{\mathbb{E}}

\usepackage{amsmath}
\usepackage{amssymb}
\usepackage{mathtools}
\usepackage{amsthm}

\usepackage{amsfonts}       
\usepackage{nicefrac}       
\usepackage{xcolor}         
\usepackage{wrapfig}

\usepackage{graphicx}
\usepackage{multirow}

\usepackage{algorithm,algorithmic}
 
\usepackage{mathrsfs}
\usepackage{appendix}
\usepackage{times}
\usepackage{latexsym}
\usepackage[]{mathtools}
\usepackage[]{amsthm} 
\usepackage{bbm}
\usepackage{subfigure}
\usepackage[]{mathtools}
\usepackage{arydshln}
\usepackage{tcolorbox}

\usepackage{microtype}
\usepackage{subfigure}
\usepackage{booktabs} 
\usepackage{tikz}
\usetikzlibrary{positioning}
\usetikzlibrary{decorations.pathreplacing}
\def\##1\#{\begin{align}#1\end{align}}
\def\$#1\${\begin{align*}#1\end{align*}}
\usepackage{enumitem}
\usepackage{threeparttable}

\usepackage[capitalize,noabbrev]{cleveref}

\theoremstyle{plain}
\newtheorem{theorem}{Theorem}[section]
\newtheorem{proposition}[theorem]{Proposition}
\newtheorem{lemma}[theorem]{Lemma}

\theoremstyle{definition}
\newtheorem{definition}[theorem]{Definition}
\newtheorem{assumption}[theorem]{Assumption}
\theoremstyle{remark}

\usepackage[textsize=tiny]{todonotes}

\icmltitlerunning{To Smooth or Not? When Label Smoothing Meets Noisy Labels}

\def\tr{\mathop{\text{tr}}\kern.2ex}
\def\mps{{\mathbf{\hat{f}_{S}}(x;D)}}
\def\mph{{\mathbf{\hat{f}_{H}}(x;D)}}
\def\bs{{\mathbf{\bar{f}_{S}}(x;D)}}
\def\bh{{\mathbf{\bar{f}_{H}}(x;D)}}

\def\mc{{\text{MC}}}

\def\bfx{{\mathbf{f}(x)}}
\def\bfX{{\mathbf{f}(X)}}
\def\bfXo{{\mathbf{f}(X_1)}}
\def\by{{\mathbf y}}
\def\bny{{\mathbf{\tilde{y}}}}

\def\bi{{i}}

\def\bo{{\mathbf 1}}

\def\P{{\mathbb P}}

\def\E{{\mathbb E}}

\long\def\comment#1{}

\def\tr{\mathop{\text{Tr}}}

\newcommand{\bel}{\begin{eqnarray}\label}
\newcommand{\eel}{\end{eqnarray}}
\newcommand{\bes}{\begin{eqnarray*}}
\newcommand{\ees}{\end{eqnarray*}}

\newcommand{\squishlist}{
\begin{list}{{{\small{$\bullet$}}}}
{\setlength{\itemsep}{3pt}      \setlength{\parsep}{1pt}
\setlength{\topsep}{1pt}       \setlength{\partopsep}{0pt}
\setlength{\leftmargin}{1em} \setlength{\labelwidth}{1em}
\setlength{\labelsep}{0.5em} } }
\newcommand{\squishend}{  \end{list}  }

\definecolor{best}{HTML}{BAFFCD}
\definecolor{issue}{HTML}{FFC8BA}
\definecolor{bad}{HTML}{FFC87C}
\newcommand{\good}[1]{\cellcolor{best}#1} 
 
\newcommand{\better}[1]{\cellcolor{issue}#1}

\newcommand{\p}{\mathbb P}

\newcommand{\D}{\mathcal D}

\newcommand{\nY}{\widetilde{Y}}
\newcommand{\nD}{\widetilde{{\mathcal {D}}}}
\newsavebox\MBox

\usepackage{float}
\usepackage{array,booktabs,tabularx,multirow,colortbl,hhline}

\newtcolorbox{mybox}{colback=blue!10!white,colframe=blue!10!white,left=1mm,top=-2mm,right=1mm,boxsep=0mm,width=8cm,before=\par\smallskip\centering,after=\par,
height=1.0cm}

\newtcolorbox{mybox2}{colback=gray!10!white,colframe=gray!10!white,left=1mm,top=-2.5mm,right=1mm,boxsep=0mm,width=8.6cm,before=\par\smallskip\centering,after=\par,
height=4cm}

\newtcolorbox{mybox3}{colback=gray!10!white,colframe=gray!10!white,left=1mm,top=-1mm,right=1mm,boxsep=0mm,width=13cm,before=\par\smallskip\centering,after=\par,
height=2.3cm}

\begin{document}

\twocolumn[
\icmltitle{{To Smooth or Not? When Label Smoothing Meets Noisy Labels}}

\icmlsetsymbol{equal}{*}

\begin{icmlauthorlist}
\icmlauthor{Jiaheng Wei}{aff1}
\icmlauthor{Hangyu Liu}{aff2}
\icmlauthor{Tongliang Liu}{aff3}
\icmlauthor{Gang Niu}{aff4}
\icmlauthor{Masashi Sugiyama}{aff4,aff5}
\icmlauthor{Yang Liu}{aff1}
\end{icmlauthorlist}

\icmlaffiliation{aff1}{University of California, Santa Cruz}
\icmlaffiliation{aff2}{Brown University}
\icmlaffiliation{aff3}{TML Lab, Sydney AI Centre, The University of
Sydney}
\icmlaffiliation{aff4}{RIKEN AIP}
\icmlaffiliation{aff5}{University of Tokyo}
\icmlcorrespondingauthor{Yang Liu}{yangliu@ucsc.edu}
\icmlkeywords{Machine Learning, ICML}

\vskip 0.3in
]

\printAffiliationsAndNotice{}  

\begin{abstract}
Label smoothing (LS) is an arising learning paradigm that uses the positively weighted average of both the hard training labels and uniformly distributed soft labels. It was shown that LS serves as a regularizer for training data with hard labels and therefore improves the generalization of the model. Later it was reported LS even helps with improving robustness when learning with noisy labels. However, we observed that the advantage of LS vanishes when we operate in a high label noise regime. Intuitively speaking, this is due to the increased entropy of $\P(\text{noisy label}|X)$ when the noise rate is high, in which case, further applying LS tends to ``oversmooth" the estimated posterior. We proceeded to discover that several learning-with-noisy-labels solutions in the literature instead relate more closely to  \emph{negative/not label smoothing} (NLS), which acts counter to LS and defines as using a negative weight to combine the hard and soft labels! We provide understandings for the properties of LS and NLS when learning with noisy labels. Among other established properties, we theoretically show NLS is considered more beneficial when the label noise rates are high. We provide extensive experimental results on multiple benchmarks to support our findings too. Code is publicly available at \url{https://github.com/UCSC-REAL/negative-label-smoothing}.

\end{abstract}

\section{Introduction}
Label smoothing (LS) \citep{szegedy2016rethinking} is an arising learning paradigm that uses positively weighted average of both the hard training labels and the uniformly distributed soft label:
\begin{align}\label{eqn:ls1}
    \by^{\text{LS},r}  = (1-r)\cdot \by +\dfrac{r}{K} \cdot \bo,
\end{align}
where we denote the one-hot vector form of hard label and an all one vector as $\by$, $\bo$ respectively. $K$ is the number of label classes, and $r$ is the smooth rate in the range of $[0,1]$. 
It was shown that LS serves as a regularizer for the hard training data and therefore improves generalization of the model. The regularizer role of LS prevents the model from fitting overly on the target class. Empirical studies have demonstrated the effectiveness of LS in improving the model performance across various benchmarks \citep{pereyra2017regularizing} (such as image classification \citep{szegedy2016rethinking}, machine translation \citep{vaswani2017attention}, language modelling \citep{chorowski2017towards}), and model calibration \citep{muller2019does}. 
\begin{figure*}[!ht]
\centering
\includegraphics[width=0.35\textwidth]{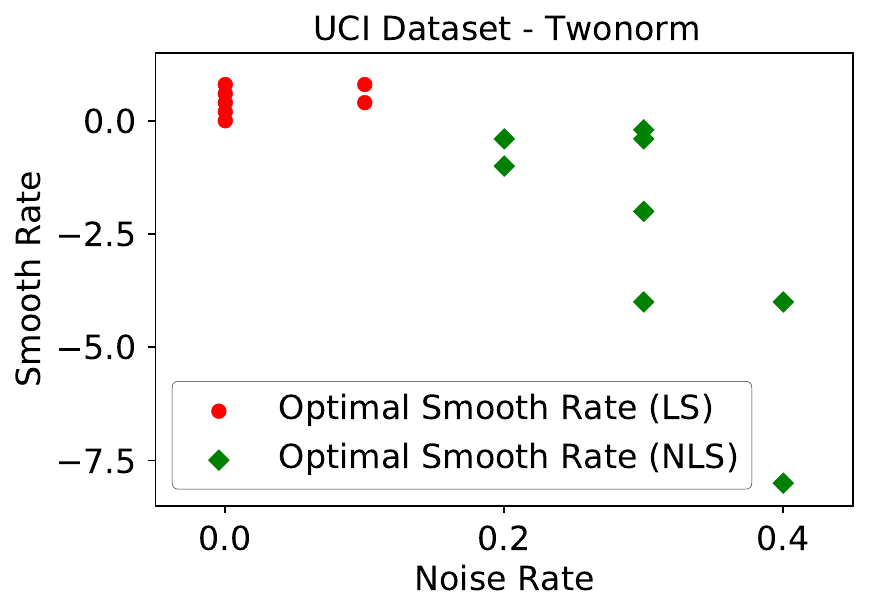}
\hspace{-0.13in}
\includegraphics[width=0.33\textwidth]{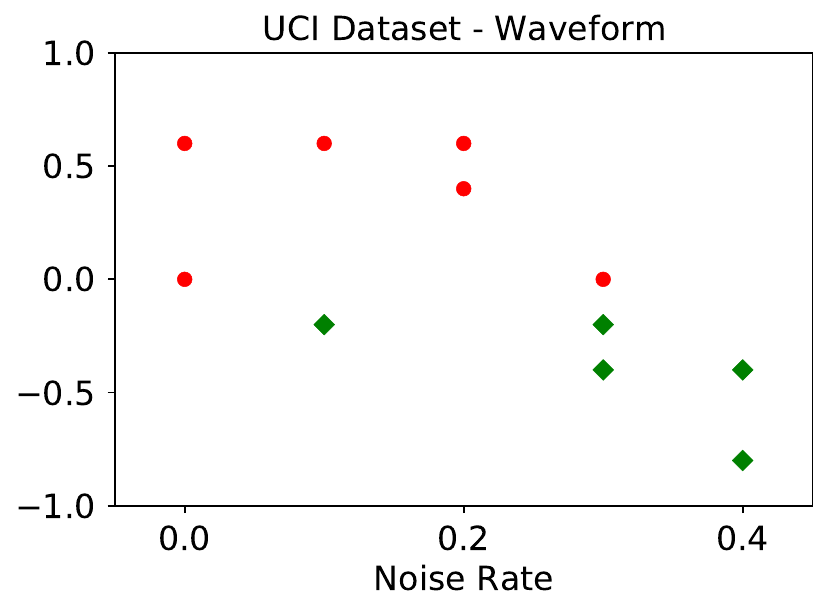}
\hspace{-0.13in}
\includegraphics[width=0.33\textwidth]{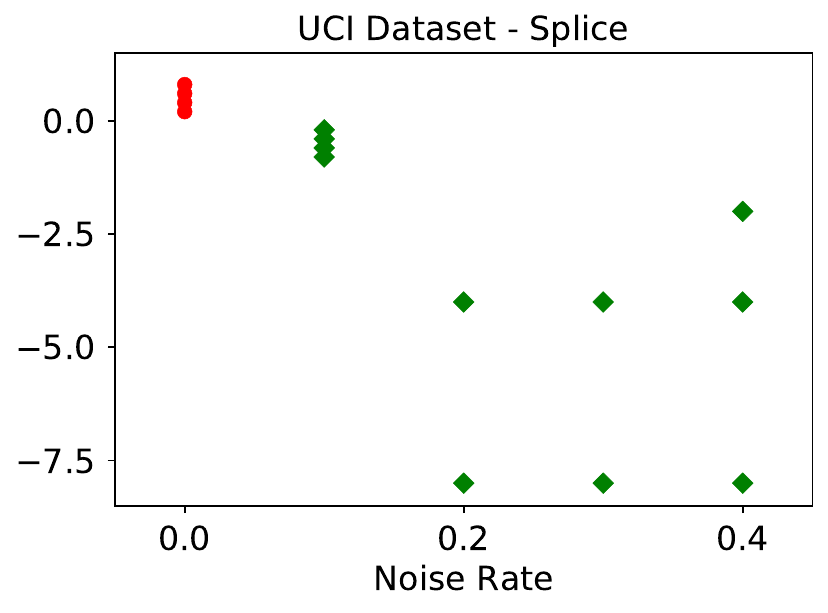}
\vspace{-0.06in}
\caption{Optimal smooth rates on UCI datasets with different label noise rates (possible to have tied smooth rates). }\label{fig:UCI}
\vspace{-0.1in}
\end{figure*}
Later it was reported LS even helps with improving robustness when learning with noisy labels \citep{lukasik2020does}. 
However, we observed that the advantage of LS vanishes when we operate in a high label noise regime: in Figure \ref{fig:UCI}, we present a set of experiments on some UCI datasets \citep{Dua:2019} {\color{black}with synthetic noisy labels}. We highlight the best two smooth rates when the classifier is trained under each label noise rate. {\color{black} Since UCI datasets are of small scales, it is possible to have tied smooth rates when evaluating the classifier on the separate clean test data.}  Indeed, non-negative smooth rates (circles colored in {\color{red}{red}}) outperform negative ones when the label noise rates are low. Nonetheless, with the increasing of noise rates, negative smooth rates $r<0$ (Eqn. (\ref{eqn:ls1}), diamonds colored in {\color{teal}{green}}) appear to be more competitive when learning with noisy labels. Intuitively speaking, this is due to the increased entropy of $\P(\text{noisy label}|X)$ when the noise rate is high, in which case, further applying LS tends to ``oversmooth" the estimated posterior. {\color{black}Motivated by this observation, we aim to provide a more thorough understanding of whether should we adopt label smoothing or not when learning with noisy labels, specifically, how to make a choice between LS and negative/not label smoothing (NLS)?

With the presence of label noise, we theoretically demonstrate that there exists a phase transition when finding the optimal label smoothing rate for $r\in(-\infty, 1]$. Particularly, when the label noise rate is low, LS is able to uncover the optimal model while NLS is considered more beneficial in a high label noise regime. Discovering that NLS differs substantially from LS in their achieved model confidence, we then proceed to explain such a transition. We also bridge the gap between NLS and several learning-with-noisy-labels solutions in the literature, including Loss Correction \citep{patrini2017making}, NLNL \citep{kim2019nlnl} and Peer Loss \citep{liu2020peer}, to further validate our results. }

We provide extensive experimental evidences to support our findings. For instance, on multiple benchmark datasets, we present the clear transition of the optimal smoothing rate going from positive to negative when we keep increasing noise rates. In particular, we show a negative smoothing rate elicits higher model confidence on correct predictions and lower confidence on wrong predictions compared with the behavior of a positive one on CIFAR-10 test data.

Our contributions summarize as follows:
\squishlist
    \item {\color{black}We provide understandings for the decision between LS and NLS, when learning with noisy labels.}
     \item We demonstrate learning with a negative smooth rate can be more robust to label noise compared with a positive rate when label noise rates are high. And this is best explained by the fact that NLS improves the confidence of model prediction. (Section \ref{sec:robust_glsr} and \ref{sec:clean_gls})
    \item 
    We show that several robust loss functions in the label-noise literature correspond to learning with NLS, under certain noise rate models. (Section \ref{sec:connect}) 
     \item Extensive empirical results validate our main theoretical conclusions. In Appendix, we discuss practical considerations to mitigate the impact of label noise, and empirically show how LS and NLS result in trade-offs in
model confidence, bias and variance of the generalization
error.
\squishend

We defer all proofs to Appendix \ref{app:proofs}. Our work primarily contributes to the literature of learning with noisy labels \citep{scott2013classification,natarajan2013learning,liu2015classification,patrini2017making,liu2020peer}.
Our core results are contingent on recent works of understanding the effect of label smoothing when training deep neural network models, {\color{black} i.e., label smoothing improves model calibration \citep{muller2019does}, more complicated forms of label smoothing \citep{li2020regularization,yuan2020revisiting}, and in particular when label noise presents \citep{lukasik2020does,liu2021importance}.} Due to the space limit, we defer a more detailed discussion of related works to Appendix \ref{app:full_related}.

\section{Preliminaries}

\subsection{Learning with noisy labels}

For a $K$-class classification task, we denote by $X\in \mathcal{X}$ a high-dimensional feature and $Y\in \mathcal{Y} :=\{1,2,...,K\}$ the corresponding label. Suppose $(X, Y)\in \mathcal{X} \times \mathcal{Y}$ are drawn from a joint distribution $\mathcal{D}$. The noisy label literature \citep{natarajan2013learning,liu2015classification,patrini2017making} considers the setting where we only have access to samples with noisy labels from $(X, \nY)$. 
Suppose random variables $(X, \nY)\in \mathcal{X} \times \widetilde{\mathcal{Y}}$ are drawn from a noisy joint distribution $\mathcal{\nD}$.  

Statistically, the random variable of noisy labels $\nY$ can be characterized by a noise transition matrix $T$, where each element $T_{i,j}$ represents the probability of flipping the clean label $Y=i$ to the noisy label $\nY = j$, i.e., $T_{ij}=\mathbb P(\nY=j|Y=i).$ 
In this paper, we concentrate on the widely adopted class-dependent label noise \citep{natarajan2013learning,liu2015classification,patrini2017making}, which assumes that the label noise is conditionally independent of features $X$, i.e.,
\vspace{-0.06in}
\begin{equation*}\label{Eq:CondIndep}
\p(\nY = j| Y = i) =\p(\nY = j| X, Y = i), \forall i,j \in [K].
\end{equation*}
For the binary classification setting, define  $e_0:=\mathbb{P}(\nY=1|Y=0)$, $e_1:=\mathbb{P}(\nY=0|Y=1)$. Without loss of generality, we assume $e_1 - e_0=e_{\Delta}\geq 0$. The binary noise transition matrix in the noisy label setting then becomes: 
{$$T=
    \begin{pmatrix}
         & 1-e_0  & e_0  \\
         & e_1 & 1 - e_1 & 
    \end{pmatrix}.
  $$}
\vspace{-0.15in}
\subsection{Learning with smoothed labels}
Let $\by_{\bi}$ be the one-hot encoded vector form of $y_i$ which generates according to $Y$. The random variable of smoothed label $ Y^{\text{LS},r}$ with smooth rate $r\in [0,1]$  generates  $\by_{\bi}^{\text{LS},r} $ as \citep{szegedy2016rethinking}:
\begin{align*}
    \by_{\bi}^{\text{LS},r}  = (1-r)\cdot \by_\bi +\dfrac{r}{K} \cdot \bo.
\end{align*}
For example, when $r=0.3$, the smoothed label of $\by_\bi=[1, 0, 0]^{\top}$ becomes $\by_\bi^{\text{LS},r=0.3}=[0.8, 0.1, 0.1]^{\top}$.

{\color{black}To enable ease of presentations (instead of highlighting a crucial concept), we unify LS \citep{szegedy2016rethinking,lukasik2020does} and NLS into the generalized label smoothing (GLS), i.e., $r \in (-\infty, 1]$:}
\begin{align}
 \by_{\bi}^{\text{GLS},r}:=(1-r)\cdot \by_{\bi}+\dfrac{r}{K}\cdot \bo,
\end{align}
where $\by_{\bi}^{\text{GLS},r}$ is given by the random variable of generalized smooth label $Y^{\text{GLS},r}$. We name the scenario $r<0$ as negative/not label smoothing (NLS). A negative $r$ indicates that the smoothed label might be negatively related to the corresponding feature and should not be (positively) smoothed. For example, when $r=-0.3$, the smoothed label of $\by_\bi=[1, 0, 0]^{\top}$ becomes $\by_\bi^{\text{GLS},r=-0.3}=[1.2, -0.1,  -0.1]^{\top}$.
We observe that the entries in $ \by_{\bi}^{\text{GLS},r}$ still add up to 1: $1-r + \frac{r}{K} \cdot K = 1$. Nonetheless we want to point out $ \by_{\bi}^{\text{GLS},r}$ is no longer a valid probability measure since for entries $y \neq y_i$, the corresponding weight will be negative ($\frac{r}{K}$) when $r<0$. This points us to the definition of an extended label distribution:
\begin{definition}[Extended label distribution]
We call $\by$ an extended label distribution if $\bo^{\top} \by =1$, but $\by$ is not necessarily entry-wise non-negative.
\end{definition}

\paragraph{What negative labels really mean}
Negative label smoothing is indeed one of such extended label distribution. We proceed the illustration using the previous three-class classification example: a one-hot label $[1, 0, 0]^{\top}$ (three elements stand for class dog (0), cat (1), deer (2), respectively) means this sample $x$ is categorized as a dog and is irrelevant to class cat and deer. LS $[0.8, 0.1, 0.1]^{\top}$ indicates that the representation $x$ might encode uncertainty and is slightly related to cat/deer (positive correlation between cat/deer and dog given $x$). NLS $[1.2, -0.1, -0.1]^{\top}$ not only implies high confidence in label dog, but it is even more so that predicting cat (1) or deer (2) should be penalized by 0.1, i.e., given any loss $\ell$ that is linear in $y$ (e.g., CE loss), this $x$ receives the loss $1.2 \cdot \ell(\mathbf{f}(x),0)- 0.1 \cdot \ell(\mathbf{f}(x),1)- 0.1 \cdot \ell(\mathbf{f}(x),2)$. Such a penalization mechanism is not uncommon and it appeared in the design of backward loss correction \citep{natarajan2013learning} $\sum_{i\in\{0,1,2\}}T^{-1}_{0, i}\cdot \ell(\mathbf{f}(x), i)$ with $T^{-1}_{0, 1}, T^{-1}_{0, 2}\leq 0$ ($T^{-1}$ is the inverse matrix of $T$), peer loss \citep{liu2020peer} $\ell(\mathbf{f}(x), 0){-\ell(\mathbf{f}(x),y_{\text{rand}})}$ where $y_{\text{rand}}=i$ with probability $\p(\nY=i)$ for $i\in \{0, 1, 2\}$, and complementary loss (more details in Section \ref{sec:connect}).

We will present the surprising power of negative labels in handling label noise, though a bit counter-intuitive at first sight. To clarify, although we may adopt the negative label for calculating the loss, the model prediction is processed by the soft-max function, so the prediction still lies on the $K$-simplex. Besides, we do not assume a strict lower bound for $r$. If $r\to -\infty$, normalizing $\by_{\bi}^{\text{GLS},r}$ by $1-r$ returns $\by_{\bi}^{\text{GLS},r}=\by_{\bi} -\frac{\bo}{K}$. We will show when imposing a negative smoothing parameter will be considered beneficial as compared to a positive one. In the main paper, we mainly focus on the binary classification task where $y_i\in \{0, 1\}$ and $K=2$, although we do include the discussion of multi-class extensions in Section \ref{app:multi}.

\subsection{Model confidence}
 Denote  a deep neural network as $f$, $\mathbf{f}(x_i)$ is the model prediction of $x_i\in X$ with element $\mathbf{f}(x_i)_{y_i}:=\p(Y=y_i|X=x_i, f)$, the binary cross-entropy loss is then defined as $\ell_{\text{CE}}(\mathbf{f}(x_i), y_i):=-\log(\mathbf{f}(x_i)_{y_i})$.  Throughout this paper, we shorthand $\ell_{\text{CE}}$ as $\ell$ for a clean presentation.
We define a key quantity, model confidence, that plays an important role in later sections. 

\begin{definition}[Confidence of model $f$ for sample $(x,y)$] 
\label{df: mc}
Given a model $f$, a sample $x$ with its target label $y\in \{0, 1\}$, the model confidence of $f$ w.r.t. sample $x$ is defined as $\text{MC}(f;x,y)=\bfx_{y}-\bfx_{1-y}.$
\end{definition}
$\text{MC}(f;x,y)$ in Definition \ref{df: mc} characterizes the difference of the predicted probability between the target class and the other class. $\text{MC}(f;x,y)=0$ simply means $f$ has no confidence on its predictions since the model can not identify the target class of $x$. $\text{MC}(f;x,y)$ is negative when $f$ gives a wrong prediction and is not confident to predict the label of $x$ as the target label $y$. To dig into how GLS influences the model confidence on correct and wrong predictions in following sections, we separate the distribution $\D$ into:
\begin{align*}
    \D^+_{f} &:= \{(X, Y)\sim \D:\text{MC}(f;X,Y) > 0 \}, \\
    \D^-_{f} &:= \{(X,Y)\sim \D:\text{MC}(f;X,Y) \leq 0\}.
\end{align*}
Similarly, we introduce the confidence of model prediction under the metric of $\ell$-loss as:
\begin{definition}[$\ell$-based confidence of model $f$ for sample $(x,y)$] 
\label{df: mcl}
Given a model $f$, a sample $x$ with its target label $y\in \{0, 1\}$, the $\ell$-based model confidence of $f$ w.r.t. sample $x$ is defined as $\mc_{\ell}(f;x, y):=-(\ell(\bfx, y)-\ell(\bfx, 1-y)).$
\end{definition}

\section{To Smooth or Not? In the View of Risk Minimization}\label{sec:robust_glsr}
{\color{black}
In this section, we aim to characterize the optimal candidates of $r$ in the unified setting to distinguish the preferences for LS and NLS, when the label noise presents.} 

Let $\bny$ be the vector form of noisy label $\tilde{y}$ obtained from $\nY$. For $r\leq 1$, we define the $r$-smoothed label of $\tilde{y}$ as $\bny^{\text{GLS},r}$, where $\bny^{\text{GLS},r}:=(1-r)\cdot \bny +\frac{r}{K}\cdot \bo$ and is generated by the random variable $\nY^{\text{GLS},r}$. Risk minimization w.r.t. smoothed noisy label distribution $\nY^{\text{GLS},r}$ is then defined as:
\begin{align}
\label{eqn:gen}
    \min_{f \in \mathcal F} ~\mathbb{E}_{(X, \nY)\thicksim \nD} \Big[\ell(\bfX, \nY^{\text{GLS},r})\Big],
\end{align}
where in above $\mathcal F$ is the hypothesis space we consider.
In Figure \ref{fig:UCI}, we have shown that given the unseen test data, learning with non-negative smooth rates may not always return the best outcome.
Based on this observation, we delve into details to show when NLS is more favorable than LS and Vanilla Loss (VL, $r=0$). We start with stating Assumption \ref{ass:opt_r}:
\begin{assumption}
\label{ass:opt_r}
We assume learning with clean data distribution $\D$ with smooth rate $r^*\leq 1$ in GLS makes the corresponding classifier $f^*_{\mathcal{D}}$ return the best performance on the unseen clean test data distribution $\D_{test}${\color{black}, where $f^*_{\mathcal{D}}$ is given by: $f^*_{\mathcal{D}}\leftarrow \arg \min_{f\in\mathcal{F}} \mathbb{E}_{(X,Y)\sim \mathcal{D}}[
\ell(\mathbf{f}(X), Y^{\text{GLS}, r^*})]$.}
\end{assumption}
Assumption \ref{ass:opt_r} simply offers us a view to initiate our analysis for the noisy label setting. To clarify, the expected risk of random variables could be approximated/replaced by the empirical one over a finite number of samples: i.e., when $\D=\{x_i, y_i\}_{i=1}^{N}, \nD=\{x_i, \tilde{y}_i\}_{i=1}^{N}$, Eqn. (\ref{eqn:gen}) becomes: $\min_{f \in \mathcal F} \frac{1}{N}\sum_{i\in [N]} \ell(\mathbf{f}(x_i), \tilde{y}_i^{\text{GLS},r}).$ In this case, {\small$\mathcal{D}$} represents the empirical distribution for the finite dataset, and {\small$\mathcal{D}_{\text{test}}$} can be thought of as the expected risk with infinite samples. With this being said, our analysis does require taking the expectation over the noisy labels (over $\widetilde{Y}|X,Y$). Besides, we don't rule out the possibility that other methods outperform LS, VL or NLS with optimal smooth rate $r^*$. At the end of this section and Appendix \ref{app:more_exp}, we will empirically test what $r^*$ usually is on various benchmarks. We denote the $r^*$ smoothed label distribution as $Y^*$: $ Y^*= Y^{\text{GLS}, r^*}$. 
With the introduction of $r^*$ and $f_{\D}^*$, our goal is then to recover the classifier $f_{\D}^*$ using the noisy training labels. We define $\lambda_1, \lambda_{2}$ and offer Theorem \ref{thm:glsr_conf}.
\[\lambda_{1}:=\Big[(e_0-\frac{r^*}{2}) + (1-2e_0)\cdot \frac{r}{2}\Big], \quad \lambda_2:=e_{\Delta}\cdot (1- r).\]
\begin{theorem}
\label{thm:glsr_conf}
The risk minimization w.r.t. $\nY^{\text{GLS},r}$ in the noisy setting (Eqn. (\ref{eqn:gen})) is equivalent to the risk w.r.t $Y^*$ defined on the clean data, with two additional bias terms:
\vspace{-0.1in}
\begin{mybox2}
\begin{align}
\label{eq:affine}
     \min_{f \in \mathcal F} ~&\underbrace{ \mathbb{E}_{(X, Y)\sim \D}   \Big[\ell(\bfX, Y^*)\Big] }_{\text{True Risk}}\notag\\&\underbrace{+ \lambda_1 \cdot \mathbb{E}_{(X, Y)\sim \D}  \Big[\ell(\bfX, 1-Y) - \ell(\bfX, Y)\Big]}_{\text{M-Inc1}}\notag\\ 
    &\underbrace{+\lambda_2 \cdot \mathbb{E}_{X, Y=1}  \Big[\ell(\bfX, 0) - \ell(\bfX, 1)\Big]}_{\text{M-Inc2}}.
\end{align}
\end{mybox2}
\end{theorem}
\vspace{-0.1in}
Remember that $\ell=\ell_{\text{ce}}$, we have: $\mc_{\ell}(f;X, Y)=\log\left(\bfX_{Y}/\left(1-\bfX_{Y}\right)\right)$, $\mc(f;X,Y)=2\bfX_{Y}-1$. Both $\log(\frac{x}{1-x})$ and $2x-1$ are monotonically increasing for $x\in (0, 1)$, model $f$ with a high $\mc_{\ell}(f;X, Y)$ has high  $\mc(f;X,Y)$. The two extra bias terms explicitly affect the model confidence. Now we proceed to answer ``what $r$ is preferred in the noisy setting''.

\subsection{Symmetric noise rates with $e_{\Delta}=0$}
Symmetric noise rates $e:=e_0=e_1$ indicates the probability of flipping to the other class is equal for both classes. In this case, $\lambda_2 = 0$, Term $\text{M-Inc2}$ is cancelled and Eqn. (\ref{eq:affine}) reduces to 
\begin{align}
    &\min_{f \in \mathcal F} ~ \underbrace{ \mathbb{E}_{(X, Y)\sim \D}   \Big[\ell(\bfX, Y^*)\Big] }_{\text{True Risk}}\notag\\&~\underbrace{+\lambda_1 \cdot \mathbb{E}_{(X, Y)\sim \D}   \Big[\ell(\bfX, 1-Y) - \ell(\bfX, Y)\Big]}_{\text{M-Inc1}}.
\end{align}
\paragraph{Noisy labels impair model confidence on Vanilla Loss}
\label{sec:exp_mc}
In the unified framework, define the optimal $r$ that will cancel the impact of Term $\text{M-Inc1}$ as:
\begin{align}
\label{eq:thred}
   \text{when}~ r_{\text{opt}} := \frac{r^*-2e}{1-2e}, ~~~\text{M-Inc1} = 0.
\end{align}
The threshold $r_{\text{opt}}$ in Eqn. \ref{eq:thred} implies:

\begin{theorem}
\label{thm:robust_glsr}
With Assumption \ref{ass:opt_r}, learning with smooth rate $r = r_{\text{opt}}$ under $(X, \nY)\thicksim \nD$ yields $f_{\D}^*$:
\squishlist
    \item When noise rate $e <r^*/2$,  $r=r_{\text{opt}}>0$ (LS);
    \item When noise rate $e =r^*/2$,  $r=0$ (VL);
    \item When noise rate $e >r^*/2$,  $r=r_{\text{opt}}<0$ (NLS). 
\squishend
\end{theorem}

\begin{figure}[!ht]
  \begin{center}
    \includegraphics[width=0.3\textwidth]{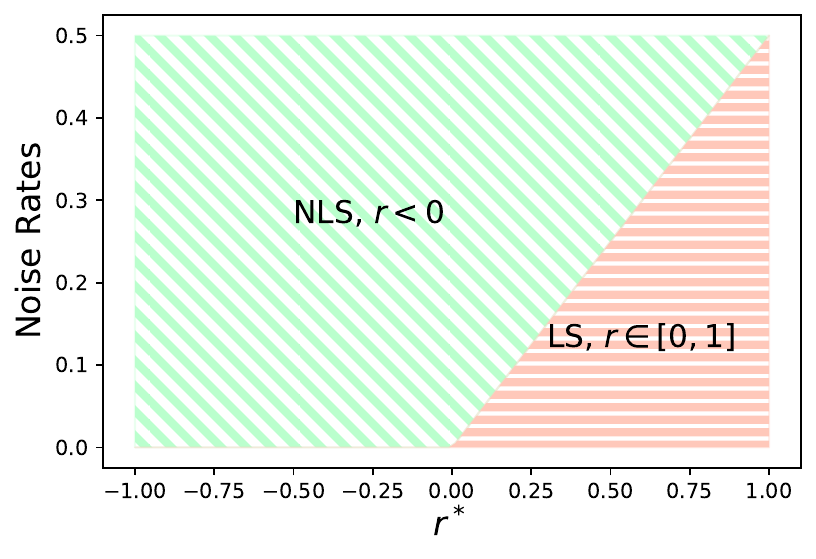}
  \end{center}
    \vspace{-15pt}
  \caption{Decision between NLS, LS given $e$, $r^*$.}
    \label{fig:ill_thres}
  \vspace{-10pt}
\end{figure}
In Theorem \ref{thm:robust_glsr}, adopting NLS when noise rate $e<\frac{r^{*}}{2}$ induces $\lambda_1<0$, Term M-Inc1 makes $f$ overly-confident on its predictions compared with $Y^*$. In Figure \ref{fig:ill_thres}, with the decreasing of $r^*$, LS is less tolerant of labels with high noise. Similarly, if $e \geq \frac{r^*}{2}$, with the decreasing of $r^*$, NLS is more robust in the high noise regime while LS makes the model $f$ become less-confident on its predictions. Clearly, NLS outperforms LS especially when noise rates are large and $r^*$ is small.

\subsection{Asymmetric noise rates with $e_{\Delta}\neq 0$}
In this case, adopting $r=\frac{r^*-2e_0}{1-2e_0}$ removes the Term M-Inc1. However, when $r<1$, Term M-Inc2 is not negligible due to assymetric noise transition matrix. As a result,  Term M-Inc2 becomes:
\begin{align*}
     e_{\Delta}\cdot \frac{1-r^*}{1-2e_0} \cdot\mathbb{E}_{X, Y=1}  \Big[\ell(\bfX, 0) - \ell(\bfX, 1)\Big],
\end{align*}
$\text{with }e_{\Delta}\cdot \frac{1-r^*}{1-2e_0}\geq 0$. Term M-Inc2 in the minimization increases the model confidence on $(X, Y=0)\sim \D^+_f$. The model will then become overly-confident with the class that has a low noise rate $e_0$. Meanwhile, Term M-Inc2 decreases the model confidence on $(X, Y=1)\sim \D^+_f$ (less-confident to the class with a high noise rate $e_1$). 
{\color{black}\subsection{Analysis of empirical risks}}
The popularity of LS is largely due to its effectiveness in practice, i.e., through the optimization of the smoothed empirical risk. Given the smooth rate $r$, potentially we could adopt the Rademacher bound on the maximal deviation between the expected risk $R_{\text{{exp}}}^r(f)$ (objective in Eqn. (\ref{eqn:gen})) and the empirical risk $R_{\text{{emp}}}^r(f):=\frac{1}{N}\sum_{i\in [N]} \ell(\mathbf{f}(x_i), \tilde{y}_i^{\text{GLS},r})$ when learning with noisy labels, formally, we have:
\begin{theorem}\label{thm:radmaker}
With probability at least $1-\delta$, we have:
{\small\begin{align*}
    &\max_{f\in\mathcal{F}}~|R_{\text{{emp}}}^r(f)-R_{\text{exp}}^r(f)|\\
    \leq &(2+|r|-r)\cdot L\cdot \mathfrak{R}(\mathcal{F})+(1-r)\cdot \left(\overline{\ell}-\underline{\ell}\right)\cdot\sqrt{\frac{\log(1/\delta)}{2N}},
\end{align*}}
where $\overline{\ell}, \underline{\ell}$ denote the upper/lower bound of $\ell$, $\mathfrak{R}$ is the Rademacher complexity.
\end{theorem}
Theorem \ref{thm:radmaker} bridges the gap between the expected risk $R_{\text{{exp}}}^r(f)$ and the empirical risk $R_{\text{{emp}}}^r(f)$ by offering an upper bound. Intuitively, with a large sample size $N$ and a low Rademacher complexity of the hypothesis space $\mathfrak{R}(\mathcal{F})$, $R_{\text{{emp}}}^r(f)$ is supposed to well-approximate $R_{\text{{exp}}}^r(f)$. When learning with finite samples, LS is popular by referring to its impacts on reducing the model confidence (or avoids over-fitting). NLS indeed may force model become confident on the prediction, including wrong ones. What we observe in practice is that neural nets firstly memorize on easy/clean samples \citep{liu2020early}, warm-up with CE and then switch to NLS significantly improves the model performance. Since in the latter stage, the model is encouraged to be more confident on learned patterns (clean samples) and less likely to over-confident on samples with wrong labels (large-loss samples). When noise rate is high, the noisy training data is already over-smoothing the training process. Think of the noisy label flipping corresponding to a certain smooth rate, and a case where a certain representation $x$, with its similar patterns, are sampled multiple times - then their associated noisy labels formed a smoothed distribution. In this case, applying the NLS corrects the over-smoothness.

\begin{table*}[!htb]
	\caption{Test accuracies of LS, VL, NLS on clean and noisy UCI Heart, Splice datasets, with best two smooth rates highlighted (green: NLS; red: VL or LS). We adopt the two independent sample T-test (5 non-negative smooth rates V.S. the last 5 rows of reported negative smooth rates) to verify the overall performance comparisons between VL/LS and NLS. $p$-value is highlighted in green if NLS generally returns a higher accuracy (i.e., $t$-value$<0$) than VL/LS, otherwise, in red. Results on more benchmark datasets are given in Appendix \ref{app:more_exp}. }
\small
		\begin{center}
		\scalebox{0.82}{\begin{tabular}{c|ccccc|ccccc}
				\hline 
		\multirow{2}{*}{Smooth Rate}  &  \multicolumn{5}{c}{\emph{UCI-Heart}}&  \multicolumn{5}{c}{\emph{UCI-Splice}}\\ 
				  &$e_i=0$ &$e_i=0.1$&$e_i=0.2$   &$e_i=0.3$ &$e_i=0.4$  &$e_i=0$ &$e_i=0.1$&$e_i=0.2$   &$e_i=0.3$ &$e_i=0.4$\\
				\hline\hline
			   $r=0.8$	  & 0.885 & 0.853  &  0.836 & 0.820  & 0.738& \better{0.980} & 0.946 & 0.919 & 0.856 & 0.760 \\
			   	 \hline 
			  $r=0.6$ 	   & \better{0.902} & 0.836  & 0.820  & 0.836  & 0.738 	&  \better{0.978} & 0.939 & 0.913 & 0.869 & 0.778 \\
			   	 \hline 
			   $r=0.4$	    & 0.885 & 0.853  & 0.836  & 0.820  &  0.771 & \better{0.978} & 0.948 & 0.922 &0.885  & 0.797\\
			   	 \hline 
			  $r=0.2$ 	    & \better{0.902} & 0.853  & 0.820  & 0.803  &  0.754 & \better{0.978} & 0.948 & 0.919 & 0.878 & 0.800\\
			   	 \hline 
			   $r=0.0$	   & \better{0.902} & 0.853  & 0.820  & 0.820  & 0.771 & 0.976 & 0.948 & 0.926 & 0.876 & 0.806  \\
			   	 \hline 
			  $r=-0.4$ 	  &   0.869  & 0.836  & 0.803  & \good{0.853}  & 0.754 & 0.961 & \good{0.956} & 0.928 & 0.880 & 0.817 \\
			  \hline 
			  $r=-0.6$	   & 0.869  & 0.836  & 0.820   & \good{0.853}  & 0.721 & 0.961 & \good{0.956} & 0.926 & 0.880 & 0.819  \\
			  \hline 
			  $r=-1.0$ 	  & 0.885   & \good{0.869}  & 0.803   & \good{0.853}  & 0.754 & 0.956 & 0.954 & 0.932 & 0.889 & 0.819\\
			   \hline 
			  $r=-2.0$ 	  & 0.885  & \good{0.869}   & 0.820   & \good{0.853} & 0.787 & 0.952 & 0.946 & 0.935 & 0.898 & \good{0.830} \\
			   \hline 
			  $r=-4.0$ 	  & 0.885   & \good{0.869}   & \good{0.853}   & \good{0.885} &  \good{0.820}  & 0.946 & 0.943 & \good{0.939} & \good{0.911} & \good{0.830}\\
			   \hline 
			  $r=-8.0$ 	 & 0.869   & \good{0.869} & \good{0.885}   & \good{0.853}  & \good{0.853} &  0.943 & 0.946 & \good{0.939} &\good{ 0.915} & \good{0.845}\\ 			  \hline\hline
			  $r_{\text{opt}}=$ & \better{$[$0.0, 0.6$]$} & \good{$[$-8.0, -1.0$]$} & \good{-8.0} & \good{-4.0} & \good{-8.0}& \better{$ 0.8$} & \good{$[$-0.6, -0.4$]$} & \good{$[$-8.0, -4.0$]$} & \good{-8.0} & \good{-8.0}
			  \\\hline\hline
			  $p\text{-value}=$ & \better{0.020} & \good{0.136} &\good{0.549} & \good{0.002} & \good{0.243}&\better{0.001} &\good{0.332}  &\good{0.002} & \good{0.015} &\good{0.005}
			  \\
	 	\hline		\end{tabular}}
			\end{center}
		\label{Tab:uci_paper1}
	\end{table*}

\subsection{Multi-class extension}\label{app:multi}

As an extension to the binary classification task, we next show how Theorem \ref{thm:robust_glsr} could be generalized to the multi-class setting under two broad families of noise transition model. We assume Assumption \ref{ass:opt_r} holds in the multi-class setting. And for $Y, \nY\in [K]$, we extend the definition of model confidence to multi-class classification tasks as:
\begin{definition}[Model confidence of sample $(x,y)$ ($K$-class classification)] 
\label{df: mc-multi}
Given a model $f$, a sample $x$ with its target label $y\in [K]$, the model confidence score of $f$ w.r.t. sample $x$ is defined as $\text{MC}(f;x,y)=\bfx_{y}-\frac{1}{K-1} \sum_{i\neq y}\bfx_{i}$.
\end{definition}\vspace{-0.1in}
\paragraph{Sparse noise transition matrix}
Sparse noise model \citep{wei2021when} assumes $K$ is an even number. For $c\in [\frac{K}{2}]$, $i_c<j_c$, sparse noise model specifies $\frac{K}{2}$ disjoint pairs of classes $(i_c, j_c)$ to simulate the scenario where particular pairs of classes are ambiguity and misleading for human annotators. The off-diagonal element of $T$ reads $T_{i_c, j_c}=e_0$, $T_{j_c, i_c}=e_1$. Suppose $e_0+e_1<1$, the diagonal entries become $T_{i_c, i_c}=1-e_0$, $T_{j_c, j_c}=1-e_1               $. Clearly, our conclusions in Theorem \ref{thm:robust_glsr} extends directly to the sparse noise transition matrix by simply splitting the $K$-class classification task into $\frac{K}{2}$ disjoint binary ones. 

\paragraph{Symmetric noise transition matrix}

Symmetric noise model \citep{kim2019nlnl} is a widely accepted synthetic noise model in the literature of learning with noisy labels. The symmetric noise model generates the noisy labels by randomly flipping the clean label to the other possible classes with probability $\epsilon$. $\forall i\neq j$, $T_{i,j}=\frac{\epsilon}{K-1}$, and the diagonal entry is $T_{i,i}=1-\epsilon$. Define the optimal $r$ under the unified setting in the multi-class setting as $r_{\text{opt}} := \frac{(K-1)\cdot r^*-K\cdot \epsilon}{(K-1)-K\cdot\epsilon}$, Theorem \ref{thm:robust_glsr} can be extended to the multi-class setting as:
\begin{theorem}
\label{thm:robust_glsr_multi}
Under Assumption \ref{ass:opt_r}, suppose the symmetric noise rate is not too large, i.e, $\epsilon<\frac{K-1}{K}$, learning with smooth rate $r = r_{\text{opt}}$ under $(X, \nY)\thicksim \nD$ yields $f_{\D}^*$:
\squishlist
\vspace{-5pt}
    \item When noise rate $\epsilon < \frac{(K-1)\cdot r^*}{K}$,  $r=r_{\text{opt}}>0$ (LS);
    \item When noise rate $\epsilon =\frac{(K-1)\cdot r^*}{K}$,  $r=0$ (VL);
    \item When noise rate $\epsilon >\frac{(K-1)\cdot r^*}{K}$,  $r=r_{\text{opt}}<0$ (NLS).
\squishend
\end{theorem}

\subsection{Clean empirical risk v.s. noisy empirical risk}
Now we empirically verify Theorem \ref{thm:glsr_conf} under symmetric noise setting, which relates the risk in the noisy setting to the clean ones. Assume the the noise label is generated through the symmetric noise transition matrix. We name the noisy risk as $\mathbb{E}_{(X, \nY)\thicksim \nD} \left[\ell(\bfX, \nY^{\text{GLS},r})\right]$, which is the objective in Eqn. (\ref{eq:affine}).

We use a UCI dataset (Waveform, binary classification) for illustration where the value of $r^*$ is approximately 0. When the noise rates are $0.1, 0.2, 0.3, 0.4$, the optimal smooth rate should be $-0.25, -0.67, -1.5, -4$ according to Eqn. (\ref{eq:thred}). The estimated noisy risk of LS/VL/NLS on these noise settings can be summarized in Table 1. Clearly, when $e=0.1$, $r=-0.25$ is closest to the estimated (clean) true risk (also returns the best test accuracy among these smooth rates). Similar observations hold for all other $e$. Learning with $r_{\text{opt}}$ on the noisy data yields the closest risk to the corresponding clean risk with $r^*$!

\begin{table}[!htb]
\centering
\caption{The difference between the empirical true risk of $Y^*$ on the clean data and empirical risk of LS/VL/NLS on noisy labels (UCI-Waveform data): $r^*$, empirical true risk, and empirical noisy risks of $r_{\text{opt}}$ under various noise levels are highlighted in purple.}
\scalebox{0.68}{
\begin{tabular}{c|c|c|c|c}
\hline 
Smooth rate & Risk (clean) & Risk ($e_i=0.1$) & Risk ($e_i=0.2$) & Risk ($e_i=0.3$) \\
\hline \hline 
 $r=0.8$ &   0.6773    &   0.6831   &  0.6873   &  0.6899      \\
\hline 
 $r=0.6$ &  0.6295  &   0.6521 &  0.6689  & 0.6833    \\
\hline 
 $r=0.4$& 0.5437&   0.5994 & 0.6408 &0.6718  \\
\hline 
 $r=0.2$& 0.4134  &  0.5212  &  0.5956 &  0.6550   \\
\hline
\cellcolor{blue!10} $\mathbf{r^*=0.0}$ &    \cellcolor{blue!10} $\textbf{0.1798}$ &  0.4057  &0.5399 & 0.6314 \\
\hline   $r=-0.25$&  -36.8095&   \cellcolor{blue!10}$\textbf{0.1983}$   & 0.4381  &    0.5957\\
\hline 
 $r=-0.67$ & -333.1283 &    -28.3508 & \cellcolor{blue!10} \textbf{0.2167} & 0.5132 \\
\hline
 $r=-1.5$ &  -97.4378   &    -61892.8047  &  -94.9509  &  \cellcolor{blue!10}$\textbf{0.1911}$     \\
\hline
\end{tabular}}
\end{table}

\subsection{What is the practical distribution of $r^*$ and $r_{\text{opt}}$?}\label{sec:r_star}
\paragraph{$r^*$ and $r_{\text{opt}}$ on UCI datasets \citep{Dua:2019}}
As for UCI datasets, we pick Twonorm and Splice for illustration. The noisy labels are generated by a symmetric noise transition matrix with noise rate $e_i=[0.1, 0.2, 0.3, 0.4]$. As highlighted in Table \ref{Tab:uci_paper1} (top of this page), $r_{\text{opt}}$ appears with positive values when the data is clean (same as $r^*$) or of a low noise rate. With the increasing of noise rates, the performance of LS results in a much larger degradation compared with NLS. We color-code different noise regimes where either VL/LS (red-ish) or NLS (green-ish) outperforms the other. Clearly there is a separation of the favored smoothing rate for different noise scenarios (upper left \& low noise for VL/LS, bottom right \& high noise for NLS).

\paragraph{$r^*$ and $r_{\text{opt}}$ on CIFAR datasets \citep{krizhevsky2009learning}}
When learning with a larger scale and more complex dataset, like CIFAR-10 and CIFAR-100, models are prone to converge on a local optimal solution rather than the global optimum. This phenomenon occurs frequently in NLS which ends up with performance degradation. Thus, in Table \ref{Tab:cifar10} and \ref{Tab:cifar100}, when learning with noisy labels, we report the better performance of LS and NLS between direct training and loading the same warm-up model. We observe that the performance of NLS is more competitive than LS when learning with clean data. Clearly, NLS outperforms LS in CIFAR-10 and CIFAR-100 under various synthetic noise settings. The gap is larger when the noise rates are high. The results of two independent sample T-test \footnote{4 non-negative smooth rates V.S. the smallest 4 negative smooth rates, $p$-value is highlighted in green if NLS generally returns a higher accuracy (i.e., $t$-value$<0$) than VL/LS, otherwise, in red.} further verify this conclusion. 
 \begin{table}[!htb]
		\caption{Test accuracy (mean$\pm$std) comparisons on symmetric noisy CIFAR-10 datasets. Best two smooth rates for each synthetic noise setting are highlighted for each $\epsilon$ (green: NLS; red: VL/LS).}
        \small
		\begin{center}
		\scalebox{0.83}{\begin{tabular}{c|cccc}
				\hline 
		\multirow{2}{*}{Smooth Rate}  &  \multicolumn{4}{c}{\emph{CIFAR-10 Symmetric}} \\ &$\varepsilon = 0.0$
				  &$\varepsilon = 0.2$ &$\varepsilon = 0.4$&$\varepsilon = 0.6$\\
				\hline\hline
			   $r=0.8$	& 92.91$\pm$0.06 & 88.88$\pm$1.61 & 81.48$\pm$2.91& 73.16$\pm$0.16 	\\
			   	 \hline 
			  $r=0.6$ &	92.33$\pm$0.09 & 87.50$\pm$1.31 & 	82.11$\pm$0.86& 73.59$\pm$0.15 \\ 
			   	 \hline 
			   $r=0.4$	& 93.05$\pm$0.04 & 87.13$\pm$0.07 & 81.50$\pm$1.42& 74.21$\pm$0.19\\
			   	 \hline 
			   $r=0.0$&	91.44$\pm$0.16 & 85.08$\pm$0.86  & 80.42$\pm$2.29& 75.34$\pm$0.13 \\
			   	 \hline \hline 
			  $r=-0.4$ 	& \good{93.55$\pm$0.06} & 87.55$\pm$0.08&  81.58$\pm$0.19& 75.95$\pm$0.13\\
			  \hline 
			  $r=-0.8$ &92.74$\pm$0.05	 &88.46$\pm$0.11 & 81.56$\pm$0.15 & 	76.15$\pm$0.14\\
			   \hline 
			  $r=-1.0$ 	&92.58$\pm$0.08 & 88.58$\pm$0.08 & 81.95$\pm$0.10 & 	76.20$\pm$0.10\\
			   \hline 
			  $r=-2.0$ 	& 	\good{93.30$\pm$0.03} & 	88.78$\pm$0.09& 	83.64$\pm$0.15 &  76.11$\pm$0.07\\
			   \hline 
			  $r=-4.0$ &	93.13$\pm$0.04 	  & 	\good{88.90$\pm$0.07}& 	\good{84.34$\pm$0.13} &  	\good{77.22$\pm$0.09}\\
			   \hline 
			  $r=-6.0$ 	&  	93.14$\pm$0.08& 	\good{88.94$\pm$0.11}&   	\good{84.52$\pm$0.13}&  	\good{77.42$\pm$0.16} \\
			   \hline \hline
			    $p\text{-value}=$ & \good{$0.0004$} & \good{0.008} & \good{0.011} & \good{$<1e-14$}\\
			   \hline
			\end{tabular}}
			\end{center}
		\label{Tab:cifar10}
	\end{table}

       \begin{table}[!htb]
		\caption{Test accuracy (mean$\pm$std) comparisons on asymmetric noisy CIFAR-10, symmetric CIFAR-100 datasets. Best two smooth rates for each synthetic noise setting are highlighted for each $\epsilon$ (green: NLS; red: VL/LS).}
        \small
		\begin{center}
		\scalebox{0.83}{\begin{tabular}{c|cc|cc}
				\hline 
		\multirow{2}{*}{Smooth Rate} &  \multicolumn{2}{c}{\emph{CIFAR-10 Asymmetric}} &  \multicolumn{2}{c}{\emph{CIFAR-100 Symmetric}} \\  &$\varepsilon = 0.2$ &$\varepsilon = 0.3$&$\varepsilon = 0.4$ &$\varepsilon = 0.6$\\
				\hline\hline
			   $r=0.8$	& \better{90.45$\pm$0.06} & 87.83$\pm$0.13&  54.04$\pm$0.93 &   39.50$\pm$0.18\\
			   	 \hline 
			  $r=0.6$ &	 90.41$\pm$0.09& 87.83$\pm$0.13 &  52.72$\pm$0.15 & 40.49$\pm$0.07\\ 
			   	 \hline 
			   $r=0.4$	 &\better{90.49$\pm$0.10}& 87.90$\pm$0.13&  54.26$\pm$0.07& 41.57$\pm$0.05 \\
			   	 \hline 
			   $r=0.0$& 88.32$\pm$0.24 & 86.27$\pm$0.32 & 48.03$\pm$0.29 &  38.11$\pm$0.14\\
			   	 \hline 
			   	 \hline
			  $r=-0.4$ 	&  87.27$\pm$1.83&\good{88.33$\pm$0.06}& 56.87$\pm$0.08&   43.70$\pm$0.16\\
			  \hline 
			  $r=-0.8$ &86.40$\pm$1.32 & \good{87.96$\pm$0.43} & 57.35$\pm$0.08 & 44.10$\pm$0.06\\
			   \hline 
			  $r=-1.0$ 	 &88.47$\pm$0.15 & 87.50$\pm$0.73 & 57.44$\pm$0.09 & 43.85$\pm$0.19\\
			   \hline 
			  $r=-2.0$ 	&88.66$\pm$0.17 & 87.27$\pm$0.70& \good{58.10$\pm$0.08}& 44.88$\pm$0.11\\
			   \hline 
			  $r=-4.0$ & 89.56$\pm$0.17& 87.29$\pm$0.59&  \good{58.35$\pm$0.09}& \good{46.38$\pm$0.05}\\
			   \hline 
			  $r=-6.0$ 	 &89.70$\pm$0.24 & 87.57$\pm$0.42& 57.73$\pm$0.10 & \good{46.46$\pm$0.09}\\
			   \hline 
			   \hline
			   $p\text{-value}=$ & \better{$<1e-7$} & \good{0.106} & \good{$<1e-14$} & \good{$<1e-15$}\\
			   \hline
			\end{tabular}}
			\end{center}
		\label{Tab:cifar100}
	\end{table}

\section{The Impacts on the Model Confidence}\label{sec:clean_gls}
Continuing the discussion of differed model confidence in the previous section, we now empirically explore how such differences distinguish LS and NLS.

\begin{figure}[!ht]
\centering
\hspace{-0.23in}
\includegraphics[width=0.17\textwidth]{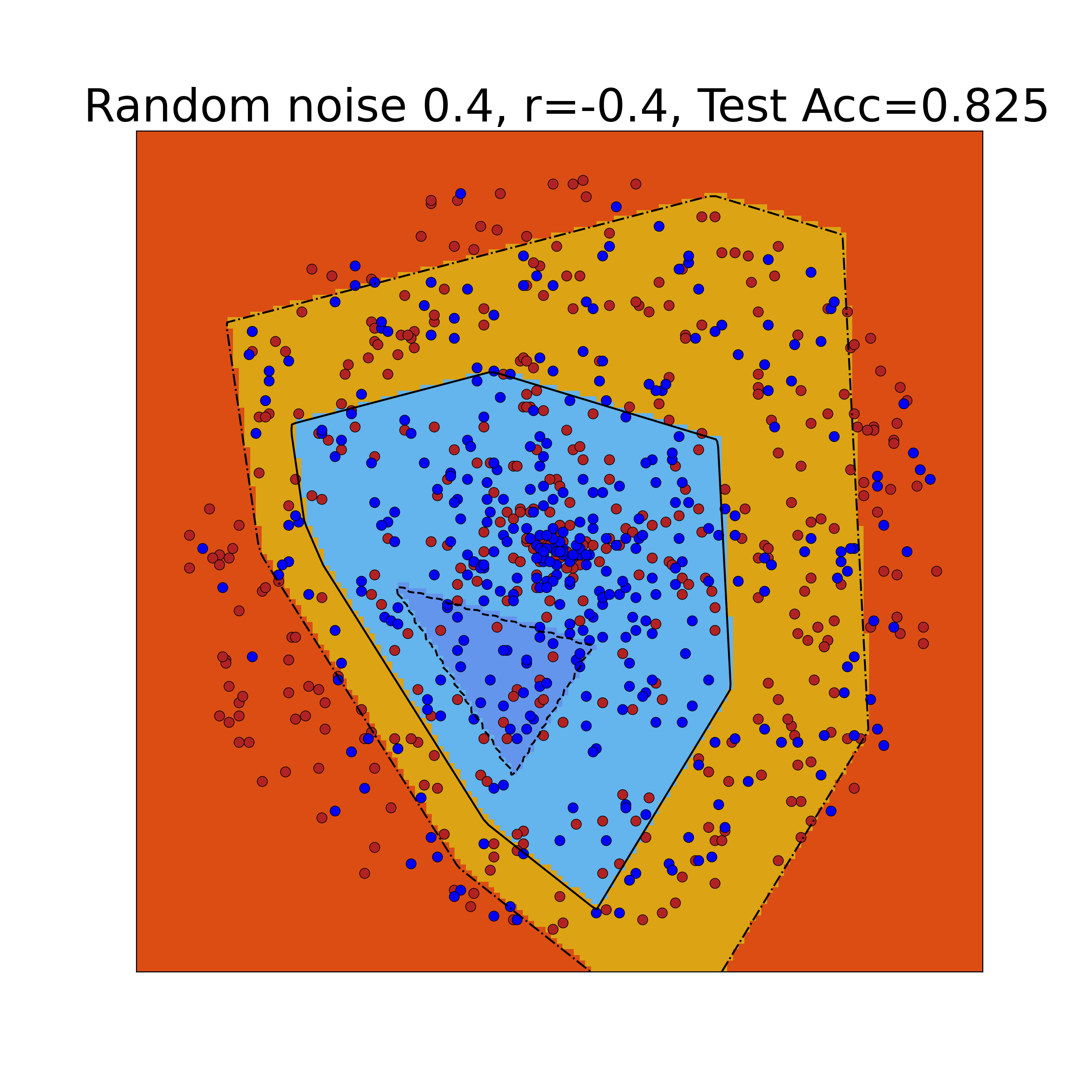}
\hspace{-0.1in}
\includegraphics[width=0.17\textwidth]{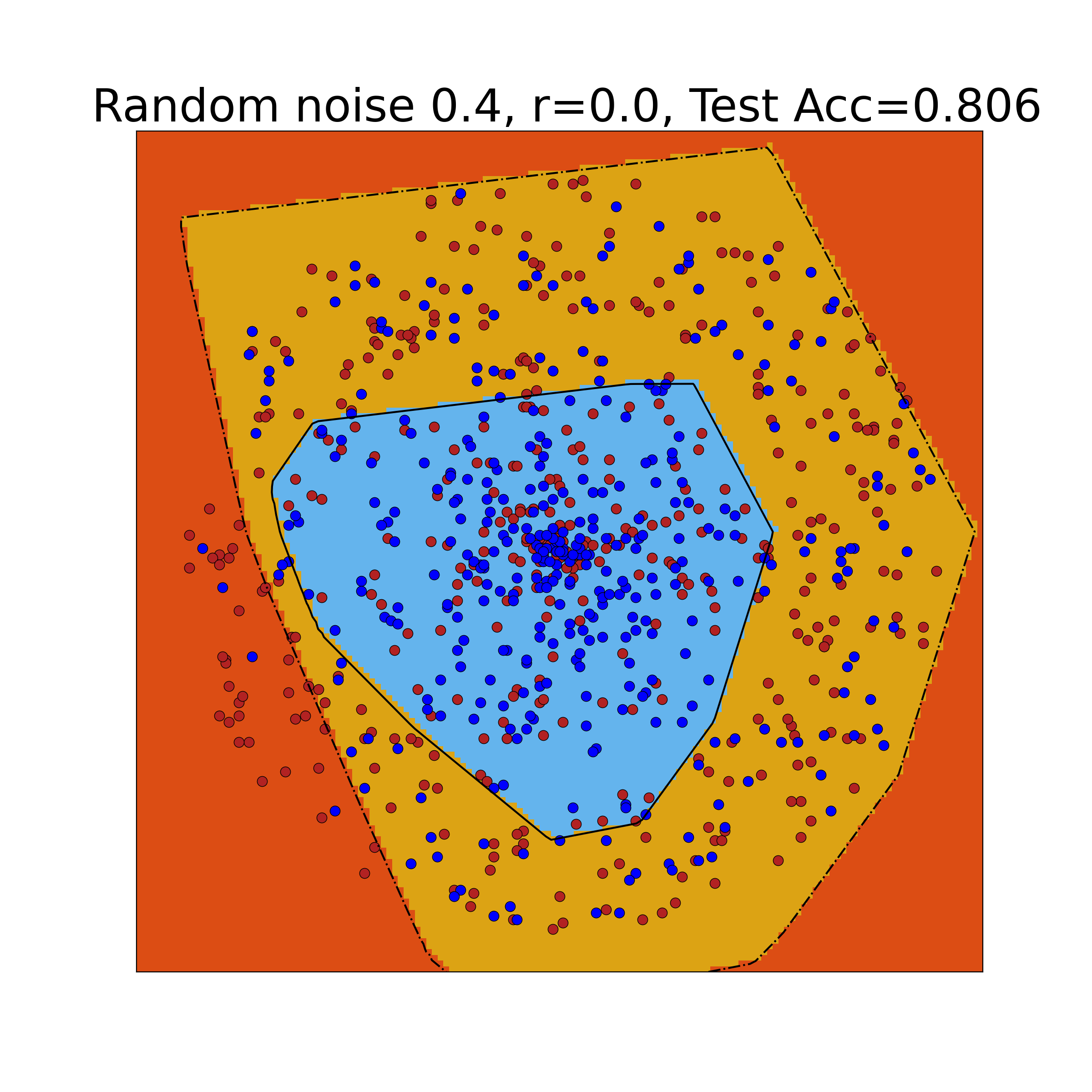}
\hspace{-0.12in}
\includegraphics[width=0.17\textwidth]{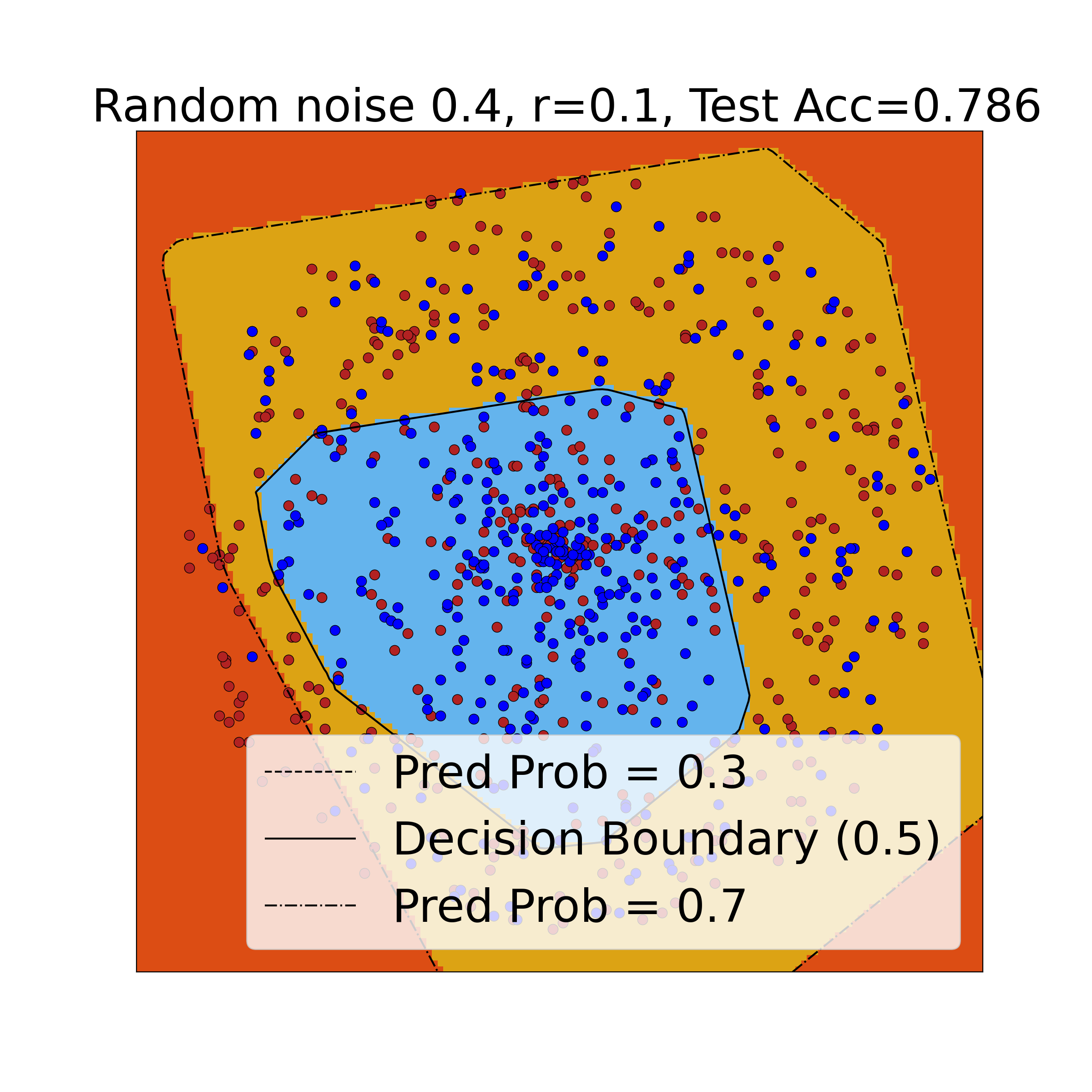}
\hspace{-0.25in}
\caption{Model confidence visualization of NLS, VL, and LS on synthetic data (Type 1) with noise rate $e_i=0.4$. The optimal smooth rate is $-0.4$. (left: NLS; middle: Vanilla Loss; right: LS). The test accuracy is annotated above each plot.}\label{fig:type1_3}
\end{figure}

Remember that when the label is clean ($e_0=e_1=0$), Eqn. (\ref{eqn:gen}) reduces to:
\begin{align}
\label{eqn:clean_gen}
    &\min_{f \in \mathcal F}~\mathbb{E}_{(X, Y)\sim \D}   \Big[ \ell(\bfX, Y)\Big] \notag\\
    &+\frac{ r}{2}\cdot\mathbb{E}_{(X, Y)\sim \D} \underbrace{ \Big[\ell(\bfX, 1-Y) - \ell(\bfX, Y)\Big]}_{\text{Term }\mc_{\ell}(f;X, Y)}.
\end{align}
{\color{black}
The difference between LS and NLS lie in the weight of Term $\mc_{\ell}(f;X, Y)$ when learning with clean labels: NLS encourages high $\mc_{\ell}(f;X, Y)$ and $\mc(f;X, Y)$ while LS has an opposite effect. 

\subsection{Side-effects of over-confident}
We adopt the generation of 2D (binary) synthetic dataset from \citep{amid2019robust} by randomly sampling two circularly distributed classes. The inner annulus indicates one class ({\color{blue}blue}), while the outer annulus denotes the other class ({\color{red}red}). We hold $20\%$ data samples for performance comparison.

\begin{figure}[!ht]
\centering
\vspace{-0.12in}
\hspace{-0.18in}
\includegraphics[width=0.18\textwidth]{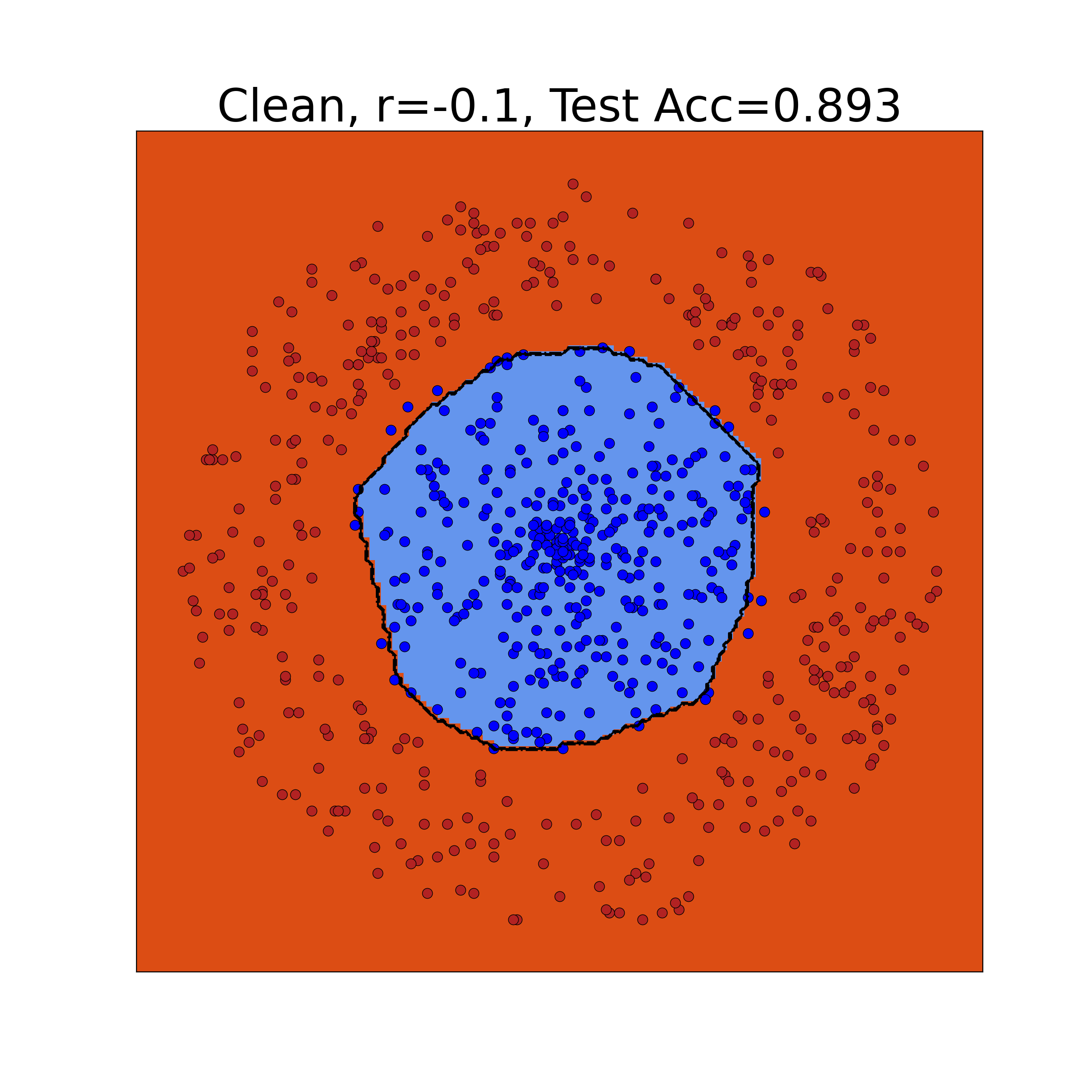}
\hspace{-0.18in}
\includegraphics[width=0.18\textwidth]{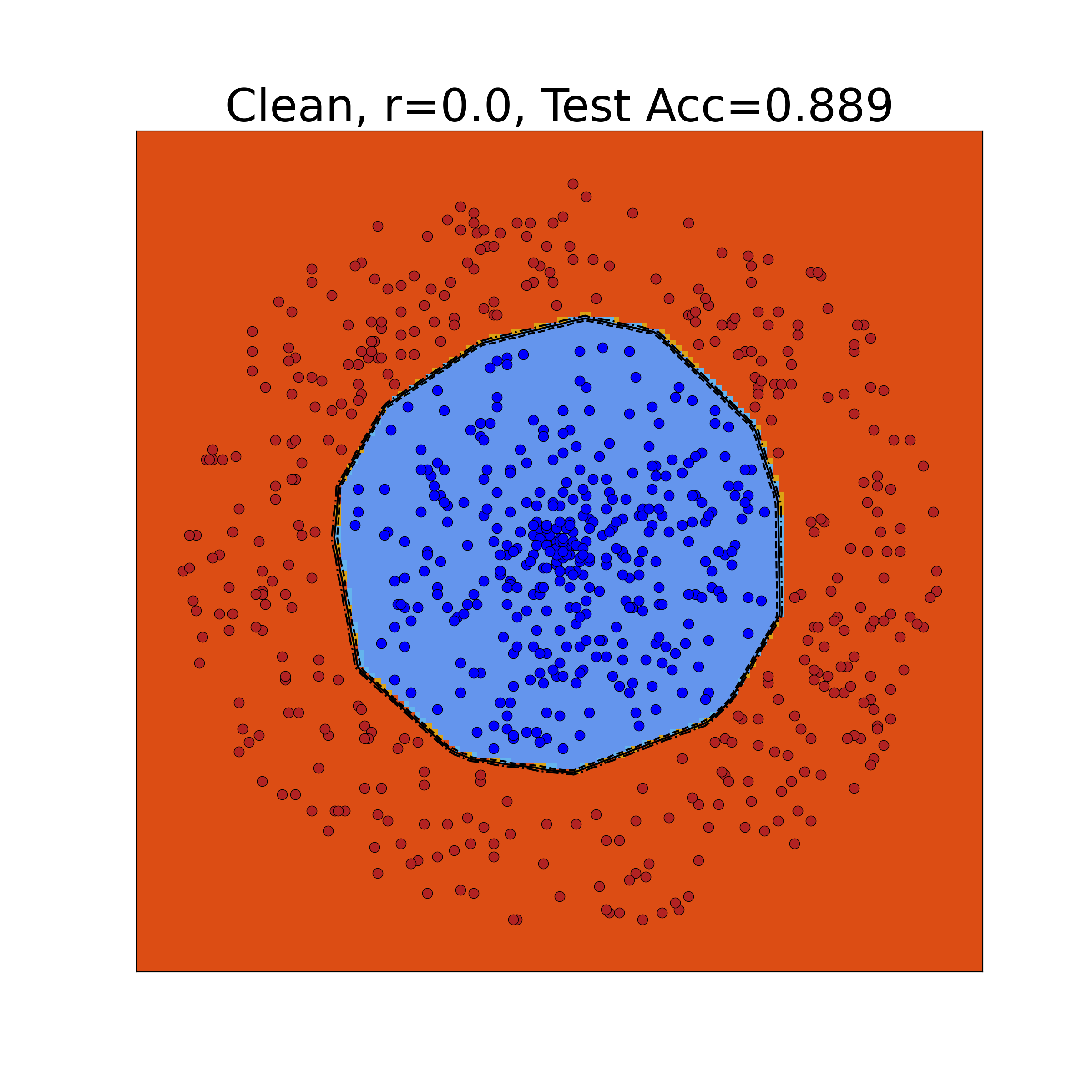}
\hspace{-0.18in}
\includegraphics[width=0.18\textwidth]{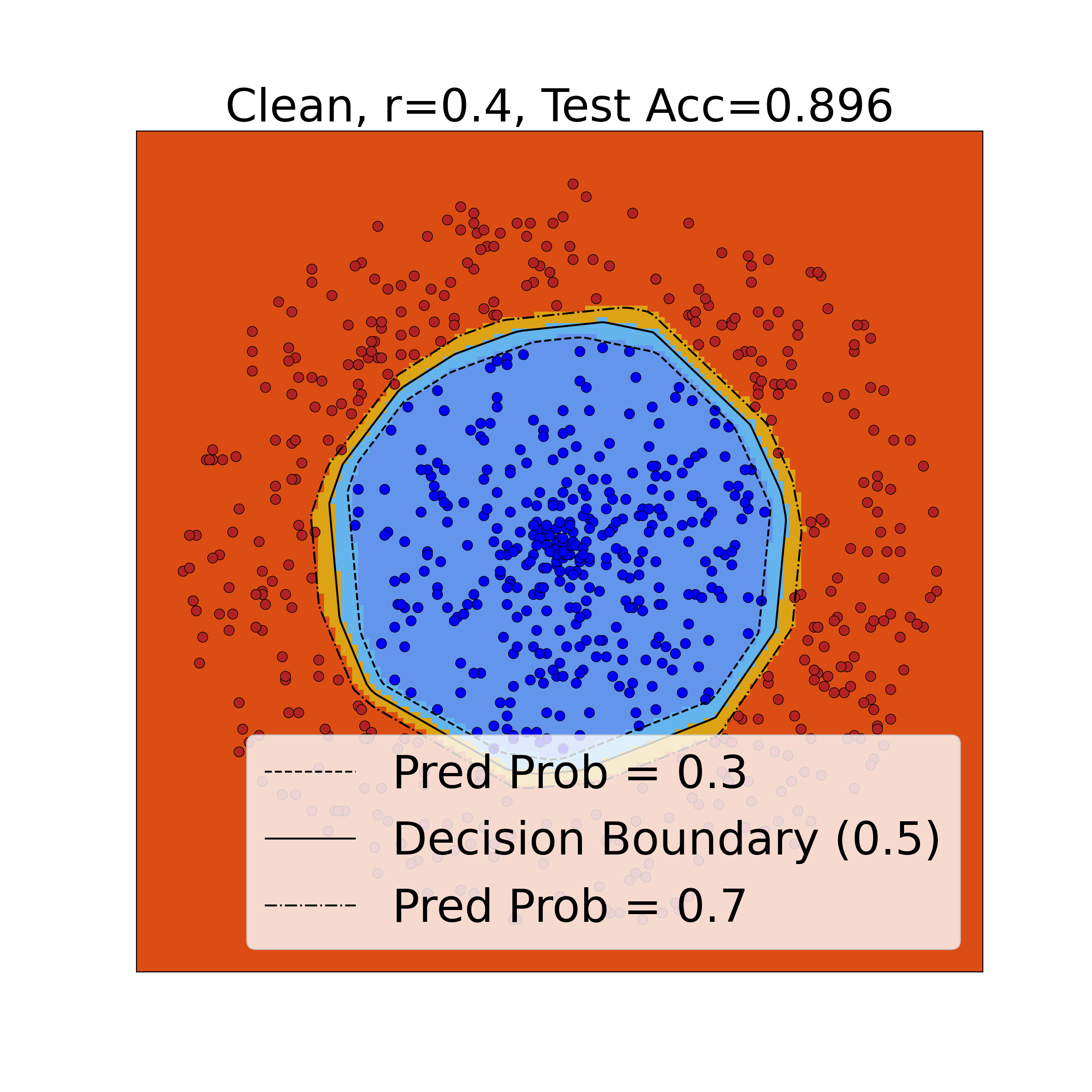}
\hspace{-0.18in}
\vspace{-0.1in}
\caption{Model confidence visualization of NLS, VL, and LS  on synthetic data (Type 1) with the clean data. The optimal smooth rate falls in $[0, 0.4]$. (left: NLS; middle: Vanilla Loss; right: LS). The test accuracy is annotated above each plot.}\label{fig:type1_1}
\vspace{-0.1in}
\end{figure}
In Figure \ref{fig:type1_1}, the colored bands depict the different levels of prediction probabilities: light blue + orange bands indicate samples that satisfy $\mc<0.4$ (low model confidence). When learning with the clean data, a non-positive smooth rate may yield over-confidence on the model prediction and a relatively low test accuracy.

\subsection{Label noise reduces model confidence} Recent works \citep{liu2021importance,cheng2021learning} have demonstrated that with the presence of label noise, learning with noisy labels directly will eventually result in unconfident model predictions. Continuing the synthetic 2D dataset, we flip the clean labels according to a symmetric noise transition matrix with noise rate $e_i$ for both classes. With the presence of label noise in Figure \ref{fig:type1_3}, the trained models generally become less confident on its predictions. Besides, when the smooth rate increases from negative to positive, more samples are of uncertain predictions. Thus, a smaller/negative smooth rate is beneficial when the noise rate increases by encouraging more confident predictions.}

\subsection{Model confidence on CIFAR-10 test dataset}
\begin{figure*}[!htb]
\centering
\includegraphics[width=0.24\textwidth]{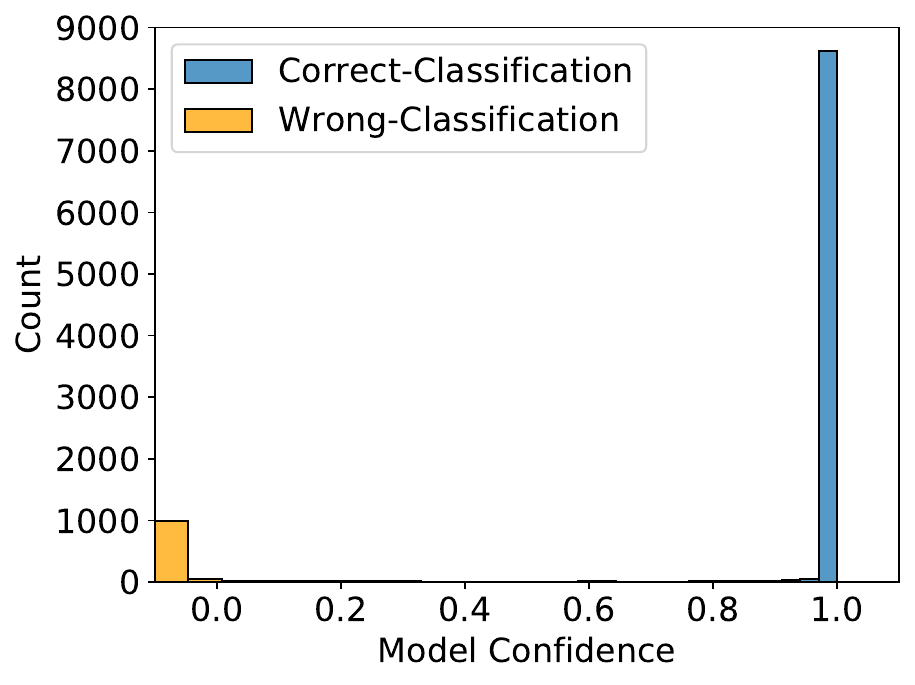}
\includegraphics[width=0.24\textwidth]{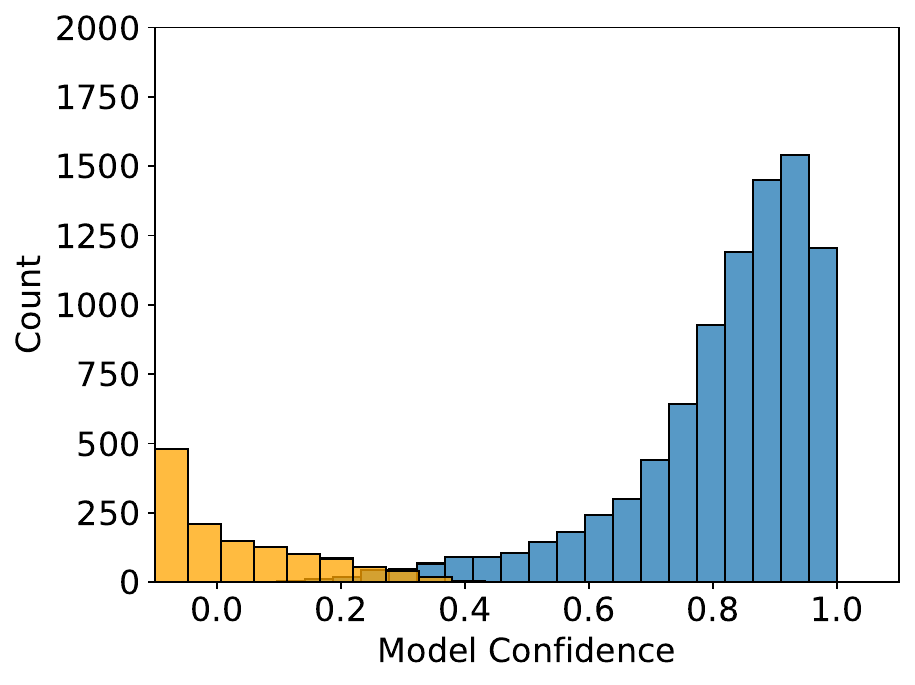}
\includegraphics[width=0.24\textwidth]{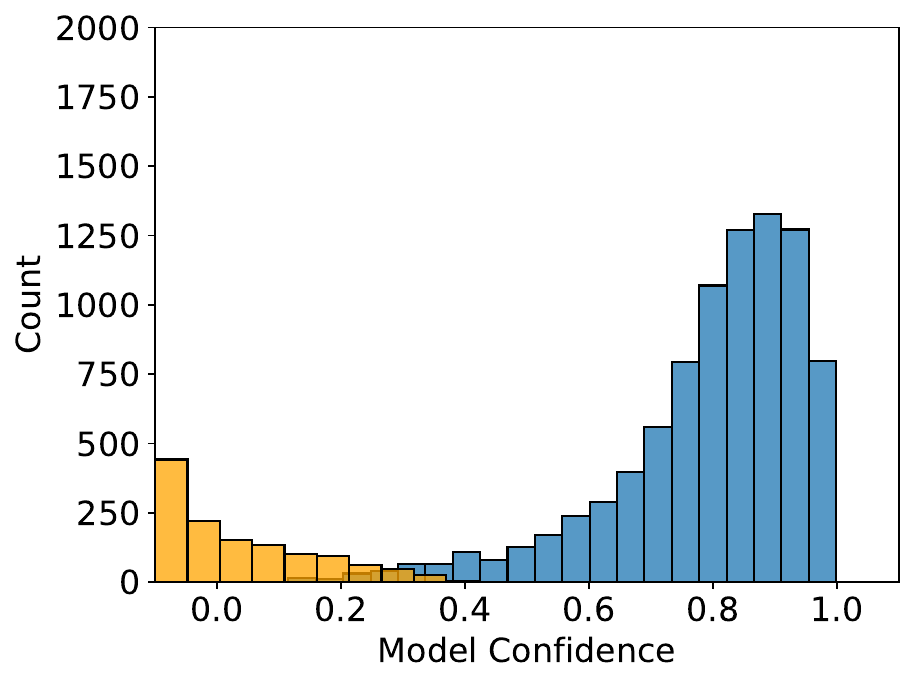}
\includegraphics[width=0.24\textwidth]{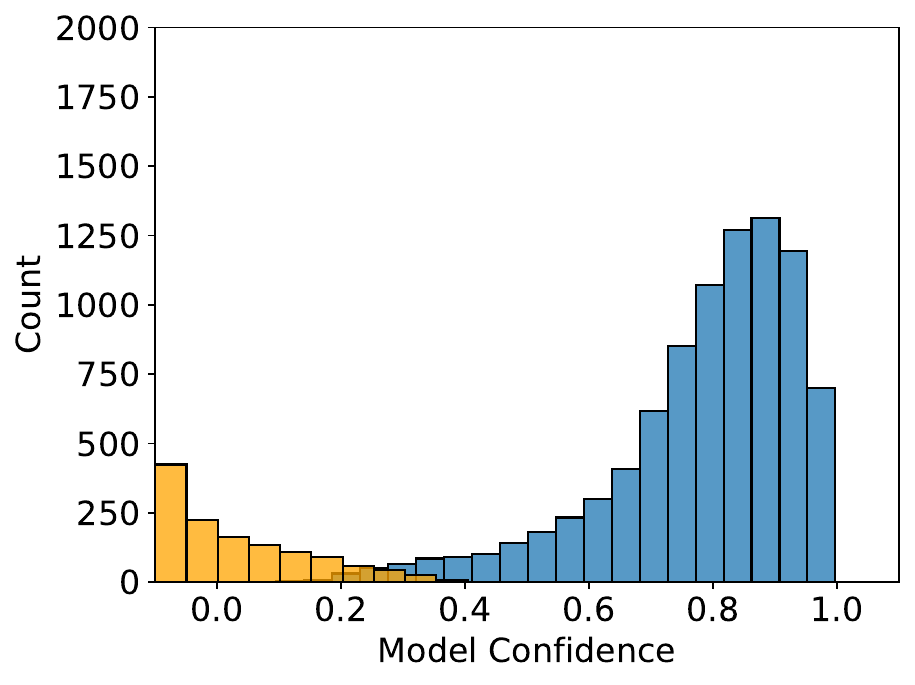}
\vspace{-0.05in}
\caption{Model confidence distribution of correct and wrong predictions on CIFAR-10 test data samples. (From left to right: NLS ($r=-0.8, -0.4$), Vanilla Loss, LS ($r=0.4$), trained on symmetric 0.2 noisy CIFAR-10 dataset).}\label{fig_mc_cifar_02}
\end{figure*}

When trained on symmetric 0.2 noisy CIFAR-10 training dataset (see Figure \ref{fig_mc_cifar_02}), with the decreasing of smooth rates (from right to left), the model confidence on correct predictions gradually approach to its maximum, while for wrong predictions, the model confidence converges to its minimum value. We observe that NLS makes the model prediction become over-confident on correct predictions and in-confident on wrong predictions.

\section{Connection to Other Robust Methods}\label{sec:connect}

In this section, we aim to theoretically explore the connection between NLS and popular robust methods such as backward/forward loss correction \citep{natarajan2013learning,patrini2017making}, NLNL \citep{kim2019nlnl} and peer loss \citep{liu2020peer}, under the unified setting. We defer the corresponding empirical verification to Appendix \ref{app:val_theory}.

\vspace{-0.06in}
\subsection{Loss correction}
Loss correction \citep{patrini2017making} studies two robust loss designs which are based on the knowledge of non-singular noise transition matrix $T$. The backward correction $\ell^{\leftarrow} (\bfX, \nY)$ re-weights the loss $\ell(\bfX, \nY)$ by $T^{-1}_{\hat{Y}, \nY}$ with $\hat{Y}$ being the model predicted label, while the proposed forward correction $\ell^{\rightarrow} (\bfX, \nY)$ multiplies the model predictions by $T$.

\begin{proposition}
\label{prop:bc_glsr}
For $r_{\text{LC}}:=\frac{2e_0}{2e_0-1}<0$, $\lambda_{\text{LC}}:=e_{\Delta} \cdot \frac{1}{1-2e_0}$, risk minimization of both backward and forward correction (with the knowledge of noise rates) are equivalent to the combination of NLS and an extra bias term Bias-LC
\vspace{-0.1in}
\begin{align*}
    &\min_{f \in \mathcal F} ~\E_{(X, \nY)\sim\nD} \Big[\ell^{\leftarrow} (\bfX, \nY)\Big]\\
    =&\min_{f \in \mathcal F} ~\E_{(X, \nY)\sim\nD} \Big[\ell^{\rightarrow} (\bfX, \nY)\Big]\\
    = &\min_{f \in \mathcal F} ~\E_{(X, \nY)\thicksim \nD} \Big[\ell(\bfX, \nY^{\text{GLS}, r_{\text{LC}}})\Big]\\
    &+\lambda_{\text{LC}}   \cdot \underbrace{ \mathbb{E}_{X, Y=1}  \Big[\ell(\bfX, 1) - \ell(\bfX, 0)\Big]}_{\text{Bias-LC}}.
\end{align*}
\end{proposition}
The incurred Bias-LC controls the model confidence on $(X,Y=1)\sim \D_{f}$. Note that when the noise rate is not substantially high, i.e., $e_0\in [0, \frac{1}{2})$, $\lambda_{\text{LC}}>0$. Then, compared with loss correction, NLS with smooth rate $r_{\text{LC}}$ makes the model $f$ to be less confident on $(X,Y=1)\sim \D_{f}^+$ and more confident on $(X,Y=1)\sim \D_{f}^-$ (wrong predictions). However, the impact of term Bias-LC is diminishing when either $e_{\Delta}\to 0$ (symmetric noise rates) or $e_0\to 0$ (low noise rates) as specified in Theorem \ref{thm:bc_glsr}.
\begin{theorem}
\label{thm:bc_glsr}
Assume the noise transition matrix is symmetric, i.e., $e_{\Delta}=0$, backward and forward loss correction are a special form of NLS with smooth rate $r_{\text{LC}}$. 
\end{theorem}

\begin{table*}[!htb]
	\vspace{-0.15in}
	\caption{{\color{black}Performance comparisons on synthetic noisy CIFAR datasets: we adopt the same model architecture for all methods (ResNet 34 \citep{he2016deep}), best achieved test accuracy is reported. }}
	\begin{center}
    \scalebox{0.7}{\begin{tabular}{c|ccc|cc|cc}
			\hline 
	\multirow{2}{*}{Method}  & \multicolumn{3}{c}{\emph{CIFAR-10, Symmetric}}  & \multicolumn{2}{c}{\emph{CIFAR-10, Asymmetric}} & \multicolumn{2}{c}{\emph{CIFAR-100, Symmetric}}\\ 
				  &$\varepsilon = 0.2$ &$\varepsilon = 0.4$&$\varepsilon = 0.6$  &$\varepsilon = 0.2$ &$\varepsilon = 0.3$ &$\varepsilon = 0.4$&$\varepsilon = 0.6$\\
				\hline\hline
				Cross Entropy & 86.45 & 82.72 & 74.04 & 88.59 & 86.14 & 48.20 & 38.27 \\
			   	Bootstrap \citep{reed2014training} & 86.06 & 81.65 & 75.26 & 87.69 & 85.51 & 47.28 & 35.81 \\
		FLC \citep{patrini2017making} & 84.85 & 84.98 & 73.97 &89.42 & 88.25 & 53.04 & 41.59\\	
SCE \citep{wang2019symmetric}  	 &  89.39 &  80.31 &   75.28&  88.07 & 85.93  &  49.34 & 38.87 \\
			APL \citep{ma2020normalized}& 88.42  & 81.27 &76.62 & 88.75 & 87.41 & 51.63 & 42.31\\Peer Loss \citep{liu2020peer} &  90.21 & 86.40  & 79.64  & 91.38 & 89.65 & 62.16 & 53.72\\
   	 	 ELR \citep{liu2020early}	 & 92.57 & 91.32 & 88.86 & 93.48 & 92.21 & 68.03 & 60.49\\
			 AUM \citep{pleiss2020identifying}  	 & 91.52 & 87.85 & 81.71 & 92.17  & 90.63 & 59.29 & 44.05\\\hline\hline
			  Label Smoothing (LS) \citep{lukasik2020does} & 90.24  & 83.78 & 75.01   & 90.61  & 88.04 & 55.17 &  41.63\\
			 Negative/Not Label Smoothing (NLS) & 89.05 & 84.85 & 77.82  & 90.02 & 88.42 & 58.47 &  46.58  \\
		   	\hline
	\end{tabular}}
			\end{center}\label{tab:robust}
	\end{table*}

\subsection{Learning from complementary labels}
Complementary labels \citep{ishida2017learning} were firstly introduced to mitigate the cost of collecting data. Rather than encouraging the model to fit directly on the target, learning from complementary labels trains the model to not fit on the complementary label which differs from the target. Later, an indirect training method ``Negative Learning'' (NL) \citep{kim2019nlnl} was proposed to reduce the risk of providing incorrect information with the presence of noisy labels and is robust to label noise in multi-class classification tasks. A more generic unbiased risk estimator of learning with complementary labels was proposed \citep{ishida2019complementary}, a popular case is: $\ell_{\text{CL}}(\bfX, \nY):=\ell(\bfX, \nY)-\ell(\bfX, 1-\nY)$.

\begin{theorem}
\label{thm:cl_glsr}
Learning from complementary labels with $\ell_{\text{CL}}$ is equivalent to NLS with smooth rate $r_{\text{CL}}\to -\infty$:
\begin{align*}
    &\min_{f \in \mathcal F} ~\E_{(X, \nY)\sim \nD} \Big[\ell_{\text{CL}}(\bfX,\nY)\Big]\\ =& \min_{f \in \mathcal F} ~\E_{(X, \nY)\thicksim \nD} [ \ell(\bfX, \nY^{\text{GLS}, r_{\text{CL}}\to -\infty})].
\end{align*}
\end{theorem}

\subsection{Peer loss functions}

Peer loss functions \citep{liu2020peer} proposed a family of robust loss measures which do not require the knowledge of noise rates. The mathematical representation of peer loss functions is $\ell_{\text{PL}}(\bfX, \nY):=\ell(\bfX, \nY)  - \ell(\bfXo, \widetilde{Y}_2)$,
where $(X_i, \nY_i)\sim \nD$. The second term of the peer loss evaluates on randomly paired data samples and labels ($X_1$ and $\widetilde{Y}_2$ for two randomly selected samples) to punish $f$ from overly fitting on noisy labels.
    
\begin{proposition}
\label{prop:pl_glsr}
For $r_{\text{PL}}:=2\cdot \p(\nY=1)$, $\lambda_{PL}:=1-r_{\text{PL}}$, risk minimization of peer loss is equivalent to negative label smoothing regularization with an extra term Bias-PL, i.e.,
\begin{align*}
    &\min_{f \in \mathcal F} ~\E_{(X, \nY)\sim \nD} \Big[\ell_{\text{PL}}(\bfX,\nY)\Big]\\= &\min_{f \in \mathcal F} ~\E_{(X, \nY)\thicksim \nD} \Big[\ell(\bfX, \nY) - \ell(\bfX, \nY^{\text{GLS}, r_{\text{PL}}})\Big]\\
    &+\lambda_{PL} \cdot \underbrace{\E_{X,\nY=1} \Big[ \ell(\bfX, 1) - \ell(\bfX, 0)\Big]}_{\text{Bias-PL}}.
\end{align*}
\end{proposition}
The incurred term Bias-PL controls the model confidence on $(X, \nY=1)\sim \nD$ and has a diminishing effect as $\p(\nY=1)\to 1/2$. Generally, the peer loss relates to the unified setting (GLS) as the negatively weighted GLS term appears to be a regularizer. Note that we have access to the $\p(\nY=1)$, we can bridge the gap by adding an estimable term \text{Bias-PL}. With some derivations, we further show in Theorem \ref{thm:pl_glsr}, when noisy priors are equal, the peer loss has an exact NLS form.
\begin{theorem}
\label{thm:pl_glsr}
When the noisy labels have equal prior, i.e., $\p(\nY=0)=\p(\nY=1)$, the peer loss is a special form of NLS regularization with the smooth rate $r_{\text{PL}}$. Besides, 
\begin{align*}
    &\min_{f \in \mathcal F} ~\E_{(X, \nY)\sim \nD} \Big[\ell_{\text{PL}}(\bfX,\nY)\Big]
    \\=&\min_{f \in \mathcal F} ~\E_{(X, \nY)\thicksim \nD} \Big[ \ell(\bfX, \nY^{\text{GLS}, r\to -\infty})\Big].
\end{align*}
\end{theorem}

{\color{black}\subsection{Practical significance}\label{App:exp_more}}
In Table \ref{tab:robust}, we compare VL(CE), LS and NLS with several robust methods in synthetic noisy CIFAR datasets. Clearly, LS and NLS can be viewed as competitive and efficient robust loss functions which outperform Cross Entropy, Bootsrap \citep{reed2014training}, SCE \citep{wang2019symmetric}, APL \citep{ma2020normalized} and Forward loss correction (FLC) \citep{patrini2017making} in most settings. 

We also provide experimental results of LS and NLS on real-world human noise benchmarks: CIFAR-N \citep{wei2022learning} and Clothing 1M \citep{xiao2015learning}, along with several baseline methods for comparisons, i.e., backward loss correction (BLC) \citep{natarajan2013learning,patrini2017making}, forward loss correction (FLC) \citep{patrini2017making}, Peer Loss (PL) \citep{liu2020peer}, and F-div \citep{wei2021when}. Table \ref{exp:real} demonstrates the effectiveness of NLS. Besides, we observe that NLS ranks 4-th among 21 existing robust methods on Clothing 1M (no extra train data, evaluated on the clean test data)\footnote{Public leaderboard of CIFAR-N, Clothing 1M: \url{http://noisylabels.com/}, \url{https://paperswithcode.com/sota/image-classification-on-clothing1m}}. This simple trick clearly reveals the importance and great potential of NLS. Nonetheless we would like to clarify that our main purposes are (instead of chasing SOTA): (1) Provide new understandings 
of whether we should smooth the label or not when learning with noisy labels. (2) Reveal the importance and effectiveness of NLS at different scenarios. The popularity of label smoothing is largely due to its simplicity and being complementary, so we expect our observations for NLS can be combined with other SOTA methods to further improve model performance in the high-noise regime. 

\begin{table}[!htb]
\centering
\caption{Performance comparisons on Clothing 1M and CIFAR-N: results of baselines are obtained through the public leader-board.}
\scalebox{0.63}{
\begin{tabular}{c|c|cccc}
\hline 
\shortstack{\textbf{Method} \\\textbf{}} &\shortstack{\textbf{Clothing} \\\textbf{1M}} & \shortstack{\textbf{CIFAR-10N} \\\textbf{Aggre}} & \shortstack{\textbf{CIFAR-10N} \\\textbf{Rand1}} & \shortstack{\textbf{CIFAR-10N} \\\textbf{Worse}}& \shortstack{\textbf{CIFAR-100N} \\\textbf{Fine}}  \\
\hline \hline 
 CE &  68.94 & 87.77  & 85.02 & 77.69  &  55.50\\
BLC & 69.13  & 88.13  & 87.14 &  77.61 &  57.14\\
 FLC & 69.84  & 88.24  & 86.88 & 79.79  & 57.01 \\
PL & 72.60  &  90.75 & 89.06 &  82.53 &  57.59\\
F-div &  73.09 &  91.64 & 89.70 &  82.53 & 57.10 \\
\hline \hline
 LS (best)&  73.44   &  91.57    & 89.80  &   82.76  &  55.84  \\
 NLS (best)&   \textbf{74.24}   &   \textbf{91.97}    &	  \textbf{90.29} &   \textbf{82.99}   &   \textbf{58.59}  \\
\hline
\end{tabular}}\label{exp:real}
\vspace{-0.1in}
\end{table}

\section{Conclusion}
In this paper, we provide understandings for whether should we adopt label smoothing or not when learning with noisy labels. We show that learning with negatively smoothed labels explicitly improves the confidence of model prediction. This key property acts as a significant role when the confidence of model prediction drops. In contrast to existing works that promote the use of positive label smoothing, we show both theoretically and empirically the advantage of a negative smooth rate when the label noise rate increases. We also bridge the gap between negative label smoothing and existing learning with noisy label solutions, which further demonstrates the importance of negative/not label smoothing. In a nutshell, our observations provide new understanding for the effects of label smoothing, especially when the training labels are imperfect. Future works include exploring the benefits of negative labels in other tasks.

\paragraph{Acknowledgement}
YL and JHW are partially supported by
the grants IIS-2007951 and IIS-2143895. TLL is partially supported by Australian Research Council Projects DE-190101473, IC-190100031, and DP-220102121. MS and GN are supported by JST CREST Grant Number JPMJCR18A2. The authors thank anonymous ICML reviewers for their comments that improved
the presentation.
\newpage
\bibliography{src/ref}

\bibliographystyle{icml2022}

\newpage
\appendix
\onecolumn
\newpage
\appendix

\begin{center}
    \section*{\Large Appendix}
\end{center}
The Appendix is organized as follows. 
\squishlist
\item Section A presents the full version of related works.
\item Section B includes empirical validations of theoretical conclusions in Section \ref{sec:connect}. 
\item Section C discusses practical considerations of the robustness for LS and NLS.
\item Section D shows additional experiments on synthetic dataset and UCI datasets. 
\item Section E illustrates the bias and variance trade-off when learning with LS and NLS from clean data.
\item Section F includes omitted proofs for theoretical conclusions in the main paper. 
\squishend

\section{Full Version of Related Works}
\label{app:full_related}
Our work supplements to two lines of related works.
\paragraph{Learning with noisy labels}
Annotated labels from human labelers usually consists of an non-negligible amount of mis-labeled data samples. Making deep neural nets perform robust training on ``noisily" labeled datasets remains a challenge. Classical approaches of learning with noisy labels assume the noisy labels are independent to features. They firstly estimate the noise transition matrix \citep{liu2015classification, menon2015learning, Scott_kernel_embedding, patrini2017making,zhu2021good,zhu2021clusterability,yang2021estimating,cheng2022instance,zhu2022beyond}, then proceed with a loss correction \citep{natarajan2013learning, patrini2017making,liu2015classification} to mitigate label noise. Recent works mainly focus on: (1) proposing robust loss functions \citep{kim2019nlnl,liu2020peer,wei2021when,englesson2021generalized,englesson2021consistency} to train deep neural nets directly without the knowledge of noise rates, or design a pipeline which dynamically select and train on ``clean" samples with small loss \citep{jiang2018mentornet,han2018co,yu2019does,yao2020searching,xia2021sample}; (2) hindering the memorization on noisy labels \cite{xia2020robust,liu2020early,cheng2021demystifying,liu2021convolutional,wei2021open,bai2021understanding,liu2022robust,yi2022learning}; (3) sample-level re-weighting to mitigate the impacts of wrong labels \cite{liu2016classification,majidi2021exponentiated,kumar2021constrained,liu2021can}. More recently, several approaches target at addressing more challenging noise settings, such as group/instance-dependent label noise \citep{cheng2021learning,wang2021fair,berthon2021confidence,zhu2021second,dawson2021rethinking,jiang2022an}, or considering more practical applications such as open-set data \cite{xia2020extended,wei2021open,xia2021instance}, partial label learning \cite{feng2020provably,lv2021robustness,wang2022pico}, samples with multiple noisy annotations \cite{wei2022deep,wei2022aggregate}.

\paragraph{Understanding the effect of label smoothing}
Learning with one-hot labels is prone to over-fitting,  soft label learning then naturally draws attentions of machine learning researchers. Successful applications of soft label learning include the label distribution learning \citep{geng2016label} which provides an instance with description degrees of all the labels. Label smoothing (LS) \citep{szegedy2016rethinking} is another arising learning paradigm that uses positively weighted average of both the hard training labels and uniformly distributed soft labels.
Empirical studies have demonstrated the effectiveness of LS in improving the model performance  \citep{pereyra2017regularizing,szegedy2016rethinking,vaswani2017attention,chorowski2017towards} and model calibration \citep{muller2019does}. However, knowledge distilling a teacher network (trained on smoothed labels) into a student network is much
less effective \citep{muller2019does}. Later, generalization effects of more advanced forms of label smoothing was studied, such as structural label smoothing \citep{li2020regularization}. More recently, it was shown that an appropriate label smoothing regularizer with reduced label variance boosts the convergence \citep{xu2020towards}. When label noise presents, \citep{liu2021importance} gives theoretical justifications for the memorizing effects of label smoothing. And the effectiveness of label smoothing in mitigating label noise is investigated in \citep{lukasik2020does}.

\section{Empirical Validations of Main Theorems}
\label{app:val_theory}
In this section, we empirically validate our main theoretical conclusions in Section \ref{sec:connect}, i.e, the connection between LS/NLS and popular methods. 

We compare the unified setting (GLS) with backward correction \citep{natarajan2013learning}, forward correction \citep{patrini2017making} and peer loss \citep{liu2020peer} on CIFAR-10 dataset. To approximate the performance of backward/forward Loss Correction, we adopt GLS with smooth rate $\frac{\epsilon}{(\epsilon-1)}$. As for the approximation of peer loss, we choose $\ell(\bfX,\nY)-\ell(\bfX, \nY^{\text{GLS}, r=0.5})$ which is equivalent to NLS when $r\to-\infty$. Experiment results in Table \ref{Tab:cifar10_connect} on CIFAR-10 under symmetric noise settings demonstrate that the equivalent forms of GLS are robust to label noise.
\vspace{-15pt}
\begin{table*}[!htb]
\caption{Comparison of test accuracies on CIFAR-10 under symmetric label noise.}\label{Tab:cifar10_connect}
\centering
{\scalebox{0.66}{
\begin{tabular}{c|ccc}
				\hline 
		\multirow{2}{*}{Method}  &  \multicolumn{3}{c}{\emph{CIFAR-10, Symmetric}}  \\ 
				  &$\varepsilon = 0.2$ &$\varepsilon = 0.4$&$\varepsilon = 0.6$   \\
				\hline\hline
			   	Backward $T$ \citep{patrini2017making} & 84.79 & 83.40 & 71.52  \\
			   	Forward $T$ \citep{patrini2017making} & 84.85 & \textbf{83.98} &73.97 \\
			   	 GLS form&  \textbf{87.33 }&  81.73 & \textbf{75.80}\\
			   	\hline\hline
			   	Peer Loss \citep{liu2020peer}  & \textbf{90.21}  & \textbf{86.40} & \textbf{79.64} \\
			   	GLS form    & 88.98 & 85.05  & 76.66   \\
			   	\hline
			\end{tabular}}}
			\vspace{-20pt}
\end{table*} 
\paragraph{Explanation of the performance gap} In practice, we adopt the same hyper-parameter setting as used for all other smooth rates for GLS form (VL, LS and NLS). Loss corrections will firstly warm-up with the cross-entropy loss, estimate the noise transition matrix with this pre-trained model, and then proceed to train with the backward/forward corrected loss. Peer loss functions adopt a dynamical adjustment for learning rate. The warming up, estimation error of noise transition matrix as well as the special hyper-parameter settings explain performance gaps.

\section{Practical Consideration of LS and NLS}\label{sec:prac_cons}
In the main paper, we theoretically show when we should adopt NLS and LS. In this section, we discuss more practical considerations, including the optimal smoothing parameter, how to reduce the impacts of bias terms, and multi-class extensions.

\subsection{The optimal smoothing parameter} 
In practice, we don't have access to noise rates $e_i$.

Our work does not intend to particularly focus on the noise rate estimation. For readers interested in the noise rate estimation, please refer to \citep{liu2015classification, menon2015learning, Scott_kernel_embedding, patrini2017making,yao2020dual,zhu2021clusterability}. To estimate $r_{\text{opt}}=\frac{r^*-2e}{1-2e}$, one can simply assume $r^*\to 0$. And the noise rate $e$ is estimable by a large family of noise estimation methods mentioned above. Our practical observations show that NLS with a CE warm-up is not sensitive to the negative smooth rate, for example, on CIFAR-10 and CIFAR-100 synthetic noisy datasets, $r<-1.0$ frequently achieves best results (see Table \ref{Tab:cifar10} in the main paper). Our current contribution focuses on understanding the generalized label smoothing, and we prefer leaving the task of identifying the optimal smooth rate to future works. 

\subsection{Making LS and NLS more robust to label noise}
There is a line of related works targeting at distinguishing clean labels from the noisy labels. Current literature in selecting clean samples from noisily labeled dataset is based on the empirical evidence that samples with noisy/wrong labels have a larger loss than clean ones. For interested readers, please refer to \citep{han2018co,jiang2018mentornet,yu2019does,yao2020searching,wei2020combating,northcutt2021confident}. Compared with the risk minimization over the clean data distribution $(X, Y)\sim \mathcal{D}$, learning directly with GLS on the noisy distribution $(X, \widetilde{Y})\sim \widetilde{\mathcal{D}}$ will result in an extra term $(e_1-e_0)\cdot (1-r)\cdot \mathbb{E}_{(X, Y=1)\sim \mathcal{D}} [\ell(\mathbf{f(X)}, 0)-\ell(\mathbf{f(X)}, 1)]$ compared to the clean scenario. Empirically, we can estimate the bias term, perform a bias correction by subtracting the estimated bias term from the objective function in Eqn. (\ref{eqn:gen}). 

Suppose we have access to a clean distribution $\D_{\text{clean}}$ which consists of selected clean samples. Denote the estimated noise rates as $\hat{e}_i$, when $e_{\Delta}\neq 0$, in order to make LS/NLS be more robust to label noise and fit on the optimal distribution $Y^*$, we improve by performing a model confidence correction on the dominating class through:
\begin{mybox3}
\begin{align*}
\label{eqn:gen-c}
    \min_{f \in \mathcal F} ~~&\mathbb{E}_{(X, \nY)\thicksim \nD}\quad \Big[\ell\big(\bfX, \nY^{\text{GLS},r}\big)\Big]\notag\\
    -&(\hat{e}_1-\hat{e}_0)\cdot (1- r) \cdot \mathbb{E}_{(X, Y=1)\thicksim \D_{\text{clean}}} \underbrace{ \Big[\ell\big(\bfX, 0\big) - \ell\big(\bfX, 1\big)\Big]}_{\text{confidence correction}}.
\end{align*}
\end{mybox3}

\section{Additional Experiment Results and Details}\label{app:more_exp}
In this section, we include more experiment results, observations and details for learning with LS/NLS.

\subsection{Experiment details on CIFAR-10, CIFAR-100}
We firstly introduce experiment details on CIFAR-10 dataset adopted in our experiment designs.

\paragraph{Training settings of clean CIFAR-10 dataset \citep{krizhevsky2009learning}}
We adopted ResNet34 \citep{he2016deep}, trained for 200 epochs with batch-size 128, SGD \citep{robbins1951stochastic} optimizer with Nesterov momentum of 0.9 and weight decay 1e-4. The learning rate of first 100 epochs is 0.1. Then it multiples with 0.1 for every 50 epochs.

\paragraph{Generating noise labels on CIFAR datasets}
We adopt symmetric noise model which generates noisy labels by randomly flipping the clean label to the other possible classes with
probability $\epsilon$. And we set $\epsilon=0.2, 0.4, 0.6$ for CIFAR-10, $\epsilon=0.4, 0.6$ for CIFAR-100. We also make use of asymmetric noise model. The asymmetric noise is generated by flipping the true label to the next class with probability $\epsilon$. We set $\epsilon=0.2, 0.3$ for CIFAR-10.

\paragraph{Training settings of synthetic noisy CIFAR datasets}
The generation of symmetric noisy dataset is adopted from \citep{cheng2021learning}. The symmetric noise rates are $[0.2, 0.4, 0.6]$. We choose two methods to train LS and NLS.
\squishlist
\item \textbf{Direct training:} this setting is the same as training on clean CIFAR-10 dataset.
\item \textbf{Warm-up:} in this case, we firstly train a ResNet34 model with Cross-Entropy loss for 120 epochs. For this warm-up, the only difference in hyper-parameter setting is the learning rate, where the initial learning rate is 0.1 and it multiplies 0.1 for every 40 epochs. After the warm-up, LS/NLS loads the same pre-trained model and trains for 100 epochs with learning rate 1e-6.
\squishend

\subsection{Why NLS is overlooked?}

When learning from a relative large scale dataset, NLS tends to push the model become overly confident early in the training. The poor performances of NLS (direct-train) in Table \ref{table:gap} explain why NLS is neglected. When there is no warm-up, training NLS directly without warming up will reach a $88\%-92\%$ test accuracy on the clean data. The performance will degrade much more significantly than LS when the noise level is high or $|r|$ is large. In Table \ref{table:gap}, we provide the comparisons between direct-train and warm-up in several settings. The improvement bring by a warm-up procedure becomes much more significantly in the high noise regime. NLS makes the classifier be overly confident at the early training which results in converging to a bad local optimum (without CE warm-up, NLS frequently results in a worse performance in CIFAR-10 and CIFAR-100). Since the model will usually fit on the clean data first, then over-fits on the noisy ones \citep{liu2020early}, a large number of approaches (such as Loss corrections \citep{patrini2017making}, Peer Loss \citep{liu2020peer}, etc) adopt a CE warm-up firstly. Note that there is no difference in the computing costs between NLS (with CE warmup) and CE loss, proceeding with NLS to enhance the model confidence makes NLS much more competitive in the high noise regime, also gives practical insights on how to make NLS work better when learning with clean data.

\begin{table*}[!htb]
\vspace{-0.1in}
		\caption{Test accuracies of GLS on assymetric noisy CIFAR-10 and symmetric CIFAR-100 (left/right denotes direct train / warm-up).}
		\begin{center}{\small\scalebox{0.85}{\begin{tabular}{c|cc|cc}
				\hline 
		\multirow{2}{*}{Smooth Rate}   &  \multicolumn{2}{c}{\emph{CIFAR-10 Asymmetric}} &  \multicolumn{2}{c}{\emph{CIFAR-100 Symmetric}}\\&$\varepsilon = 0.2$ &$\varepsilon = 0.3$&$\varepsilon = 0.4$ &$\varepsilon = 0.6$\\
				\hline\hline
			   $r=0.8$	& 87.89 / 90.51  & 86.38 / 87.97&  54.78 / 51.27 &   40.21 / 39.80\\
			   	 \hline 
			  $r=0.6$  & 89.14 / 90.55& 85.97 / 88.01 &  52.83 / 52.88 & 39.64  / 40.57\\ 
			   	 \hline 
			   $r=0.4$	 & 88.23 / \textbf{90.61}& 86.95 / 88.04&  51.40 / 54.36&   38.29 / 41.63 \\
			   	 \hline 
			  $r=-0.4$  & 19.71 / 89.60& 21.86 / \textbf{88.42}& 40.30  / 56.97&  31.35 / 43.91\\
			  \hline 
			  $r=-0.8$	 & - / 89.02 &- / 88.28 & 22.63 / 57.45 & 26.75 / 44.19\\
			   \hline 
			  $r=-1.0$  &- / 88.68 & - / 88.29 & - / 57.53 & - / 44.59\\
			   \hline 
			  $r=-2.0$ 	&- / 88.86 & - / 88.13& - / 58.21& - / 45.47\\
			   \hline 
			  $r=-4.0$ 	&- / 89.80& - / 88.20& - / \textbf{58.47}& - / 46.86\\
			   \hline 
			  $r=-6.0$ 	 &- / 90.02 & - / 88.18&- / 57.87 &- / \textbf{47.18} \\
			   \hline 
			\end{tabular}}}\label{table:gap}
			\end{center}\vspace{-10pt}
	\end{table*}

\subsection{Experiment details on synthetic datasets and UCI}
We introduce experiment details on synthetic  datasets and UCI datasets adopted in our experiment designs.
\paragraph{Generation of synthetic dataset} In the synthetic (Type 1) dataset, we generate 500 points for both classes. Class +1 distributes inside the circle with radius 0.25. Class -1 generates by randomly sampling 500 data points in the annulus with inner radius 0.28 and outer radius 0.45. As for synthetic (Type 2) dataset, we uniformly assign labels for 50\% samples in the annulus (with inner radius 0.22, outer radius 0.31) based on Type 1 dataset.

\vspace{-0.1in}
\paragraph{Generating noisy labels on synthetic datasets and UCI datasets}
 Note that these datasets are all binary classification datasets, each label in the training and validation set is flipped to the other class with probability $e$, and we set $e=0.1, 0.4$ for synthetic Type 1 dataset, $e=0.1, 0.3$ for synthetic Type 2 dataset.

\vspace{-0.1in}
\paragraph{Training settings of synthetic datasets}
For both types of synthetic datasets, we adopted a three-layer ReLU Multi-Layer Perceptron (MLP), trained for 200 epochs with batch-size 128 and Adam \citep{kingma2014adam} optimizer. The initial learning rate is 0.1, and it multiplies 0.1 for every 40 epochs. 

\vspace{-0.1in}
\paragraph{Training settings of UCI datasets \cite{Dua:2019}}
We adopted \citep{liu2020peer} a two-layer ReLU Multi-Layer Perceptron (MLP) for classification tasks on multiple UCI datasets, trained for 1000 episodes with batch-size 64 and Adam \citep{kingma2014adam} optimizer. We report the best performance for each smooth rate under a set of learning rate settings, $[0.0007, 0.001, 0.005, 0.01, 0.05]$.

\subsection{Additional experiment on $r^*$ and $r_{\text{opt}}$}

\paragraph{$r^*$ and $r_{\text{opt}}$ on synthetic dataset}
We generate 2D (binary) synthetic dataset by randomly sampling two circularly distributed classes. The inner annulus indicates one class (blue), while the outer annulus denotes the other class (red). Clearly, the generated synthetic dataset is well-separable (Type 1) and we hold $20\%$ data samples for performance comparison. The noise transition matrix takes a symmetric form with noise rate $e_i$ for 
both classes. To simulate the scenario where the clean data may not be perfectly separated due to a non-negligible amount of uncertainty samples clustering at the decision boundary, we flip the label of 50\% samples near the intersection of two annulus to the other class (Type 2). As specified in Table \ref{Tab:r_star_syn}, $r^*= [0.1, 0.4]$ for Type 1 data and $r^*=[0.0, 0.2]$ for Type 2 data. With the presence of label noise, the distribution of $r_{\text{opt}}$ shifts from non-negative ones to negative values. Even though NLS fails to outperform LS on clean data, we observe that NLS is less sensitive to noisy labels. Data with high level noise rates clearly favor NLS with a low smooth rate! 
\begin{table*}[!htb]
\vspace{-0.15in}
	\caption{Test accuracies of GLS on clean and noisy synthetic data. We report best test accuracy for each method. $r_{\text{opt}}$ and the corresponding test accuracy are highlighted (green: NLS; red: CE or LS).}
    \scriptsize\centering
    \begin{tabular}{c|ccc|ccc}
				\hline 
		\multirow{2}{*}{Method}  &  \multicolumn{3}{c}{\emph{Synthetic data (Type 1)}} &  \multicolumn{3}{c}{\emph{Synthetic data (Type 2)}}\\ 
				  &$e_i=0$ &$e_i=0.2$ &$e_i=0.4$   
				  &$e_i=0$ &$e_i=0.2$   &$e_i=0.4$\\
				\hline\hline
			   LS & \better{0.896} & 0.878 & 0.786 &\better{0.894}  & 0.848 & 0.842\\
	 	\hline
	 	Vanilla Loss & 0.889 & 0.882 & 0.806 & \better{0.894} & 0.875 & 0.868\\
	 	\hline
	 	NLS & 0.893 & \good{0.885} & \good{0.825} & 0.883 & \good{0.884} & \good{0.875}\\
	 	\hline\hline
	 	$r_{\text{opt}}=$ & \better{[0.1, 0.4]} & \good{-0.2} & \good{-0.4} & \better{[0, 0.2]} & \good{-0.3} & \good{-0.5}\\
	 	\hline
			\end{tabular}
		\vspace{-10pt}
		\label{Tab:r_star_syn}
	\end{table*}

\paragraph{$r^*$ and $r_{\text{opt}}$ on more UCI datasets}
We further test the performance of generalized label smoothing on 7 more UCI datasets (Heart, Breast 1, Breast 2, Diabetes, German, Image and Waveform). Our observation remains unchanged: there exists a general trend that with the increasing of noise rates, NLS becomes much more competitive than LS. Here, we attach the results of 4 additional UCI datasets for illustration. 

\begin{table*}[!htb]
	\vspace{-0.15in}
\caption{Test accuracy comparisons on clean and noisy UCI datasets (Image, Waveform, Heart, Banana) with best two smooth rates (green: NLS; red: CE or LS).}\scriptsize
		\begin{center}
		\scalebox{0.9}{\begin{tabular}{c|ccccc|ccccc}
				\hline 
		\multirow{2}{*}{Smooth Rate}  &  \multicolumn{5}{c}{\emph{Image}} &  \multicolumn{5}{c}{\emph{Waveform}}\\ 
				  &$e_i=0$ &$e_i=0.1$&$e_i=0.2$   &$e_i=0.3$ &$e_i=0.4$   
				  &$e_i=0$ &$e_i=0.1$&$e_i=0.2$   &$e_i=0.3$ &$e_i=0.4$\\
				\hline\hline
			   $r=0.8$	  & \better{0.993} &  0.983& \better{0.973} & \better{0.946} &  0.875 & 0.939 & 0.935 & 0.931 & 0.927 & 0.885\\
			   	 \hline 
			  $r=0.6$ 	  & \better{0.993} &  \better{0.987}& 0.970 & \better{0.939} &  0.869& \better{0.943} & \better{0.943} &  \better{0.943}& 0.929 & 0.901\\
			   	 \hline 
			   $r=0.4$	  & \better{0.997} &  0.980& \better{0.973} & \better{0.939} & 0.865 & 0.941 & 0.937 & \better{0.943} & 0.931 & 0.905\\
			   	 \hline 
			  $r=0.2$ 	  & \better{0.993} &  \better{0.993}& 0.966 & 0.936 & 0.875 & 0.941 & 0.935 &  0.933& 0.931 & 0.913\\
			   	 \hline 
			   $r=0.0$	  & 0.990 &  0.976& 0.963 & 0.929 & 0.865 & \better{0.945} & 0.935 & 0.937 & \better{0.933} & 0.911\\
			   	 \hline 
			  $r=-0.2$ 	  & 0.912 &  0.96& 0.953 & 0.919 &  0.872& 0.937 & \good{0.939} & 0.939 & \good{0.933} & 0.907\\
			  \hline 
			  $r=-0.4$	  & 0.882 &  0.923& 0.953 & 0.936 & 0.872 & 0.925 & 0.937 & 0.939 & \good{0.933} & \good{0.917}\\
			  \hline 
			  $r=-0.8$ 	  & 0.842 &  0.882& 0.926 & 0.933 &  0.872& 0.921 & 0.925 &  0.939& 0.931 & \good{0.923}\\
			   \hline 
			  $r=-1.0$ 	  & 0.832 &  0.869& 0.909 & 0.929 & \good{0.882}& 0.921 & 0.923 & 0.933 & 0.929 & 0.907\\
			   \hline 
			  $r=-2.0$ 	  & 0.818 &  0.815& 0.889 & 0.909 & \good{0.906} & 0.911 & 0.913 & 0.921 & 0.927 & 0.911\\
	 	\hline 	\hline 
		\multirow{2}{*}{Smooth Rate}  &  \multicolumn{5}{c}{\emph{Twonorm}} &  \multicolumn{5}{c}{\emph{Banana}}\\ 
				  &$e_i=0$ &$e_i=0.1$&$e_i=0.2$   &$e_i=0.3$ &$e_i=0.4$   
				  &$e_i=0$ &$e_i=0.1$&$e_i=0.2$   &$e_i=0.3$ &$e_i=0.4$\\
				\hline\hline
			   $r=0.8$	& \better{0.990} &   \better{0.990} & 0.986 & 0.982 & 0.968  & 0.896 &  \better{0.893} &  \better{0.876} & 0.847  & 0.790\\
			   	 \hline 
			  $r=0.6$ 	  & \better{0.990} &   0.989 & 0.987 & 0.981 & 0.972 & \better{0.903}  & 0.881  & \better{0.876}  &    0.855& 0.811 \\
			   	 \hline 
			  $r=0.4$	  & \better{0.990} &   \better{0.990} & 0.987 & 0.983 & 0.971 & \better{0.900}  & 0.887  & 0.874  &    \better{0.859}&  0.807 \\
			   	 \hline 
			  $r=0.2$ 	  & \better{0.990} &  0.989 & 0.986 & 0.985 & 0.969  & 0.896  & \better{0.894}  & \better{0.876}  &    0.856& 0.810 \\
			   	 \hline 
			    $r=0.0$	  & \better{0.990} &  0.989 & 0.987 & 0.985 & 0.973 & 0.897  & 0.881  & 0.871  &    0.849 & 0.833  \\
			   	 \hline 
			   $r=-0.4$	  & 0.986 &   0.988 & \good{0.988} & \good{0.986} & 0.972   & 0.847  & 0.874  & 0.859  & 0.853   & \good{0.840} \\
			  \hline 
			  $r=-0.6$ 	  & 0.986 &  0.988 &0.987 & 0.984 &  0.974 & 0.845  & 0.864  & 0.861  & \good{0.859}    & \good{0.837} \\
			  \hline 
			   $r=-1.0$ 	  & 0.986 &  0.986 & \good{0.988}& 0.985 & 0.977  & 0.796  &  0.812 & 0.852  & 0.854   & 0.811  \\
			   \hline 
			  $r=-2.0$ 	  & 0.986 &  0.986 & 0.986 & \good{0.986} & 0.978   & 0.759  & 0.764  &  0.819 & 0.852   & 0.819 \\
			   \hline 
			   $r=-4.0$ 	  & 0.986 &  0.986& 0.986 & \good{0.986} & \good{0.983}  & 0.718  &0.723   & 0.738  &  0.787   & 0.813 \\
			   \hline 
			  $r=-8.0$ 	  & 0.986 &  0.986 & 0.986 & 0.985 & \good{0.986}  & 0.703  & 0.700  & 0.699  & 0.735    & 0.735\\
	 	\hline
			\end{tabular}}
			\end{center}
		\vspace{-0.1in}
		\label{Tab:more_uci}
	\end{table*}

The noisy labels are generated by a symmetric noise transition matrix with noise rate $e_i=[0.1, 0.2, 0.3, 0.4]$. As highlighted in Table \ref{Tab:more_uci}, $r_{\text{opt}}$ appears with positive values when the data is clean (same as $r^*$) or of a low noise rate. With the increasing of noise rates, NLS becomes more competitive than LS. We color-code different noise regimes where either LS (red-ish) or NLS (green-ish) outperforms the other. Clearly, there is a separation of the favored smoothing rate for different noise scenarios (upper left \& low noise for LS, bottom right \& high noise for NLS).

\paragraph{$r^*$ and $r_{\text{opt}}$ on AGNews}
We next provide an additional empirical justification of Theorem \ref{ass:opt_r}. Note that when we have access to $r^*$, Theorem \ref{thm:robust_glsr_multi} reveals what smooth rate recovers the performance on the clean data when learning with noisy labels. We adopt an NLP dataset AGNews \citep{zhang2015character} for illustration. In Table \ref{tab:agnews}, we do observe that $r_{\text{opt}}$ achieves the best performance for most noise settings.
\vspace{-0.1in}
\begin{table}[!htb]
	\caption{Test accuracy comparisons on clean and symmetric noisy AGNews dataset. Highlighted numbers indicate the best performance under each $\epsilon$.}
\small
		\begin{center}
		\scalebox{0.75}{\begin{tabular}{c|ccccc}
				\hline 
		\multirow{2}{*}{Smooth Rate}  &  \multicolumn{5}{c}{AGNews (4 classes)}\\ 
				  &$\epsilon=0$ &$\epsilon=0.1$&$\epsilon=0.2$   &$\epsilon=0.3$ &$\epsilon=0.4$\\
				\hline\hline
			   $r=0.4$	   &  86.33 & 85.55   & 83.93   & 82.29  & 79.80  \\
			  $r=0.2$ 	    &  87.79 &    86.99&  85.67  & 83.47  & 81.04 \\
			  \cellcolor{blue!10}  $r=0.0$	  & \cellcolor{blue!10}  \textbf{88.20} &  87.79  &  86.80  &  85.24 & 82.39\\
			   $r=-0.15$	  & 85.04  &  \cellcolor{blue!10} \textbf{88.00}  & 87.47    & 85.83  & 83.09 \\
			  $r=-0.2$ 	 & 84.08  &  87.30  &   87.50 &  85.85 & 83.34  \\
			   $r=-0.36$	  & 81.39  &   84.47 &  \cellcolor{blue!10} \textbf{87.75}   & 86.14  & 83.62 \\
			  $r=-0.4$ 	 &  80.76 &  83.99  &  87.28  &  86.36 & 83.96   \\
			  $r=-0.6$	  & 77.62  & 80.80   &  84.68  & \cellcolor{blue!10}  \textbf{87.26} & 84.37\\
			   $r=-0.67$	  & 76.70  &   79.91 &   83.87  & 87.21 & 84.58 \\
			   $r=-1.14$	  &  72.38 &  74.84  &   78.28  & 82.45  &\cellcolor{blue!10} \textbf{86.43} \\
			  \hline  $r=r_{\text{opt}}=\frac{(K-1)r^*-K\epsilon}{(K-1)-K\epsilon}$ & \textbf{88.20}  &  \textbf{88.00}  & \textbf{87.75}  & 87.21  &\textbf{86.43} \\
	 	\hline		\end{tabular}}
			\end{center}
		\label{tab:agnews}\vspace{-0.15in}
	\end{table}

\subsection{Additional experiment results on model confidence}

\begin{figure}[!htb]
\centering
\vspace{-0.15in}
\includegraphics[width=0.21\textwidth]{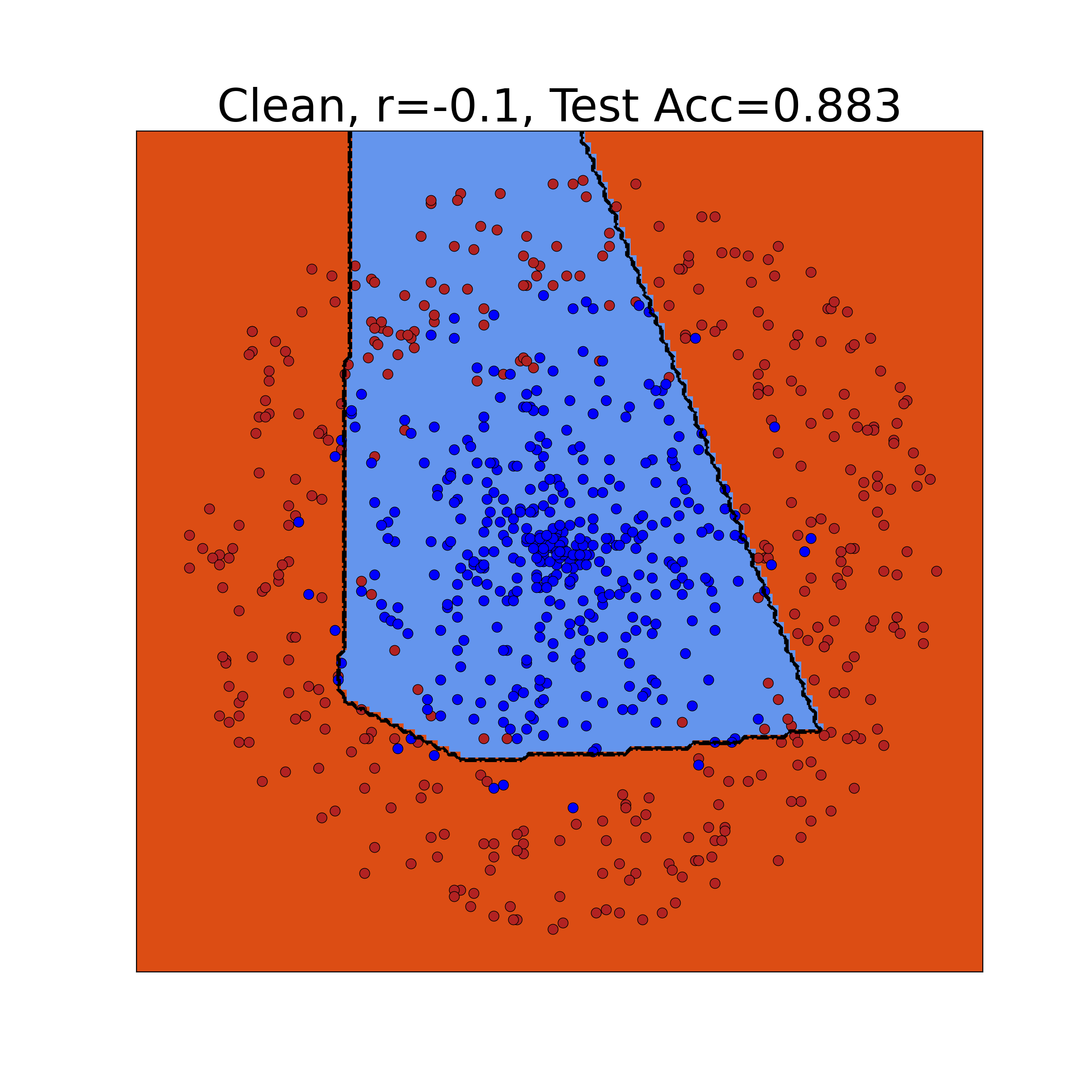}
\includegraphics[width=0.21\textwidth]{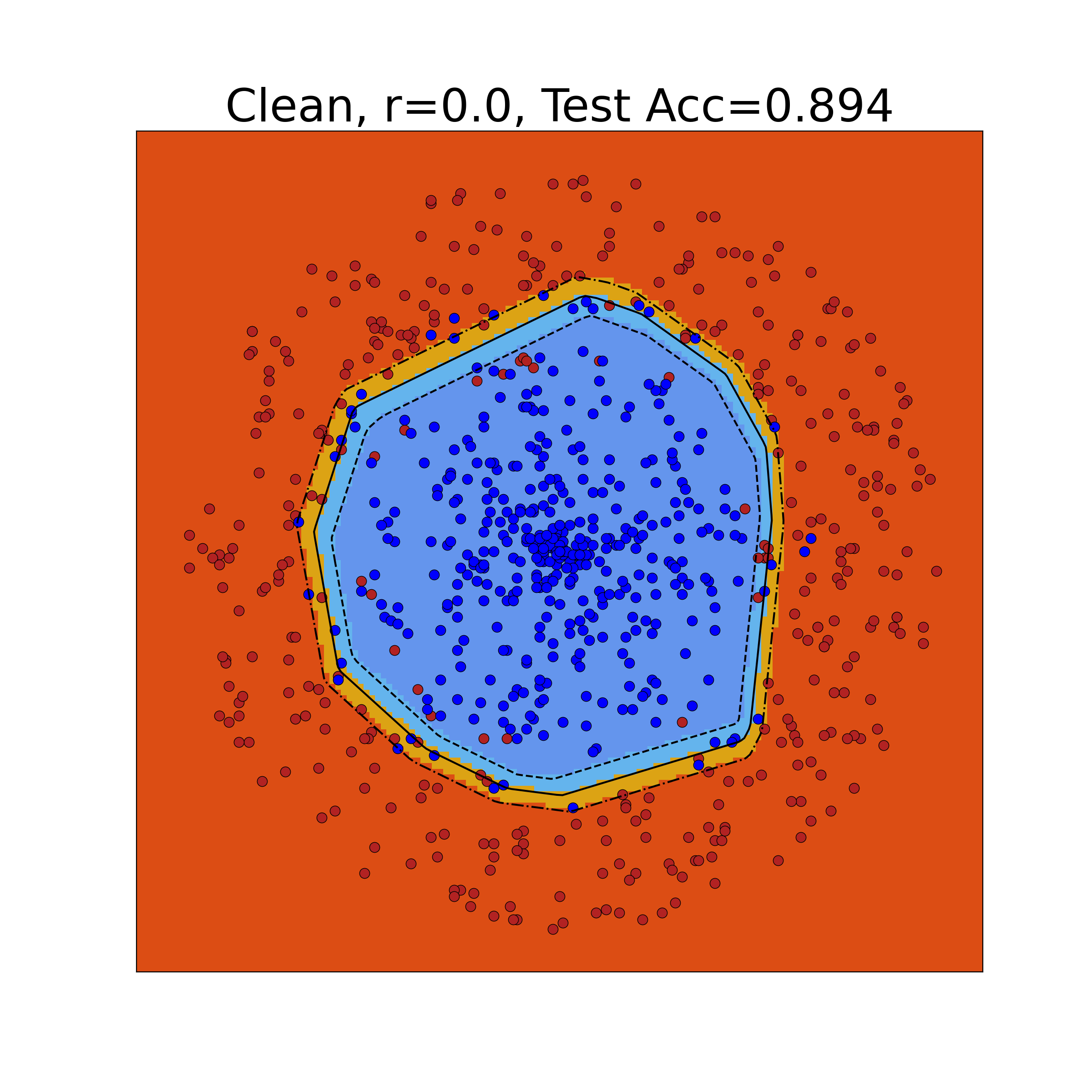}
\includegraphics[width=0.21\textwidth]{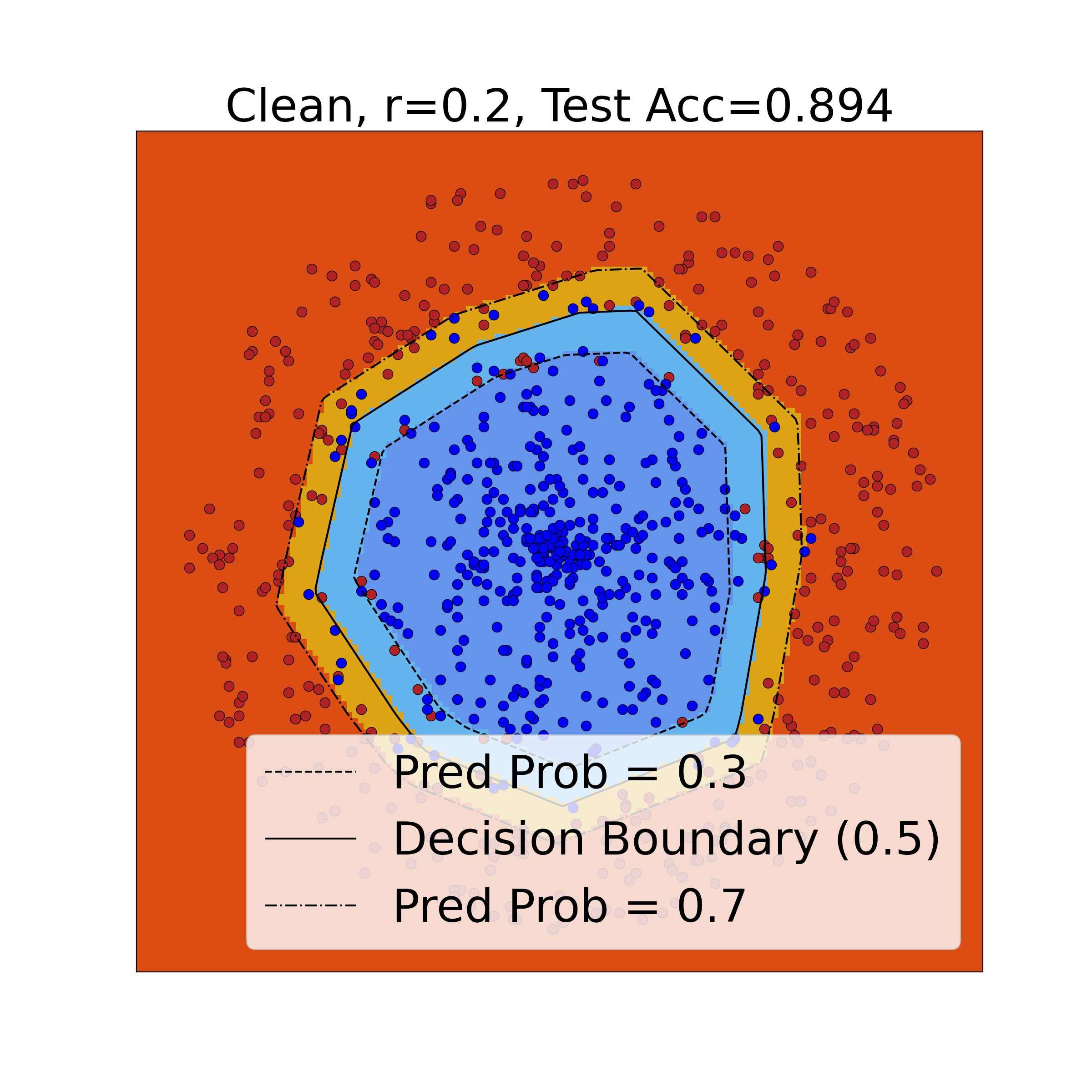}
\vspace{-0.15in}
\caption{Model confidence visualization of NLS, VL and LS on synthetic data (Type 2) with the clean data. $r^*\in [0, 0.2]$. (left: NLS; middle: Vanilla Loss; right: LS).}\label{fig:type2_1}\vspace{-0.1in}
\end{figure}

\begin{figure}[!htb]
\centering
\vspace{-0.15in}
\includegraphics[width=0.21\textwidth]{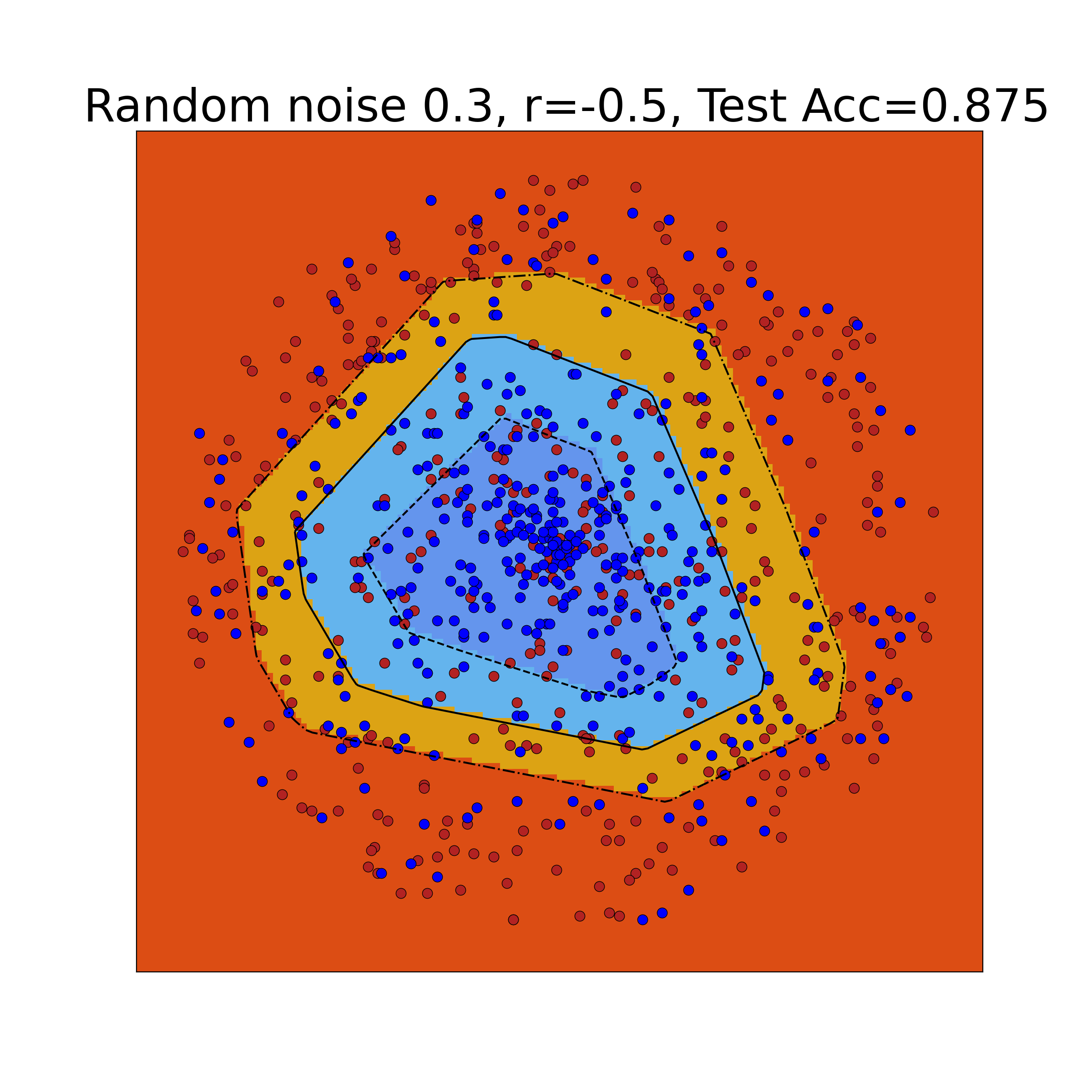}
\includegraphics[width=0.21\textwidth]{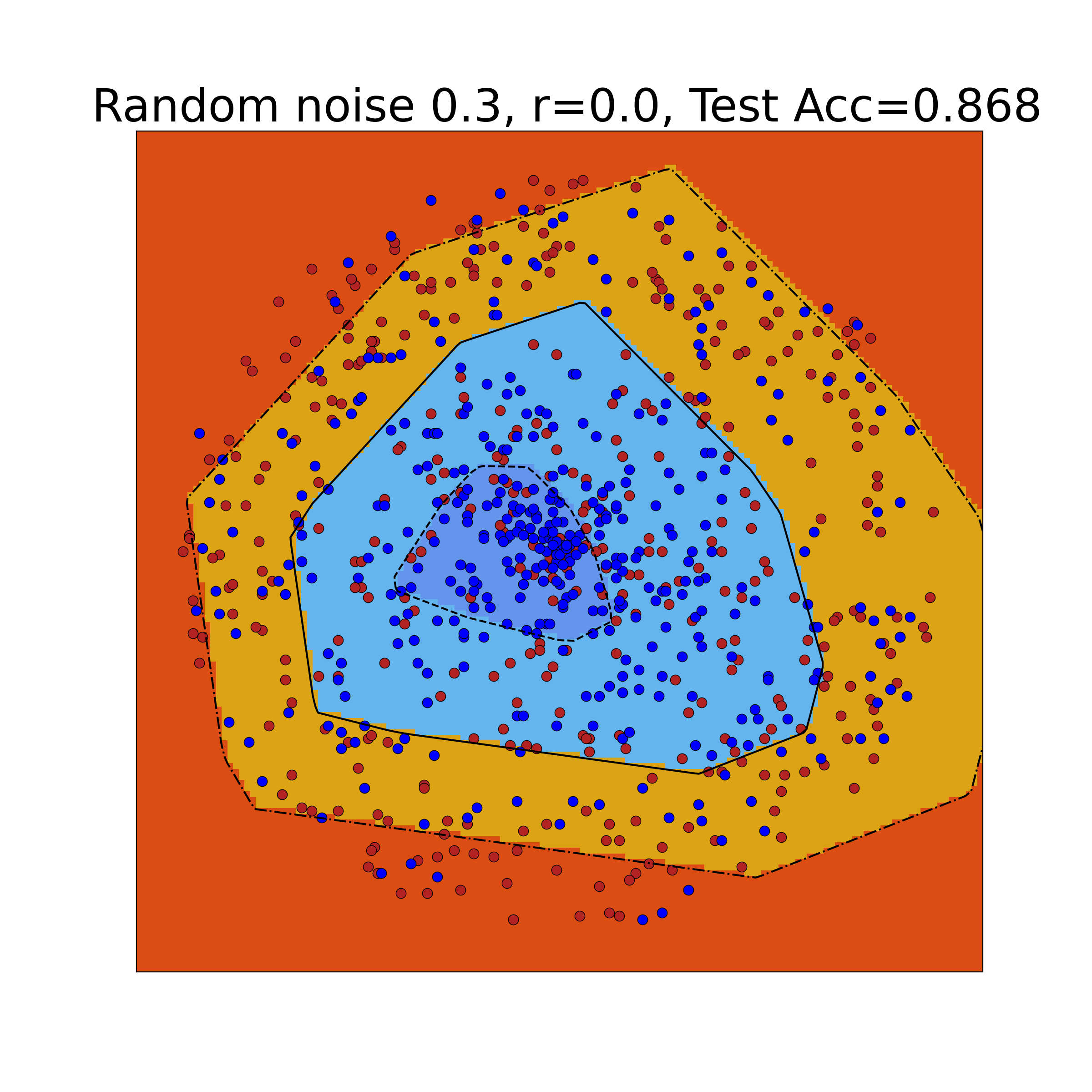}
\includegraphics[width=0.21\textwidth]{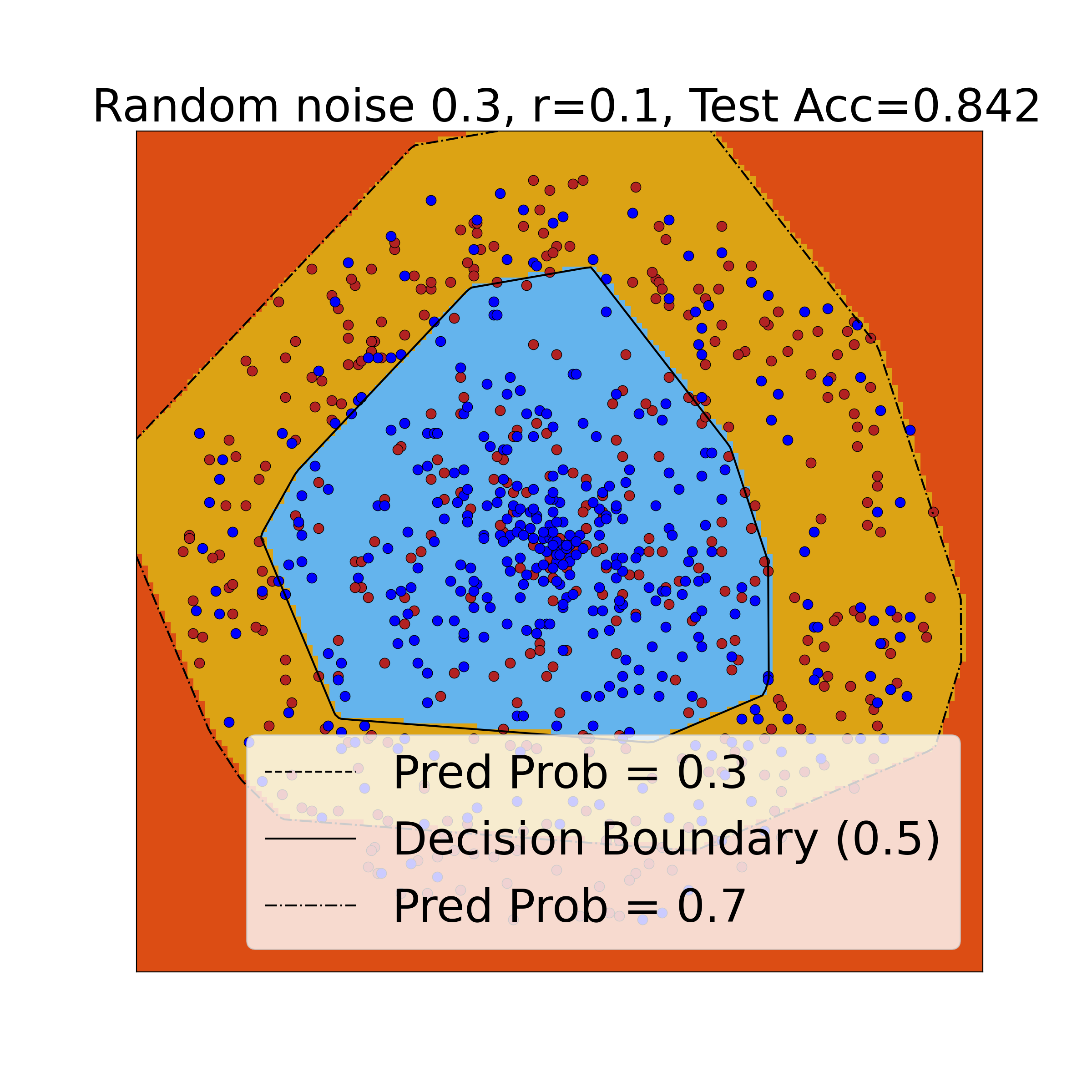}
\vspace{-0.15in}
\caption{Model confidence visualization of NLS, VL and LS on synthetic data (Type 2) with noise rate $e_i=0.3$. $r_{\text{opt}}=-0.5$. (left: NLS; middle: Vanilla Loss; right: LS).}\label{fig:type2_3}
\end{figure}

\paragraph{NLS improves model confidence on Synthetic Type 2 dataset}
In this case, the clean data that are close to decision boundary distributes randomly. In Figure \ref{fig:type2_1}-\ref{fig:type2_3}, the colored bands depict the different levels of prediction probabilities. When the smooth rate increases from negative to positive, more samples fall in the orange and light blue band which indicates uncertain predictions. When the smooth rate increases from negative to positive, learning with smoothed labels will result in more uncertain predictions. With the increasing of noise rates ($e_i=0\to 0.4$), Learning with a fixed smooth rate generally becomes less confident on its predictions. Thus, a smaller smooth rate is required when the noise rate increases.

\vspace{-0.2in}

\subsection{Effect of LS and NLS on pre-logits}
\begin{figure}[!htb]
\centering
\includegraphics[width=0.30\textwidth]{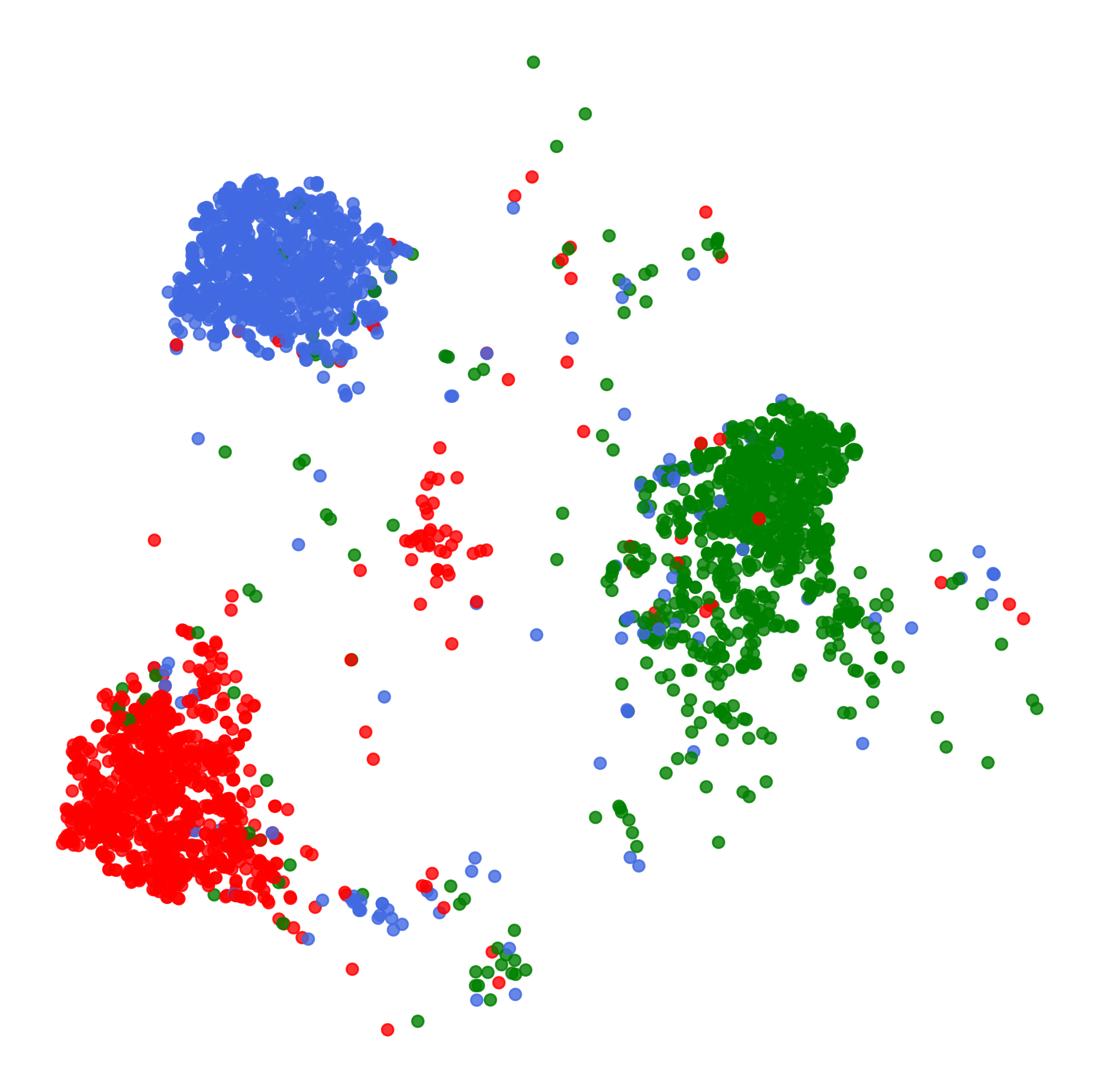}
\hspace{0.1in}
\includegraphics[width=0.30\textwidth]{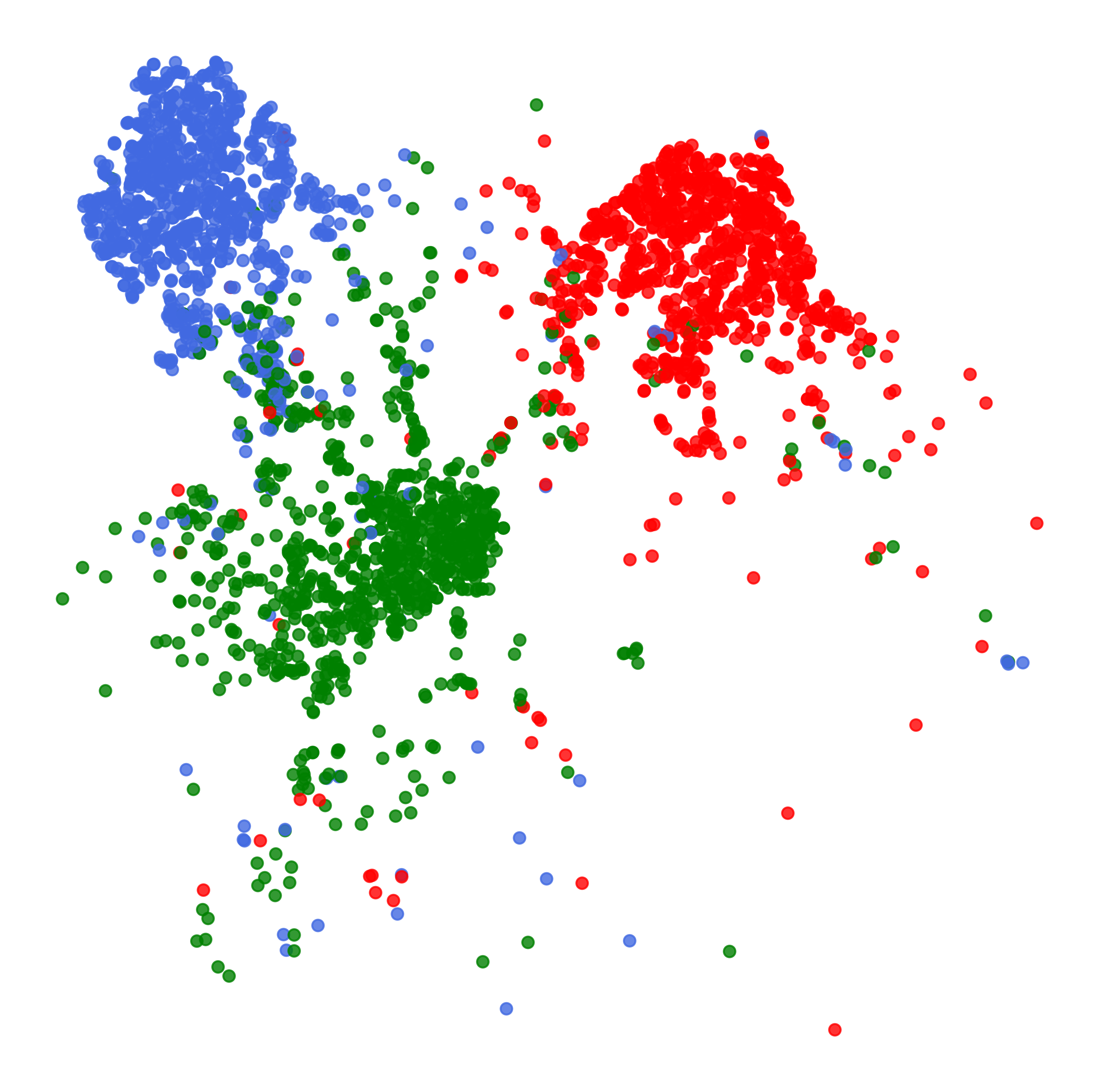}
\hspace{0.1in}
\includegraphics[width=0.30\textwidth]{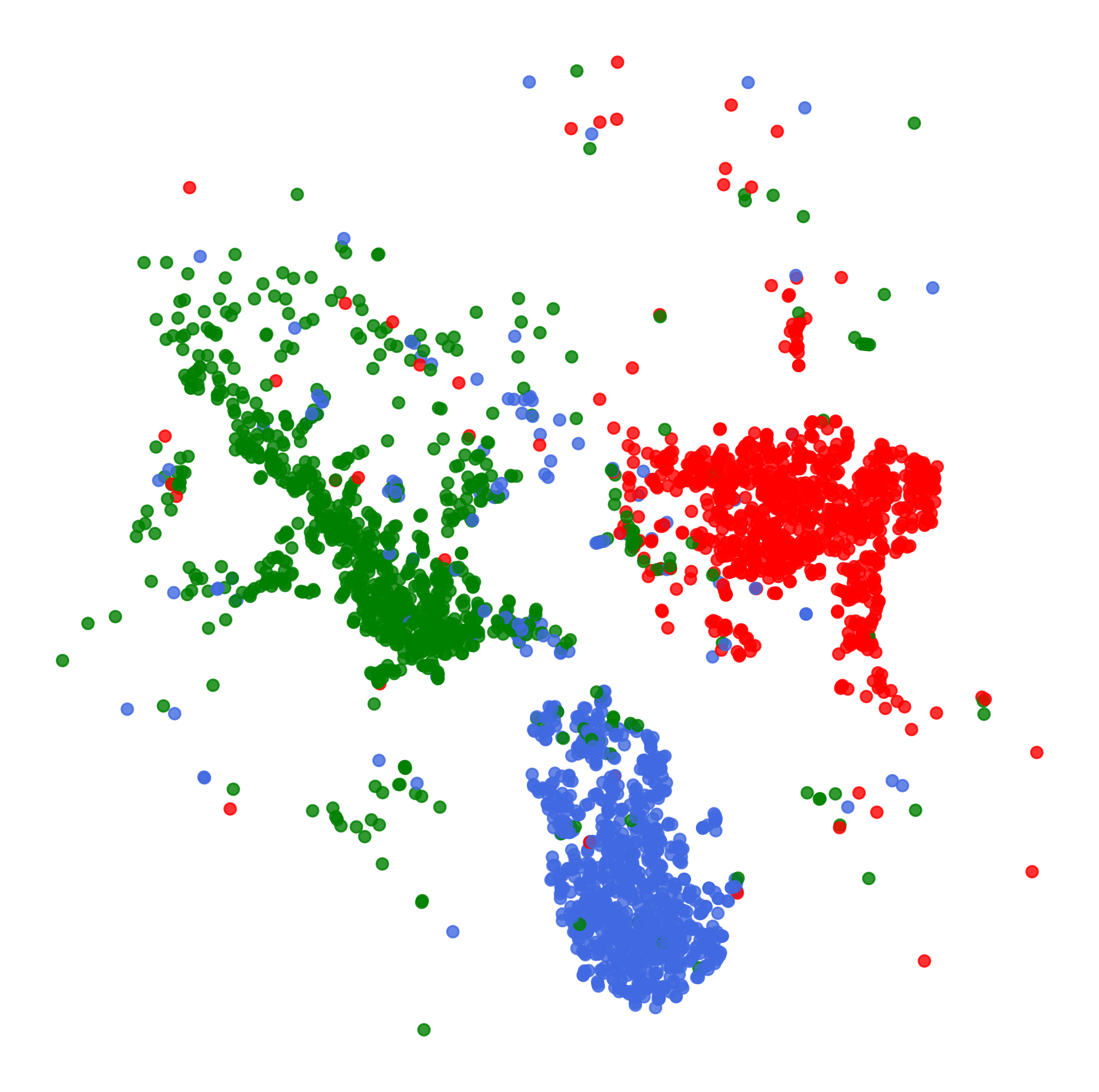}
\vspace{-0.1in}
\caption{Effect of GLS on pre-logits (left: NLS; middle: Vanilla Loss; right: LS; trained with symmetric 0.2 noisy CIFAR-10 training dataset).}\label{tsne_cifar_02}
\end{figure}

We visualise the pre-logits of a ResNet-34 for three classes on CIFAR-10. We adopt the method from \citep{muller2019does} which illustrates how representations differ
between penultimate layers of networks trained with different smooth rates in GLS. In Figure \ref{tsne_cifar_02}, NLS makes the model $f$ be confident on her predictions and the distances between three clusters are clearly larger than those appeared in Vanilla Loss and LS.

\section{Bias and Variance Trade-off of Learning with Smoothed Labels}
\label{app:bias_variance}
Denote $\hat{f}_{\text{H}}$, $\hat{f}_{\text{S}}$ as pre-trained models on the training dataset $D$ w.r.t. hard labels and soft labels, respectively. The vector form of the prediction w.r.t. sample $x$ given by $\hat{f}_{\text{H}}$ and $\hat{f}_{\text{S}}$ are $\mph$ and $\mps$. For the ease of presentation, we relate notations with subscript H/S to hard/soft labels without further explanation. Given the sample $x$ and the one-hot label $\by$, we denote the averaged model prediction by:
\begin{align*}
\bh:=\frac{1}{Z_\text{H}}\exp^{\E_{ D}\log(\mph)},\quad \bs:=\frac{1}{Z_\text{S}}\exp^{\E_{ D}\log(\mps)},
\end{align*}
where $Z_{\text{H}}, Z_{\text{S}}$ are normalization constants. The bias of model prediction is defined as the KL divergence $D_{\text{KL}}$ between target distribution (one-hot encoded vector form) $\by$ and the averaged model prediction. 
\begin{align*}
    \text{Bias}_\text{H}:=\E_{x, \by}\Big[\by \log\frac{\by}{\bh}\Big], \quad
    \text{Bias}_\text{S}:=\E_{x, \by}\Big[\by \log\frac{\by}{\bs}\Big].
\end{align*}
While the variance of model prediction measures the expectation of KL divergence between the averaged model prediction and model prediction over $D$:
\begin{align*}
    \text{Var}_\text{H}:=\E_{D}\Bigg[\E_{x, \by}\Big[ \bh \log\Big(\frac{\bh}{\mph}\Big)\Big]\Bigg], \quad  \text{Var}_\text{S}:=\E_{D}\Bigg[\E_{x, \by}\Big[ \bs \log\Big(\frac{\bs}{\mps}\Big)\Big]\Bigg].
\end{align*}

Empirical observation from \citep{zhou2021rethinking} shows that the variance brought by learning with positive soft labels given by a teacher's model \citep{hinton2015distilling} is less than the direct training w.r.t hard labels. As an extension, we are interested in how LS/NLS interferes with the bias and variance of model prediction. 

\begin{figure}[!ht]
\centering
  \begin{center}
    \includegraphics[width=0.44\textwidth]{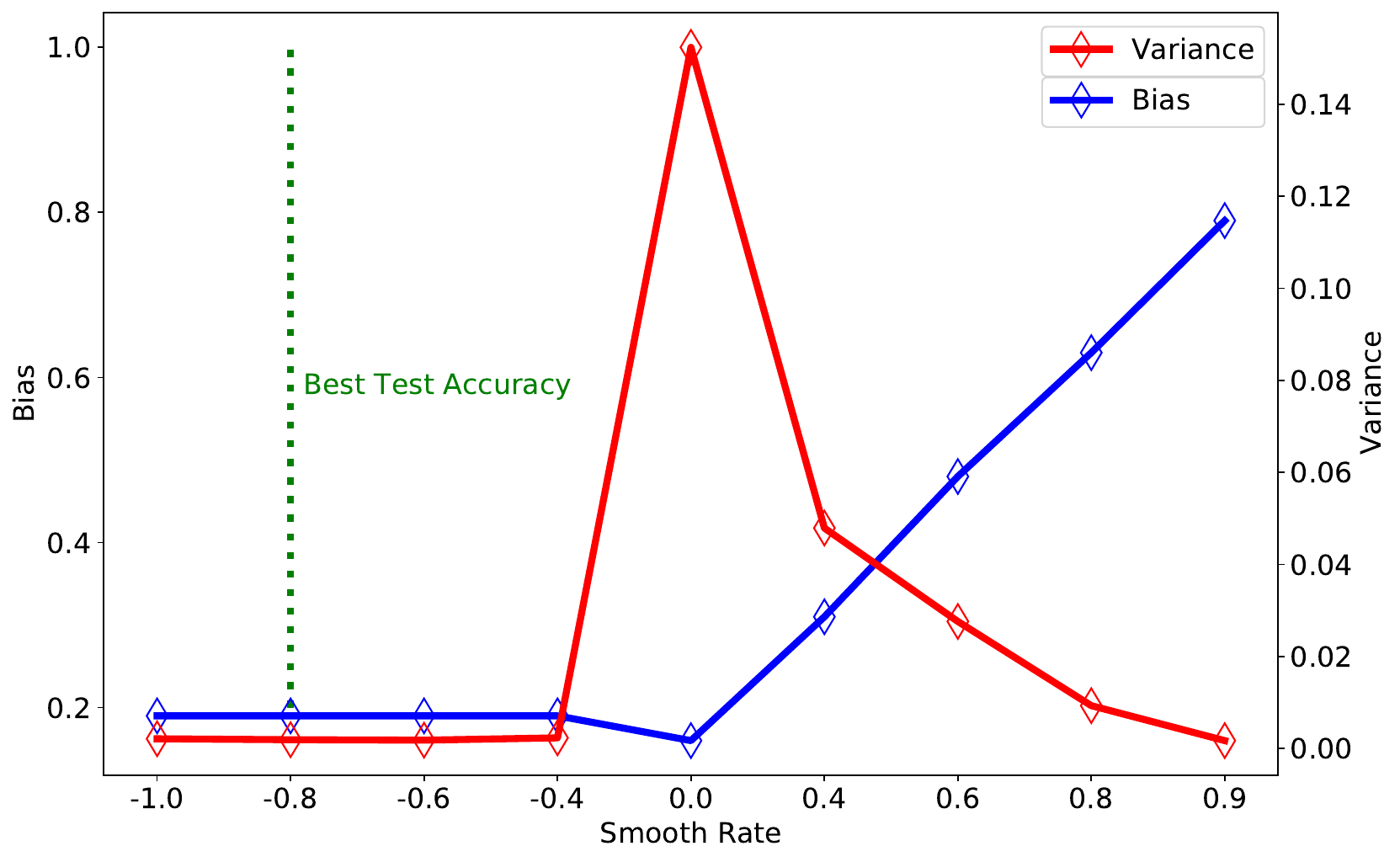}
  \end{center}
    \vspace{-15pt}
  \caption{Bias and variance of pre-trained LS/VL/NLS models on clean CIFAR-10 test dataset.}
    \label{fig:bv_cifar10}
\end{figure}

\paragraph{Bias and variance of LS/NLS on clean dataset} We introduce our empirical observation regarding the role of LS/NLS in bias and variance trade-off in Figure \ref{fig:bv_cifar10}. We select nine smooth rates of LS/NLS for illustration. Each smooth rate setting of LS/NLS trains on the CIFAR-10 dataset for 5 times with different data augmentations. To estimate the variance and bias of pre-trained models, we adopt the implementation in \citep{yang2020rethinking}. Empirical results show that learning directly with a larger positive smooth rate typically results in lower variance and higher bias. In Figure \ref{fig:bv_cifar10}, we can observe almost constant bias values and very low variance for NLS. This is best explained by the warm-up of pre-trained models and the fact that NLS pushes the classifier to give confident predictions. As for LS, with the increase of smooth rate, the overall bias has an increasing tendency while the variance has the decreasing pattern. Especially when the smooth rate approaches to 1, i.e., $r=0.9$, the variance is close to 0.

\section{Omitted Proofs}\label{app:proofs}

We observe that NLS connects to a special case of label smoothing regularization \citep{szegedy2016rethinking}. We highlight this in Theorem \ref{thm:glsr}.

\begin{theorem}
\label{thm:glsr}
$\forall r\in [0, 1]$, NLS with smooth rate $-r$ is a special form of label smoothing regularization:
\vspace{-0.06in}
\begin{align*}
&\min_{f \in \mathcal F} \mathbb{E}_{(X, \nY)\thicksim \nD}\Big[ \ell(\bfX, \nY^{\text{GLS},-r})\Big] =\min_{f \in \mathcal F} \mathbb{E}_{(X, \nY)\thicksim \nD}\Big [2 \cdot \ell(\bfX, \nY) -\ell(\bfX, \nY^{\text{GLS}, r})\Big].
\end{align*}
\vspace{-0.06in}
\end{theorem}
\subsection{Proof of Theorem \ref{thm:glsr}}
Before we prove Theorem \ref{thm:glsr}, we first introduce Lemma \ref{lm:loss_nls}.
\begin{lemma}
\label{lm:loss_nls}
$\forall (x, \by^{\text{GLS}, r})$, $
\ell\big(\bfx, \by^{\text{GLS}, r}\big)= \big(1-\frac{r}{2}\big)\cdot \ell\big(\bfx, y\big)+\frac{r}{2}\cdot \ell\big(\bfx, 1-y\big).
$
\end{lemma}

\paragraph{Proof of Lemma \ref{lm:loss_nls}}
\begin{proof}
For CE loss, due to its linear property w.r.t. the label, we directly have:
\begin{align*}
    \ell(\bfx, \by^{\text{GLS}, r})&=\ell\big(\bfx, (1-r)\cdot \by+\frac{r}{2}\cdot \bo\big)=\big(1-\frac{r}{2}\big)\cdot \ell\big(\bfx, y\big)+\frac{r}{2}\cdot \ell\big(\bfx, 1-y\big).
\end{align*}
\end{proof}
\paragraph{Proof of Theorem \ref{thm:glsr}}
\begin{proof}
Based on Lemma \ref{lm:loss_nls}, with a bit of math, for NLS, we have:
\begin{align*}
    &\min_{f \in \mathcal F} \mathbb{E}_{(X, \nY)\thicksim \nD} \Big[\ell\big(\bfX, \nY^{\text{GLS}, -r}\big)\Big]\\
    =&\min_{f \in \mathcal F} \mathbb{E}_{(X, \nY)\thicksim \nD} \Big[\big(1+\frac{r}{2}\big)\cdot \ell\big(\bfX, \nY\big)-\frac{r}{2}\cdot \ell\big(\bfX, 1-\nY\big)\Big]\\
    =&\min_{f \in \mathcal F} \mathbb{E}_{(X, \nY)\thicksim \nD} \Big[\big[\big(1+\frac{r}{2}\big)+ \big(1-\frac{r}{2}\big)\big]\cdot \ell\big(\bfX, \nY\big) - \big[\big(1-\frac{r}{2}\big)\cdot \ell\big(\bfX, \nY\big) + \frac{r}{2}\cdot \ell\big(\bfX, 1-\nY\big)\big]\Big]\\
    =&\min_{f \in \mathcal F} \mathbb{E}_{(X, \nY)\thicksim \nD}\Big [2 \cdot \ell\big(\bfX, \nY\big) -\ell\big(\bfX, \nY^{\text{GLS}, r}\big)\Big].
\end{align*}

\end{proof}

\subsection{Proof of Theorem \ref{thm:glsr_conf}}
\begin{proof}
\begin{align*}
    Eqn.\ref{eqn:gen} 
    =& \min_{f \in \mathcal F}~ \E_{(X, \nY)\thicksim \nD} \Big[\underbrace{\big(1-\frac{ r}{2}\big)}_{:=c_1} \cdot \ell\big(\bfX, \nY\big) + \underbrace{\frac{r}{2}}_{:=c_2} \cdot \ell\big(\bfX, 1-\nY\big)\Big]\\
    =&\min_{f \in \mathcal F}~ \mathbb{E}_{X, Y=0} \Big[\p(\nY=0|Y=0)\cdot \Big(c_1\cdot \ell\big(\bfX, 0\big) + c_2\cdot \ell\big(\bfX, 1\big)\Big) \\
    &+\p(\nY=1|Y=0)\cdot \Big(c_1\cdot \ell\big(\bfX, 1\big) + c_2\cdot \ell\big(\bfX, 0\big)\Big)\Big]\\
    &+ \mathbb{E}_{X, Y=1} \Big[\p(\nY=0|Y=1)\cdot \Big(c_1\cdot \ell\big(\bfX, 0\big) + c_2\cdot \ell\big(\bfX, 1\big)\Big) \\
    & +\p(\nY=1|Y=1)\cdot \Big(c_1\cdot \ell\big(\bfX, 1\big) + c_2\cdot \ell\big(\bfX, 0\big)\Big)\Big]\\
    =&\min_{f \in \mathcal F}~ \mathbb{E}_{X, Y=0} \Big[\big[(1-e_0)\cdot c_1 +e_0 \cdot c_2 \big]\cdot \ell\big(\bfX, 0\big) + \big[(1-e_0)\cdot c_2 +e_0 \cdot c_1 \big]\cdot \ell\big(\bfX, 1\big)\Big]\\
    &+ \mathbb{E}_{X, Y=1} \Big[\big[(1-e_1)\cdot c_1 +e_1 \cdot c_2 \big]\cdot \ell\big(\bfX, 1\big) + \big[(1-e_1)\cdot c_2 +e_1 \cdot c_1 \big]\cdot \ell\big(\bfX, 0\big)\Big]\\
    =&\min_{f \in \mathcal F}~ \mathbb{E}_{X, Y=0} \Big[\big[(1-e_0)\cdot c_1 +e_0 \cdot c_2 \big]\cdot \ell\big(\bfX, 0\big) + \big[(1-e_0)\cdot c_2 +e_0 \cdot c_1\big ]\cdot \ell\big(\bfX, 1\big)\Big]\\
    &+ \mathbb{E}_{X, Y=1} \Big[\big[(1-e_0)\cdot c_1 +e_0 \cdot c_2 \big]\cdot \ell\big(\bfX, 1\big) + \big[(1-e_0)\cdot c_2 +e_0 \cdot c_1 \big]\cdot \ell\big(\bfX, 0\big)\Big]\\
    &+\mathbb{E}_{X, Y=1} \Big[e_{\Delta}\cdot (c_2-c_1)\cdot \ell\big(\bfX, 1\big) - e_{\Delta}\cdot (c_2-c_1)\cdot \ell\big(\bfX, 0\big)\Big]\\ =& \min_{f \in \mathcal F}~ \mathbb{E}_{(X, Y)\sim \D}  \Big[\big[(1-e_0)\cdot c_1 +e_0 \cdot c_2 \big]\cdot \ell\big(\bfX, Y\big) +  \big[(1-e_0)\cdot c_2 +e_0 \cdot c_1 \big]\cdot \ell\big(\bfX, 1-Y\big) \Big]\\
    &-e_{\Delta}\cdot (c_1-c_2)\cdot \mathbb{E}_{X, Y=1}  \Big[\ell\big(\bfX, 1\big) - \ell\big(\bfX, 0\big)\Big]\\
    =& \min_{f \in \mathcal F}~ \mathbb{E}_{(X, Y)\sim \D}\Big[ (c_1+c_2)\cdot \ell\big(\bfX, Y\big) \Big] \\
    &+  \big[(1-e_0)\cdot c_2 +e_0 \cdot c_1 \big]\cdot \mathbb{E}_{(X, Y)\sim \D}\Big[\ell\big(\bfX, 1-Y\big) - \ell\big(\bfX, Y\big)\Big] \\
    &-e_{\Delta}\cdot (c_1-c_2)\cdot \mathbb{E}_{X, Y=1}  \Big[\ell\big(\bfX, 1\big) - \ell\big(\bfX, 0\big)\Big]\\
    =& \min_{f \in \mathcal F}~ \mathbb{E}_{(X, Y)\sim \D}\Big[ (c_1+c_2)\cdot \ell\big(\bfX, Y^*\big) \Big] \\
    &+  \big[-\frac{r^*}{2}+(1-e_0)\cdot c_2 +e_0 \cdot c_1 \big]\cdot \mathbb{E}_{(X, Y)\sim \D}\Big[\ell\big(\bfX, 1-Y\big) - \ell\big(\bfX, Y\big)\Big] \\
    &-e_{\Delta}\cdot (c_1-c_2)\cdot \mathbb{E}_{X, Y=1}  \Big[\ell\big(\bfX, 1\big) - \ell\big(\bfX, 0\big)\Big]\\
   =&\min _{f \in \mathcal F}~ \underbrace{ \mathbb{E}_{(X, Y)\sim \D}   \Big[\ell\big(\bfX, Y^*\big)\Big] }_{\text{True Risk}}\underbrace{+  \lambda_1 \cdot \mathbb{E}_{(X, Y)\sim \D} \Big[\ell\big(\bfX, 1-Y\big) - \ell\big(\bfX, Y\big)\Big]}_{\text{M-Inc1}}\notag\\ 
    &\underbrace{+ \lambda_2 \cdot \mathbb{E}_{X, Y=1}  \Big[\ell\big(\bfX, 0\big) - \ell\big(\bfX, 1\big)\Big]}_{\text{M-Inc2}}.
\end{align*}
\end{proof}

\subsection{Proof of Theorem \ref{thm:radmaker}}
\begin{proof}
With the Rademacher bound on the maximal deviation between risks and empirical ones, for $\forall f\in \mathcal{F}$ and with probability at least $1-\delta$, we have:
\begin{align*}
   \max_{f\in\mathcal{F}}|R_{\text{{emp}}}^r(f)-R_{\text{exp}}^r(f)|\leq& 2 \mathfrak{R}(\ell^{\text{GLS}, r} \circ \mathcal{F})+\left(\overline{\ell^{\text{GLS}, r}}-\underline{\ell^{\text{GLS}, r}}\right)\cdot\sqrt{\frac{\log(1/\delta)}{2N}},
\end{align*}
where we define $\ell^{\text{GLS}, r}\big(\bfx, \by^{\text{GLS}, r}\big)= \big(1-\frac{r}{2}\big)\cdot \ell\big(\bfx, y\big)+\frac{r}{2}\cdot \ell\big(\bfx, 1-y\big)$, and $\mathfrak{R}$ indicates the Rademacher complexity. If $\ell$ is $L$-Lipshitz for every $y$, then for any $\mathbf{f}_1(x), \mathbf{f}_2(x)$, we have: $|\ell(\mathbf{f}_1(x), y)-\ell(\mathbf{f}_2(x), y)|\leq L|[\mathbf{f}_1(x)]_y-[\mathbf{f}_2(x)]_y|$. $\ell^{\text{GLS}, r}$ is also $L^{r}$-Lipshitz such that for CE loss, we have:
\begin{align*}
    &\left|\big(1-\frac{r}{2}\big)\cdot \ell\big(\mathbf{f}_1(x), y\big)+\frac{r}{2}\cdot \ell\big(\mathbf{f}_1(x), 1-y\big)-\big(1-\frac{r}{2}\big)\cdot \ell\big(\mathbf{f}_2(x), y\big)-\frac{r}{2}\cdot \ell\big(\mathbf{f}_2(x), 1-y\big)\right|\\
    = & \left|\big(1-\frac{r}{2}\big)\cdot \ell\big(\mathbf{f}_1(x), y\big)+\frac{r}{2}\cdot \ell\big(\bo-\mathbf{f}_1(x), y\big)-\big(1-\frac{r}{2}\big)\cdot \ell\big(\mathbf{f}_2(x), y\big)-\frac{r}{2}\cdot \ell\big(\bo-\mathbf{f}_2(x), y\big)\right|\\
    = & \left|\big(1-\frac{r}{2}\big)\cdot \left(\ell\big(\mathbf{f}_1(x), y\big)-\ell\big(\mathbf{f}_2(x), y\big)\right)+\frac{r}{2}\cdot \left(\ell\big(\bo-\mathbf{f}_1(x), y\big)- \ell\big(\bo-\mathbf{f}_2(x), y\big)\right)\right|\\
    \leq & \left|\big(1-\frac{r}{2}\big)\cdot \left(\ell\big(\mathbf{f}_1(x), y\big)-\ell\big(\mathbf{f}_2(x), y\big)\right)\right|+\left|\frac{r}{2}\cdot \left(\ell\big(\bo-\mathbf{f}_1(x), y\big)- \ell\big(\bo-\mathbf{f}_2(x), y\big)\right)\right|\\
    \leq &\underbrace{(1+\frac{|r|-r}{2})L}_{\text{defined as  }L^{r}}|\mathbf{f}_1(x)-\mathbf{f}_2(x)|.
\end{align*}
Note that $(1-\frac{r}{2})\geq \frac{r}{2}$ so we need to concentrate on the term $\ell(\bfx, y)$, what is more, $\underline{\ell}(\bfx, y)=\overline{\ell}(\bfx, 1-y)$. We then have:
\begin{align*}
  \overline{\ell^{\text{GLS}, r}}=(1-\frac{r}{2})\cdot \overline{\ell}+\frac{r}{2}\cdot \underline{\ell};\quad  \underline{\ell^{\text{GLS}, r}}=(1-\frac{r}{2})\cdot \underline{\ell}+\frac{r}{2}\cdot \overline{\ell};\\
\end{align*}
 Thus we have:
\begin{align*}
    \overline{\ell^{\text{GLS}, r}} - \underline{\ell^{\text{GLS}, r}} &= (1-\frac{r}{2})\cdot \overline{\ell}+\frac{r}{2}\cdot \underline{\ell} - (1-\frac{r}{2})\cdot \underline{\ell} - \frac{r}{2}\cdot \overline{\ell}\\
    &=(1-\frac{r}{2})\cdot \left(\overline{\ell}-\underline{\ell}\right)-\frac{r}{2}\cdot \left(\overline{\ell}-\underline{\ell}\right)\\
    &=(1-r)\cdot \left(\overline{\ell}-\underline{\ell}\right).
\end{align*}
Thus, we finally have:
\begin{align*}
   \max_{f\in\mathcal{F}}|R_{\text{{emp}}}^r(f)-R_{\text{exp}}^r(f)|\leq& (2+|r|-r)\cdot L\cdot \mathfrak{R}(\mathcal{F})+(1-r)\cdot \left(\overline{\ell}-\underline{\ell}\right)\cdot\sqrt{\frac{\log(1/\delta)}{2N}}.
\end{align*}
\end{proof}

\subsection{Proof of Proposition \ref{prop:bc_glsr}}
\begin{proof}
The risk minimization of backward correction is equivalent to:
    \begin{align*}
        \E_{(X, \nY)\sim\nD} \Big[\ell^{\leftarrow} \big(\bfX, \nY\big)\Big]=&\E_{(X, Y)\sim \D} \Big[\ell\big(\bfX, Y\big)\Big]. ~~~\text{(By Theorem 1 in \citep{patrini2017making})}
    \end{align*}
The risk minimization of forward correction is equivalent to:
    \begin{align*}
        \E_{(X, \nY)\sim\nD} \Big[\ell^{\rightarrow} (\bfX, \nY)\Big]=&\E_{(X, Y)\sim \D} \Big[\ell(\bfX, Y)\Big]. ~~~\text{(By Theorem 2 in \citep{patrini2017making})}
    \end{align*}
    Theorem 1 and 2 in \citep{patrini2017making} demonstrate that forward and backward corrected losses  equal the original loss $\ell$ computed on the clean data in expectation. Thus, for $r_{\text{LC}}=\frac{2e_0}{2e_0-1}$, by Theorem \ref{thm:glsr_conf} (adopt $r^*=0$), we have:
    \begin{align*}
        &\min_{f \in \mathcal F}~ \E_{(X, \nY)\thicksim \nD}\quad \Big[\ell\big(\bfX, \nY^{\text{GLS}, r_{\text{LC}}}\big)\Big]+\lambda_{\text{LC}}   \cdot \underbrace{ \mathbb{E}_{X, Y=1}   \Big[\ell\big(\bfX, 1\big) - \ell\big(\bfX, 0\big)\Big]}_{\text{Bias-LC}}\\
     =&\min_{f \in \mathcal F}~ \mathbb{E}_{(X, Y)\sim \D}  \Big[\ell\big(\bfX, Y\big)\Big] +\Big[e_0 + (1-2e_0)\cdot \frac{ r_{\text{LC}}}{2}\Big]\cdot \mathbb{E}_{(X, Y)\sim \D}  \Big[\ell\big(\bfX, 1-Y\big) - \ell\big(\bfX, Y\big)\Big]\\
     &+ e_{\Delta}\cdot (1-r_{\text{LC}}) \cdot \mathbb{E}_{X, Y=1}  \Big[\ell\big(\bfX, 0\big) - \ell\big(\bfX, 1\big)\Big] + \lambda_{\text{LC}} \cdot \mathbb{E}_{X, Y=1}  \Big[\ell\big(\bfX, 1\big) - \ell\big(\bfX, 0\big)\Big]\\
     =&\min_{f \in \mathcal F}~ \mathbb{E}_{(X, Y)\sim \D}  \Big[\ell\big(\bfX, Y\big)\Big]+ e_{\Delta}\cdot \Big(\frac{1}{1-2e_0} - \frac{1}{1-2e_0}\Big) \cdot \mathbb{E}_{X, Y=1}  \Big[\ell\big(\bfX, 0\big) - \ell\big(\bfX, 1\big)\Big] \\
     =&\min_{f \in \mathcal F}~ \mathbb{E}_{(X, Y)\sim \D}  \Big[\ell\big(\bfX, Y\big)\Big].
    \end{align*}
    Thus,
 \begin{align*}
    &\min_{f \in \mathcal F}~ \E_{(X, \nY)\sim\nD} \Big[\ell^{\leftarrow} \big(\bfX, \nY\big)\Big]=\min_{f \in \mathcal F}~ \E_{(X, \nY)\sim\nD} \Big[\ell^{\rightarrow} \big(\bfX, \nY\big)\Big]\\
    = &\min_{f \in \mathcal F}~ \E_{(X, \nY)\thicksim \nD} \Big[\ell\big(\bfX, \nY^{\text{GLS}, r_{\text{LC}}}\big)\Big]+\lambda_{\text{LC}} \cdot \underbrace{ \mathbb{E}_{X, Y=1}   \Big[\ell\big(\bfX, 1\big) - \ell\big(\bfX, 0\big)\Big]}_{\text{Bias-LC}}.
\end{align*}
\end{proof}

\subsection{Proof of Theorem \ref{thm:bc_glsr}}
\begin{proof}
Based on Proposition \ref{prop:bc_glsr}, when $e_{\Delta}=0$, $\lambda_{\text{LC}}=0$, we directly have:
\begin{align*}
    &\min_{f \in \mathcal F}~ \E_{(X, \nY)\sim\nD} \Big[\ell^{\leftarrow} \big(\bfX, \nY\big)\Big]=\min_{f \in \mathcal F}~ \E_{(X, \nY)\sim\nD} \Big[\ell^{\rightarrow} \big(\bfX, \nY\big)\Big]\\
    = &\min_{f \in \mathcal F}~ \E_{(X, \nY)\thicksim \nD} \Big[\ell\big(\bfX, \nY^{\text{GLS}, r_{\text{LC}}}\big)\Big].
\end{align*}
\end{proof}

\subsection{Proof of Theorem \ref{thm:cl_glsr}}
\begin{proof}
Note that
\begin{align*}
    &\min_{f \in \mathcal F}~ \E_{(X, \nY)\sim \nD} \Big[\ell_{\text{CL}}\big(\bfX,\nY\big)\Big]=\min_{f \in \mathcal F}~ \E_{(X, \nY)\sim \nD}\Big[\ell\big(\bfX, \nY\big)-\ell\big(\bfX, 1-\nY\big)\Big].
\end{align*}
We have:
\begin{align*}
    &\min_{f \in \mathcal F}~ \E_{(X, \nY)\thicksim \nD} \Big[ \ell\big(\bfX, \nY^{\text{GLS}, r_{\text{CL}}}\big)\Big]\\
    =&\min_{f \in \mathcal F}~ \E_{(X, \nY)\thicksim \nD} \Big[ \big(1-\frac{r_{\text{CL}}}{2}\big)\cdot \ell\big(\bfX, \nY\big) + \frac{r_{\text{CL}}}{2}\cdot \ell\big(\bfX, 1-\nY\big)\Big]\\
     \Leftrightarrow&\min_{f \in \mathcal F}~ \E_{(X, \nY)\thicksim \nD} \Big[ \ell\big(\bfX, \nY\big) + \frac{r_{\text{CL}}}{2-r_{\text{CL}}}\cdot \ell\big(\bfX, 1-\nY\big)\Big].
\end{align*}
When $r_{\text{CL}}\to -\infty$, we have $\frac{r_{\text{CL}}}{2-r_{\text{CL}}}\to -1$. Thus, 
\begin{align*}
    \min_{f \in \mathcal F}~ \E_{(X, \nY)\sim \nD} \Big[\ell_{\text{CL}}\big(\bfX,\nY\big)\Big]= \min_{f \in \mathcal F}~ ~&\E_{(X, \nY)\thicksim \nD} \Big[ \ell\big(\bfX, \nY^{\text{GLS}, r_{\text{CL}}\to -\infty}\big)\Big].
\end{align*}
\end{proof}

\subsection{Proof of Proposition \ref{prop:pl_glsr}}
\begin{proof}

Note that:
    \begin{align*}      &\mathbb{E}_{(X, \nY)\thicksim \nD}   \Big[\ell\big(\bfX, \nY\big)\Big]-\mathbb{E}_{(X, \nY)\thicksim \nD}   \Big[\ell\big(\bfX, \nY^{\text{GLS}, r}\big)\Big]\\
    =&\mathbb{E}_{(X, \nY)\thicksim \nD}  \Big[1-\big(1-\frac{r}{2}\big)\cdot \ell\big(\bfX, \nY\big)-\frac{r}{2}\cdot \ell\big(\bfX, 1-\nY\big)\Big]\\
     =&\frac{r}{2}\cdot\mathbb{E}_{(X, \nY)\thicksim \nD}  \Big[\ell\big(\bfX, \nY\big) -\ell\big(\bfX, 1-\nY\big)\Big].
    \end{align*}
And we have:
\begin{align*}
    &\mathbb{E}_{(X_i, \nY_i)\thicksim \nD} \Big[\ell\big(\bfXo, \tilde{Y_2}\big)\Big]\\
     =&\E_{X} \Big[\p(\nY=0)\cdot \ell\big(\bfX, 0\big)+\big(1-\p(\nY=0)\big)\cdot \ell\big(\bfX, 1\big)\Big]\\
     =& \E_{X,\nY=0}\Big[\p(\nY=0)\cdot \ell\big(\bfX, 0\big)+\big(1-\p(\nY=0)\big)\cdot \ell\big(\bfX, 1\big)\Big] \\
     &+\E_{X,\nY=1}\Big[\p(\nY=0)\cdot \ell\big(\bfX, 0\big)+\big(1-\p(\nY=0)\big)\cdot \ell\big(\bfX, 1\big)\Big]\\
     =& \E_{X,\nY=0}\Big[\p(\nY=0)\cdot \ell\big(\bfX, 0\big)+\big(1-\p(\nY=0)\big)\cdot \ell\big(\bfX, 1\big)\Big] \\
     &+\E_{X,\nY=1}\Big[\big(1-\p(\nY=0)\big)\cdot \ell\big(\bfX, 0\big)+ \p(\nY=0) \cdot \ell\big(\bfX, 1\big)\Big]\\
     &+\big(1-2\cdot \p(\nY=0)\big)\cdot \E_{X,\nY=1}\Big[ \ell\big(\bfX, 1\big) - \ell\big(\bfX, 0\big)\Big].
    \end{align*}
Thus, 
\begin{align*}
        &\min_{f \in \mathcal F}~ \mathbb E_{(X,\nY)\thicksim \nD}
     \Big[\ell_{\text{PL}}\big(\bfX,\nY\big)\Big]=\min_{f \in \mathcal F}~ \mathbb E_{(X,\nY)\thicksim \nD}\bigg[ \ell\big(\bfX, \nY\big)  - \ell\big(\bfXo, \tilde{Y_2}\big) \bigg]\\
        =&\min_{f \in \mathcal F}~ \mathbb E_{(X,\nY)\thicksim \nD} \Big[\ell\big(\bfX, \nY\big)\Big]  - \mathbb E_{(X_i,\nY_i)\thicksim \nD} \Big[\ell\big(\bfXo, \tilde{Y_2}\big) \Big]\\
        =&\min_{f \in \mathcal F}~  \E_{X,\nY=0} \Big[\ell\big(\bfX, 0\big) \Big]+\E_{X,\nY=1} \Big[\ell\big(\bfX, 1\big) \Big]\\
        &- \E_{X,\nY=0}\Big[\p(\nY=0)\cdot \ell\big(\bfX, 0\big)+\big(1-\p(\nY=0)\big)\cdot \ell\big(\bfX, 1\big)\Big] \\
     &-\E_{X,\nY=1}\Big[\big(1-\p(\nY=0)\big)\cdot \ell\big(\bfX, 0\big)+ \p(\nY=0) \cdot \ell\big(\bfX, 1\big)\Big]\\
     &-\big(1-2\cdot \p(\nY=0)\big)\cdot \E_{X,\nY=1}\Big[ \ell\big(\bfX, 1\big) - \ell\big(\bfX, 0\big)\Big]\\
     =&\min_{f \in \mathcal F}~  \E_{X,\nY=0} \Big[\big(1-\p(\nY=0)\big)\cdot \big[\ell\big(\bfX, 0\big)-\ell\big(\bfX, 1\big) \big] \Big]\\
     &+\E_{X,\nY=1} \Big[\big(1-\p(\nY=0)\big)\cdot \big[\ell\big(\bfX, 1\big)-\ell\big(\bfX, 0\big) \big] \Big]\\
     &-\big(1-2\cdot \p(\nY=0)\big)\cdot \E_{X,\nY=1}\Big[ \ell\big(\bfX, 1\big) - \ell\big(\bfX, 0\big)\Big]\\
     =&\min_{f \in \mathcal F}~  \E_{(X,\nY)\thicksim \nD} \Big[\big(1-\p(\nY=0)\big)\cdot \big[\ell\big(\bfX, \nY\big)-\ell\big(\bfX, 1-\nY\big) \big] \Big]\\
     &-\big(1-2\cdot \p(\nY=0)\big)\cdot \E_{X,\nY=1}\Big[ \ell\big(\bfX, 1\big) - \ell\big(\bfX, 0\big)\Big].
\end{align*}
 Thus, for $r_{\text{PL}}=2\cdot \p(\nY=1), \lambda_{\text{PL}}=1-r_{\text{PL}}$, we have:
 \begin{align*}
     &\E_{(X,\nY)\thicksim \nD}\Big[\ell_{\text{PL}}(\bfX, \nY\big)\Big]-\Bigg[\E_{(X,\nY)\thicksim \nD}\Big[\ell(\bfX, \nY\big)\Big]-\E_{(X, \nY)\thicksim \nD} \Big[\ell\big(\bfX, \nY^{\text{GLS}, r_{\text{PL}}}\big)\Big]\Bigg]\\
     =&\E_{(X,\nY)\thicksim \nD} \Big[\big(1-\p(\nY=0)\big)\cdot \big[\ell\big(\bfX, \nY\big)-\ell\big(\bfX, 1-\nY\big) \big] \Big]\\
     &-\big(1-2\cdot \p(\nY=0)\big)\cdot \E_{X,\nY=1}\Big[ \ell\big(\bfX, 1\big) - \ell\big(\bfX, 0\big)\Big]\\
     &-\frac{r_{\text{PL}}}{2}\cdot\mathbb{E}_{(X, \nY)\thicksim \nD}  \Big[\ell\big(\bfX, \nY\big) -\ell\big(\bfX, 1-\nY\big)\Big]\\
     =&\E_{(X,\nY)\thicksim \nD} \Big[\big(1-\p(\nY=0)-\p(\nY=1)\big)\cdot \big[\ell\big(\bfX, \nY\big)-\ell\big(\bfX, 1-\nY\big) \big] \Big]\\
     &-\big(2\cdot \p(\nY=1)-1\big)\cdot \E_{X,\nY=1}\Big[ \ell\big(\bfX, 1\big) - \ell\big(\bfX, 0\big)\Big]\\
     =&\lambda_{\text{PL}}\cdot \E_{X,\nY=1}\Big[ \ell\big(\bfX, 1\big) - \ell\big(\bfX, 0\big)\Big].
 \end{align*}
 And we can conclude that: 
 \begin{align*}
    \min_{f \in \mathcal F}~ \E_{(X, \nY)\sim \nD} \Big[\ell_{\text{PL}}(\bfX,\nY)\Big]= \min_{f \in \mathcal F}~ ~&\E_{(X, \nY)\thicksim \nD} \Big[\ell(\bfX, \nY) - \ell(\bfX, \nY^{\text{GLS}, r_{\text{PL}}})\Big]\\
    +&\lambda_{PL} \cdot \underbrace{\E_{X,\nY=1} \Big[ \ell(\bfX, 1) - \ell(\bfX, 0)\Big]}_{\text{Bias-PL}}.
\end{align*}

\end{proof}

\subsection{Proof of Theorem \ref{thm:pl_glsr}}
\begin{proof}
When $\p(\nY=0)=\p(\nY=1)$, according to Proposition \ref{prop:pl_glsr}, we have $\lambda_{PL}=0$ and:
\begin{align*}
    \min_{f \in \mathcal F}~ \E_{(X, \nY)\sim \nD} \Big[\ell_{\text{PL}}\big(\bfX,\nY\big)\Big]=& \min_{f \in \mathcal F}~ \E_{(X, \nY)\thicksim \nD} \Big[\ell\big(\bfX, \nY\big) - \ell\big(\bfX, \nY^{\text{GLS}, r_{\text{PL}}}\big)\Big]\\
    = &\min_{f \in \mathcal F}~ \E_{(X, \nY)\thicksim \nD} \Big[\frac{r_{\text{PL}}}{2} \cdot \ell\big(\bfX, \nY\big) -\frac{r_{\text{PL}}}{2}\cdot  \ell\big(\bfX, 1-\nY\big)\Big]\\
    \Leftrightarrow &\min_{f \in \mathcal F}~ \E_{(X, \nY)\thicksim \nD} \Big[ \ell\big(\bfX, \nY\big) - \ell\big(\bfX, 1-\nY\big)\Big].
\end{align*}
When $r_{\text{PL}}\to -\infty$, we further have:
\begin{align*}
    \min_{f \in \mathcal F}~ \E_{(X, \nY)\sim \nD} \Big[\ell\big(\bfX, \nY^{\text{GLS}, r_{\text{PL}}}\big)\Big]\Leftrightarrow&\min_{f \in \mathcal F}~ \E_{(X, \nY)\thicksim \nD} \Big[ \ell\big(\bfX, \nY\big) + \frac{r_{\text{CL}}}{2-r_{\text{CL}}}\cdot \ell\big(\bfX, 1-\nY\big)\Big]\\
    \Leftrightarrow &\min_{f \in \mathcal F}~ \E_{(X, \nY)\thicksim \nD} \Big[ \ell\big(\bfX, \nY\big) - \ell\big(\bfX, 1-\nY\big)\Big].
\end{align*}
Thus, Theorem \ref{thm:pl_glsr} is proved.
\end{proof}

\subsection{Proof of Theorem \ref{thm:robust_glsr}}
\begin{proof}
Note that the optimal $r$ that will cancel the impact of Term $\text{M-Inc1}$ is:
\begin{align*}
    r_{\text{opt}} := \frac{r^*-2e}{1-2e}.
\end{align*}
\begin{itemize}[leftmargin=*]
\vspace{-5pt}
    \item When $e<\frac{r^*}{2}$, $r_{\text{opt}}> 0$. In this case, learning LS with smooth rate $r_{\text{opt}}$ results in:
    \begin{align*}
        \min_{f \in \mathcal F}~ \mathbb{E}_{(X, \nY)\thicksim \nD} \Big[\ell(\bfX, \nY^{\text{GLS},r=r_{\text{opt}}})\Big]
    =\min_{f \in \mathcal F}~    \mathbb{E}_{(X, Y)\sim \D}   \Big[\ell\big(\bfX, Y^*\big)\Big],
    \end{align*}
    which yields $f_{\D}^*$;
    \item When $e=\frac{r^*}{2}$, $r_{\text{opt}}= 0$. Learning with the Vanilla Loss yields $f_{\D}^*$ since:
    $$\min_{f \in \mathcal F}~ \mathbb{E}_{(X, \nY)\thicksim \nD} \Big[\ell(\bfX, \nY)\Big]
    =\min_{f \in \mathcal F}~    \mathbb{E}_{(X, Y)\sim \D}   \Big[\ell\big(\bfX, Y^*\big)\Big];$$
    \item Similarly, when $e>\frac{r^*}{2}$, learning NLS with $r=r_{\text{opt}}< 0$ yields $f_{\D}^*$.
\end{itemize}
\end{proof}

\subsection{Proof of Theorem \ref{thm:robust_glsr_multi}}
\begin{proof}
Denote $p_i=\p(Y=i)$ as the clean label distribution, $\tilde{p}_i=\p(\nY=i)$ as the clean label distribution. Let $\epsilon'=\frac{K\cdot \epsilon}{K-1}$, we have:
    \begin{align*}
        & \E_{(X,\nY)\sim \nD}\Big[(1-r)\cdot \ell\big(\bfX, \nY\big)\Big]+\E_{X}\Big[\sum_{i\in [K]}\frac{r}{K}\cdot \ell\big(\bfX, i\big)\Big]\\
        =&\Bigg[\sum_{i\in [K]}\E_{(X,\nY)\sim \nD,Y=i}\Big[(1-r)\cdot  \ell\big(\bfX,\nY\big)\Big]\Bigg]+\E_{X}\Big[\sum_{i\in [K]}\frac{r}{K}\cdot \ell(\bfX, i)\Big]\\
     =&\Bigg[(1-r)\cdot\sum_{i\in [K]}\E_{(X,\nY)\sim \nD,Y=i}\Big[  \sum_{j\in[K]}T_{i,j}\cdot \ell\big(\bfX,\nY=j\big)\Big]\Bigg]+\E_{X}\Big[\sum_{i\in [K]}\frac{r}{K}\cdot \ell\big(\bfX, i\big)\Big]\\
     =&\Bigg[(1-r)\cdot\sum_{i\in [K]}\E_{X,Y=i}\Big[  (1-\epsilon')\cdot \ell\big(\bfX,i\big)+\sum_{j\in [K]}\frac{\epsilon'}{K}\cdot \ell\big(\bfX,j\big) \Big]\Bigg]+\E_{X}\Bigg[\sum_{i\in [K]} \frac{r}{K}\cdot \ell\big(\bfX, i\big)\Bigg]\\
         =&\Bigg[(1-r)\cdot\sum_{i\in [K]}\E_{X,Y=i}\Big[  \Big(1-\epsilon'\Big)\cdot \ell\big(\bfX,i\big)\Big]\Bigg]+\E_{X}\Bigg[\Big[\frac{(1-r)\cdot \epsilon'}{K}+\frac{r}{K}\Big]\sum_{j\in [K]} \ell\big(\bfX,j\big)\Bigg]\\
         =&\Bigg[\underbrace{(1-r)\cdot\Big(1-\epsilon'\Big)}_{:=c_3} \E_{(X,Y)\sim \D}\Big[ \ell\big(\bfX,Y\big)\Big]\Bigg]+\E_{X}\Bigg[\underbrace{\Big[\frac{(1-r)\cdot  \epsilon'}{K}+\frac{r}{K}\Big]}_{:=c_4}\sum_{j\in [K]} \ell\big(\bfX,j\big)\Bigg]\\
         =&\Bigg[\frac{c_3}{1-r^*}\cdot\E_{(X,Y)\sim \D}\Big[ \ell\big(\bfX,Y^*\big)-\frac{r^*}{K}\sum_{j\in[K]}\ell\big(\bfX,j\big)\Big]\Bigg]+\Bigg[c_4\cdot \E_{X}\Big[\sum_{j\in [K]} \ell\big(\bfX,j\big)\Big]\Bigg]\\
         =&\underbrace{\Bigg[\frac{c_3}{1-r^*}\cdot\E_{(X,Y)\sim \D}\Big[ \ell\big(\bfX,Y^*\big)\Big]\Bigg]}_{\text{True Risk}}+\underbrace{\Bigg[\Big(c_4-\frac{c_3\cdot r^*}{(1-r^*)\cdot K}\Big)\cdot \E_{X}\Big[\sum_{j\in [K]} \ell\big(\bfX,j\big)\Big]\Bigg]}_{\text{M-Inc1}}.
    \end{align*}
Adopting $r_{\text{opt}}=\frac{r^*-\epsilon'}{1-\epsilon'}$, with a bit of math, the weight of Term M-Inc1 becomes 0 and 
\begin{align*}
    & \E_{(X,\nY)\sim \nD}\Big[ \ell\big(\bfX,Y^{\text{GLS},r_{\text{opt}} }\big)\Big]\\=&\E_{(X,\nY)\sim \nD}\Big[(1-r_{\text{opt}})\cdot \ell\big(\bfX, \nY\big)\Big]+\E_{X}\Big[\sum_{i\in [K]}\frac{r_{\text{opt}}}{K}\cdot \ell\big(\bfX, i\big)\Big]\\=&\Bigg[\frac{c_3}{1-r^*}\cdot\E_{(X,Y)\sim \D}\Big[ \ell\big(\bfX,Y^*\big)\Big]\Bigg]\\
    \Leftrightarrow & \E_{(X,Y)\sim \D}\Big[ \ell\big(\bfX,Y^*\big)\Big].
\end{align*}
\end{proof}

\end{document}